\documentclass[a4paper,12pt]{article}

\usepackage[margin=1in]{geometry}
\usepackage{graphicx} 
\usepackage[most]{tcolorbox}
\usepackage{natbib}
\usepackage{amsmath}
\usepackage{booktabs}
\usepackage{caption}
\usepackage{subcaption}
\usepackage{url}
\usepackage{amssymb}
\usepackage{pifont}
\usepackage{makecell}
\usepackage{float}
\usepackage{tabularx}
\newcommand{\cmark}{\checkmark}
\newcommand{\xmark}{\ding{55}}
\usepackage{rotating}
\usepackage[colorlinks=true, linkcolor=black, citecolor=gray,urlcolor=black]{hyperref}
\usepackage{placeins}
\usepackage{authblk}
\usepackage{fontawesome5} 
\usepackage{array}
\usepackage{multirow}
\usepackage{titlesec}
\usepackage{enumitem}

\usepackage[table]{xcolor}
\usepackage{ragged2e}

\usepackage[framemethod=tikz]{mdframed}

\definecolor{warnbd}{HTML}{FB8C00}
\definecolor{warnbg}{RGB}{205,140,60}

\definecolor{HOFP}{RGB}{134,170,201}   
\definecolor{HOFC}{RGB}{170,157,199}  
\definecolor{HOFS}{RGB}{244,230,159}     
\definecolor{HOFE}{RGB}{187,213,164}    

\definecolor{HoFDefinition}{RGB}{90,110,130}

\newtcolorbox{warningbox}{
  colback=warnbg!6,       
  colframe=warnbd!70,     
  boxrule=0pt,            
  borderline west={1.4pt}{0pt}{warnbd},
  left=6pt,
  right=6pt,
  top=5pt,
  bottom=5pt,
  sharp corners,          
  enhanced                
}

\newtcolorbox{pillarcard}[3]{
  colback=#3!15,
  colframe=#3!85,
  boxrule=0.4pt,
  arc=3pt,
  left=6pt,
  right=6pt,
  top=6pt,
  bottom=6pt,
  fonttitle=\bfseries,
  title={#1},
  before upper={\textit{#2}\par\smallskip}
}

\newtcolorbox{definitionbox}{
  colback=white,
  colframe=HoFDefinition!90,
  boxrule=0pt,
  borderline west={1.2pt}{0pt}{HoFDefinition},
  left=6pt,
  right=6pt,
  top=4pt,
  bottom=4pt
}

\newcommand{\HoFbarSec}[1]{\textcolor{#1}{\rule[-0.55ex]{3pt}{2.8ex}}\hspace{8pt}}

\newcommand{\HoFShortOrLong}[2]{\ifstrempty{#1}{#2}{#1}}

\NewDocumentCommand{\Psubsection}{ O{} m }{%
  \subsection[\HoFShortOrLong{#1}{#2}]{\texorpdfstring{\HoFbarSec{HOFP}}{}#2}%
}
\NewDocumentCommand{\Csubsection}{ O{} m }{%
  \subsection[\HoFShortOrLong{#1}{#2}]{\texorpdfstring{\HoFbarSec{HOFC}}{}#2}%
}
\NewDocumentCommand{\Ssubsection}{ O{} m }{%
  \subsection[\HoFShortOrLong{#1}{#2}]{\texorpdfstring{\HoFbarSec{HOFS}}{}#2}%
}
\NewDocumentCommand{\Esubsection}{ O{} m }{%
  \subsection[\HoFShortOrLong{#1}{#2}]{\texorpdfstring{\HoFbarSec{HOFE}}{}#2}%
}

\definecolor{Xinyuan}{RGB}{228, 26, 28}   
\definecolor{Eren}{RGB}{0, 0, 255}  
\definecolor{Lixiao}{RGB}{77, 175, 74}   
\definecolor{Ransalu}{RGB}{152, 78, 163}  

\title{Humanoid Factors: Design Principles for \\AI Humanoids in Human Worlds}

\author[1,$\dagger$]{Xinyuan Liu}
\author[2]{Eren Sadikoglu}
\author[1]{Ransalu Senanayake}
\author[3]{Lixiao Huang}

\affil[1]{School of Computing and Augmented Intelligence, Arizona State University, AZ, USA}
\affil[2]{School of Manufacturing Systems and Networks, Arizona State University, AZ, USA}
\affil[3]{Human Systems Engineering, Arizona State University, AZ, USA}

\affil[$\dagger$]{Corresponding author: \texttt{xinyua11@asu.edu}}

\date{}

\date{}

\begin{document}

\maketitle

\begin{abstract}
    Human factors research has long focused on optimizing environments, tools, and systems to account for human performance. Yet, as humanoid robots begin to share our workplaces, homes, and public spaces, the design challenge expands. We must now consider not only factors for humans but also factors for humanoids, since both will coexist and interact within the same environments. Unlike conventional machines, humanoids introduce expectations of human-like behavior, communication, and social presence, which reshape usability, trust, and safety considerations. In this article, we introduce the concept of \emph{humanoid factors} as a framework structured around four pillars---physical, cognitive, social, and ethical---that shape the development of humanoids to help them effectively coexist and collaborate with humans. This framework characterizes the overlap and divergence between human capabilities and those of general-purpose humanoids powered by \emph{AI foundation models}. To demonstrate our framework's practical utility, we then apply the framework to evaluate a real-world humanoid control algorithm, illustrating how conventional task completion metrics in robotics overlook key human cognitive and interaction principles. We thus position humanoid factors as a foundational framework for designing, evaluating, and governing sustained human--humanoid coexistence.
\end{abstract}

\begin{figure}[h]
    \centering
    \includegraphics[width=0.95\textwidth]{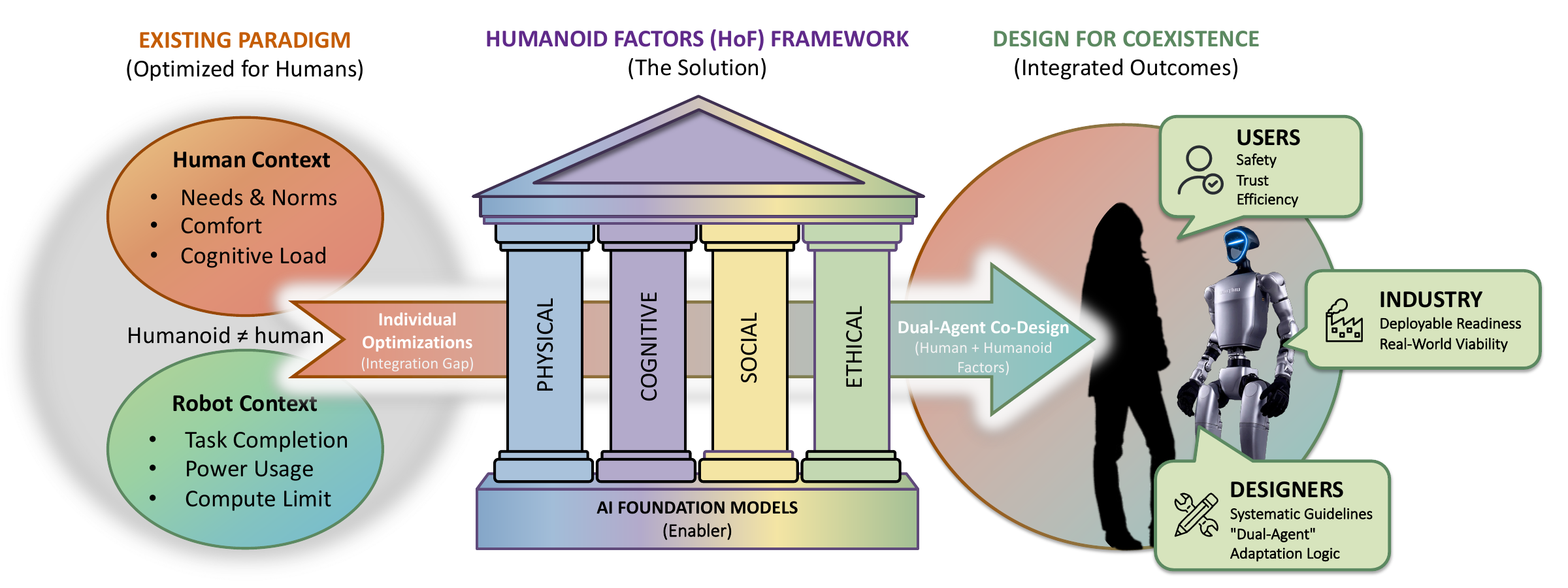}
    \caption{The proposed Humanoid Factors framework for humanoid design.}
    \label{fig:venn}
\end{figure}

\newpage
\tableofcontents
\newpage

\section{Introduction}

For most of human history, the design of environments and tools has been centered on a single intelligent agent—the human. Human Factors (HF) emerged to optimize safety, efficiency, and usability within this human-only design paradigm. Robots, when introduced, were built as specialized instruments: function dictated form, and the robot’s role was tightly bound to a narrow operational context.


\begin{definitionbox}
\textbf{Definition}. \underline{Human Factors (HF)} and ergonomics use knowledge of \underline{human} abilities and limitations to design systems, organizations, jobs, machines, tools, and consumer products for safe, efficient, and comfortable \underline{human use}.
\flushright--\cite{HFES_Definition}
\end{definitionbox}

That assumption is now being overturned. Robots are no longer designed only as fixed, single-purpose tools; instead, we are beginning to see \emph{generalist} humanoid robots that can flexibly perform many tasks in everyday environments. This transformation is driven by large-scale \emph{AI foundation models}: machine learning models trained on vast and diverse datasets and adaptable to a wide range of downstream tasks and contexts~\citep{bommasani2021opportunities}. When such models are embedded in humanoid robots, they provide a common substrate for perception, reasoning, and control, enabling robots to generalize across tasks and domains, and to improve through continuous robot learning at scale~\citep{team2025gemini}. Humanoids powered by these models are no longer merely preprogrammed machines; they are becoming systems that can interpret goals, learn from experience, and adapt their behavior to different situations and user preferences. As a result, the design challenge becomes mutual adaptation of humans and humanoids in shared, learning-rich environments. This new reality introduces a fundamental question:
\emph{How should we design environments, tasks, and interactions for ecosystems in which humans and humanoids coexist?}

For understanding human performance, the field of Human Factors provides a foundation framework for analyzing how people perceive, act, and make decisions. A central principle within this field is Human-Centered Design~\citep{xu2023enabling, shneiderman2020human}, a design philosophy and process that ensures products, systems, and services are developed with a deep understanding of users' needs, contexts, and experiences. Implicit in these approaches is the assumption that humans are the primary intelligent agents within the system.

As humanoid robots, which are also now intelligent, become co-actors in these same environments, the foundational goals of Human Factors remain, yet their application must broaden to encompass how both humans and humanoids jointly perceive, act, and adapt. For humanoid robots to acquire human-like sensorimotor and cognitive skills, they must continually fine-tune their underlying AI foundation models. As a result, humanoids inherit not only our physical affordances~\citep{gundawar2025pac}---doorways, stairs, tools---but also our social affordances~\citep{kaufmann2007culture}, such as expectations of eye contact, emotional tone, and moral reasoning. These overlapping capabilities and expectations create both opportunities for seamless collaboration and risks of misalignment. 

To address this gap, we introduce Humanoid Factors, a framework that extends Human Factors thinking to the design and evaluation of systems in which humanoid robots and humans interact as joint participants. 

\begin{tcolorbox}
\textbf{Definition}. \underline{Humanoid Factors (HoF)} is the \underline{design of humanoids} and their surrounding environments to \underline{enable humanoids} to safely, efficiently, and intelligibly adapt to and sustain operation in human-centric settings.
\end{tcolorbox}

We therefore introduce Humanoid Factors (HoF) as the complementary counterpart to Human Factors. Whereas HF optimizes environments, tools, and workflows for human performance, HoF specifies the design, evaluation, and adaptation requirements for humanoid performance within human-built spaces. Together, HF and HoF enable design for coexistence: a mixed-agent ergonomics in which human and humanoid capabilities are specified separately (HF for humans, HoF for humanoids) and then integrated through interaction protocols, safety envelopes, and task-allocation rules. Practically, this means starting from generalist humanoids and customizing them for applications such as manufacturing, caregiving, education, and beyond.

In this view, design becomes ecological: the separation keeps accountability clear yet invites integration through a broader philosophy of design for coexistence, in which humans and humanoids operate as coordinated agents within shared environments. Generalist humanoids built on AI foundation models can serve as base platforms, customized through HoF-guided adapters, constraints, and peripherals to meet domain needs. When integrated with HF considerations, these co-adaptive systems establish shared ergonomics (physical compatibility), shared cognition (mutual legibility of intent), shared sociality (trust and communication norms), and shared ethics (accountability and dignity in collaboration). 


A growing need therefore exists to articulate a framework of Humanoid Factors that accounts for the physical, cognitive, social, and ethical pillars shaping the design and operation of humanoid robots in human environments. Rather than positioning this as a simple extension of Human Factors, we view it as a complementary perspective that focuses on how humanoid systems---endowed with human-like embodiment and adaptive intelligence powered by foundation models---can be systematically designed, evaluated, and refined to function safely and intelligibly alongside people. 

Broadening the current interaction-centric, moment-level, task-specific focus of robots and human~\citep{rodriguez2021human}, HoF adopts a system-level, lifecycle-level, and environment-centric perspective to ask: How should a humanoid be designed so that it can safely, predictably, and acceptably live and operate in human environments? Although research on humanoid robots has advanced rapidly in the past three years~\citep{tong2024advancements, cao2025humanoid}, current literature remains fragmented across robotics \citep{cao2025humanoid}, machine learning \citep{xiao2025robot}, and ergonomics \citep{electronics14234734}, with limited guidance on consistent principles for humanoid form, behavior, and integration in shared environments. We organize HoF into four pillars because the main challenges of humanoid coexistence arise from four recurring but interacting sources: embodiment in human-built spaces (Physical), internal reasoning and action selection (Cognitive), coordination with human expectations and communication norms (Social), and questions of safety, responsibility, and acceptable conduct (Ethical). These pillars are intended as analytic lenses rather than mutually exclusive boxes; in deployment, the same failure often spans several at once. This paper therefore develops Humanoid Factors as a dedicated framework to consolidate these perspectives and to provide a foundation for systematic study, evaluation, and co-design of humanoids and the environments they inhabit.

The article proceeds by establishing the Humanoid Factors framework in Section 2 and analyzing how AI foundation models enable Humanoid Factors in Section 3. As an example, we then demonstrate how to utilize this framework to evaluate an AI algorithm deployed in a real humanoid robot in Section 4. Finally, Section 5 proposes considerations for research and product design, followed by concluding remarks in Section 6.

\section{Humanoid Factors Framework}
\label{sec:framework}
As humanoid robots enter the same spaces of humans, the design must shift to a dual agent, co-adaptive view in which humans and humanoids learn the capabilities, limitations and behavioral regularities of each other over time~\citep{nikolaidis2013human, ajoudani2018progress}. For such design thinking, we require a complementary lens: {Humanoid Factors}, the systematic study of how humanoid embodiment, intelligence, and social presence shape their performance in human-designed environments, and how these factors must be understood to enable effective human--humanoid coexistence~\citep{dautenhahn2007socially, fong2003survey}.

This need arises because humanoids differ fundamentally from conventional automation. Unlike purpose-built machines designed for specific tasks, humanoids combine human-like morphology with increasingly general-purpose intelligence, creating three fundamental shifts:
\begin{enumerate}
    \item \textbf{From automation to embodiment.} Unlike conventional automation, humanoids inherit human environmental affordances by aligning with human scale, reach, and signaling. This alignment allows them to leverage existing tools and workflows, assume monotonous or dangerous tasks, and collaborate more naturally and safely with people~\citep{takayama2008beyond}. From an AI perspective, embodiment in human environments also mitigates data scarcity in training AI foundation models: humanoids can leverage large volumes of human activity recordings and demonstrations, which are largely unavailable to most other robotic platforms. These factors jointly enable the deployment of general-purpose robotic capabilities in human-built spaces without extensive retrofitting. Humanoids share a partial overlap in capabilities with humans: they can perform certain tasks humans can, while humans retain abilities that humanoids do not, and vice versa. This asymmetric overlap gives rise to distinct strengths, limitations, and failure modes that must be accounted for in design and evaluation~\citep{gundawar2025pac}.
    \item \textbf{From machines as tools to social actors.} Unlike conventional machines that are interpreted primarily as tools, humanoids' human-like appearance and motion cues lead people to attribute intent, understanding, and social competence to them. As a result, users naturally expect humanoids to communicate, coordinate, and respond in human-like ways, even when their underlying capabilities do not fully support such expectations~\citep{mori2012uncanny}. This expectation shift affects how instructions are given, how errors are interpreted, and how trust is formed during interaction. When humanoid behavior falls short of these socially inferred expectations, it can lead to confusion, misuse, or erosion of trust, highlighting the need to explicitly account for expectation management, transparency, and communicative behavior in humanoid system design and evaluation.
    \item \textbf{From technical safety to socio-ethical governance.} Traditional automation emphasizes technical safety guarantees such as collision avoidance, fault tolerance, and reliability under predefined conditions~\citep{brunke2022safe, corso2020interpretable, lasota2017survey}. In contrast, humanoids introduce broader trust, safety, and ethical considerations that arise from their anthropomorphic form, autonomy, and intelligence \citep{winfield2018ethical}. Because humanoids operate in shared social spaces and are perceived as agents rather than tools, questions of accountability, appropriate behavior, and responsibility become inseparable from system performance. This shift necessitates governance frameworks that extend beyond engineering safeguards to include social norms, ethical constraints, and institutional oversight, addressing not only whether a humanoid can act safely, but whether it should act, under whose authority, and with what consequences when failures occur.
\end{enumerate} 

These shifts indicate that the promise of humanoids is not defined solely by capability, such as mobility or dexterity, but by compatibility with human environments, norms, and expectations. Effective deployment therefore requires legible behavior, reliable communication, and sustained co-adaptation to human practices and social conventions~\citep{hancock2011meta}. Achieving this compatibility demands that embodiment, cognition, social interaction, and ethics be treated as co-equal design constraints rather than isolated considerations.

\begin{figure}[h]
    \centering
    \includegraphics[width=\textwidth]{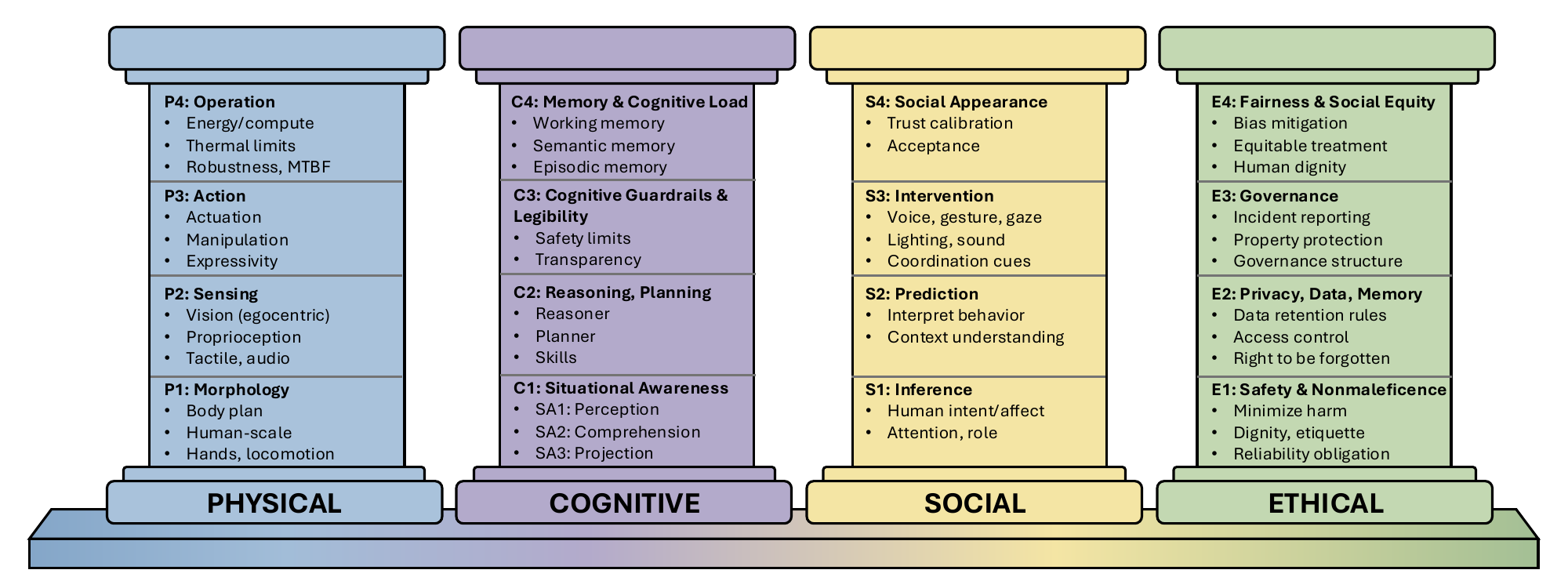}
    \caption{The Four Pillars of Humanoid Factors.}
    \label{fig:4_pillars}
\end{figure}

As summarized in Figure~\ref{fig:4_pillars}, we therefore structure the Humanoid Factors framework around four pillars---\textbf{Physical (P)}, \textbf{Cognitive (C)}, \textbf{Social (S)}, and \textbf{Ethical (E)}---that capture the key ways humanoid systems differ from both traditional automation and humans themselves. Each pillar emphasizes a distinct evaluative question: whether the humanoid fits the physical environment and task demands, whether it reasons and acts in ways that remain legible under uncertainty, whether it coordinates with human expectations and social norms, and whether its behavior remains accountable and acceptable. The layers within each pillar should be read as a pragmatic decomposition for design and evaluation, not as a claim that real deployments can be neatly partitioned into independent modules. In practice, failures often cut across pillars: for example, a motion that deviates from human expectation may simultaneously reflect embodiment limits, policy-design choices, social miscalibration, and safety implications.

\Psubsection{Physical Pillar (\textbf{P})}

The Physical Pillar establishes how a humanoid's body enables it to operate, signal, 
and adapt within human environments. Rather than enumerating hardware variations, 
we organize the physical design space into a four–layer embodiment model that 
distinguishes (i) what the robot \emph{is}, (ii) how it \emph{perceives}, (iii) how it \emph{acts and communicates}, and 
(iv) how it \emph{sustains operation}. This \emph{layered embodiment structure} avoids overloading the 
definition of ``humanoid'' with incidental engineering choices and instead 
grounds the Physical pillar in functional principles that generalize across 
morphologies and future platforms.

\subsubsection{Layer P1: Morphology (What the humanoid \emph{is})}
The morphological layer defines the invariant, architecture-level physical properties of a humanoid robot---its fixed body structure, proportions, and degrees of freedom~\citep{tong2024advancements}. These properties are predetermined and remain constant during operation, setting the physical preconditions for compatibility with human-designed environments and affordances. In other words, this layer answers the fundamental question: \emph{Is this system physically 
capable of inhabiting and acting within human environments?}

To answer this question, we first define what a humanoid is:
\begin{definitionbox}
\textbf{Definition.} We define a \underline{humanoid} as a robot designed to exploit human-scale advantages (e.g., doors, stairs, furniture, tools) and to present human-compatible signals for coordination. A platform is in scope if it combines:
(i) a torso and head/neck sensing cluster for human-compatible signaling,
(ii) two upper limbs enabling \emph{bimanual} manipulation, and
(iii) \emph{bipedal} locomotion and anthropomorphic hands that support \emph{in-hand} manipulation.
A realistic face is not required for the humanoid; what matters is \emph{legible} posture, gaze, and approach behavior.
\end{definitionbox}

\paragraph{Morphology variants.} As detailed in Table~\ref{tab:m-variants}, platforms that embody a subset of these morphological properties (i.e., M-Static, M-Wheel, M-Biped) are within scope for ablation studies in research and industrial settings (see~\citet{tong2024advancements} for such recent variants). However, when reporting results on these variants, we recommend explicit classification using the defined IDs and a statement regarding generalization limits relative to the canonical anthropomorphic humanoid (M-Full).

\begin{table}[h]
\centering
\caption{Embodiment classes based on morphological ({M}) features. The \emph{Class ID} uses descriptive suffixes to allow for flexible categorization. All-terrain capability denotes the physical endowment necessary for mobility in human environments, including stairs and other non-planar surfaces.}
\label{tab:m-variants}

\resizebox{\textwidth}{!}{%
    \renewcommand{\arraystretch}{1.2} 
    \begin{tabular}{l l c c c c c}
    \toprule
    & & \multicolumn{2}{c}{\textbf{Locomotion}} & \multicolumn{3}{c}{\textbf{Structure \& Manipulation}} \\
    \cmidrule(lr){3-4} \cmidrule(lr){5-7}
    \textbf{Class ID} & 
    \textbf{Descriptor} & 
    \makecell{\textbf{Body}\\\textbf{Type}} & 
    \makecell{\textbf{All-terrain}\\\textbf{Capable}} & 
    \makecell{\textbf{Human}\\\textbf{Size}} & 
    \makecell{\textbf{Torso}\\\textbf{+ Arms}} & 
    \makecell{\textbf{Dexterous}\\\textbf{Hands}} \\
    \midrule

    \textbf{M-Static} & 
    Upper-body & 
    Static & 
    \xmark & 
    Optional & 
    \cmark & 
    Optional \\

    \textbf{M-Wheel} & 
    Mobile Manipulator & 
    Wheeled & 
    \xmark & 
    \cmark & 
    \cmark & 
    Optional \\

    \textbf{M-Biped} & 
    Locomotion Biped & 
    Legged & 
    \cmark & 
    \cmark & 
    \cmark & 
    \xmark \\

    \textbf{M-Full} & 
    Full Humanoid & 
    Legged & 
    \cmark & 
    \cmark & 
    \cmark & 
    \cmark \\

    \bottomrule
    \end{tabular}%
}
\end{table}

\begin{figure}[h]
    \centering
    \includegraphics[width=0.95\textwidth]{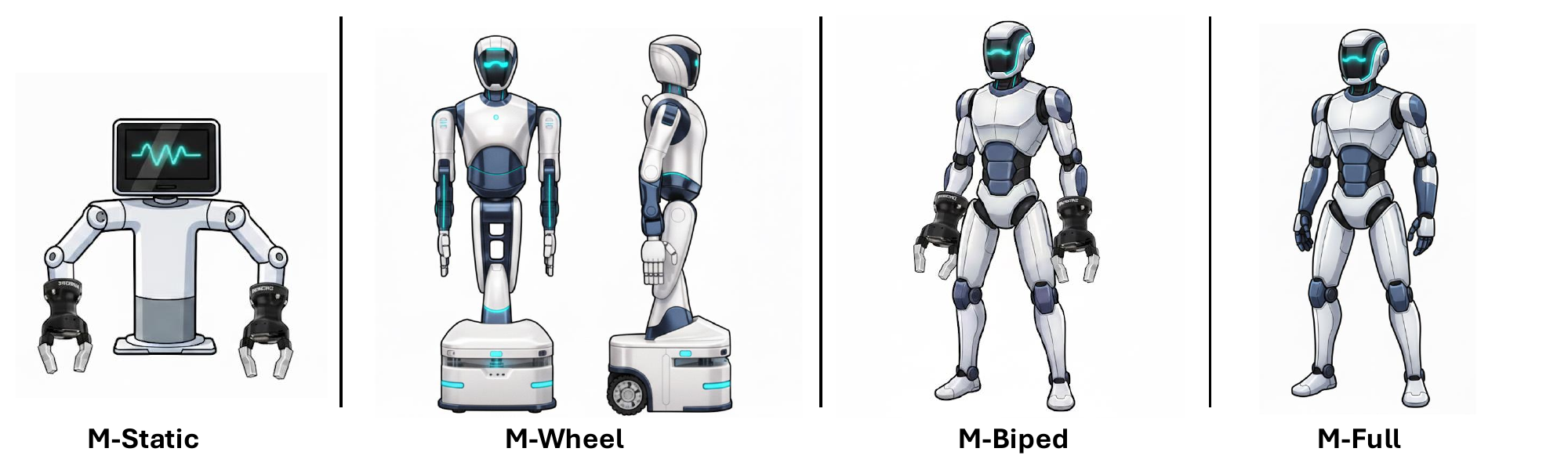}
    \caption{Visual examples of the four morphological embodiment classes: M-Static (Upper-body), M-Wheel (Mobile Manipulator), M-Biped (Locomotion Biped), and M-Full (Full Humanoid).}
    \label{fig:venn}
\end{figure}

\subsubsection{Layer P2: Sensing (What the humanoid \emph{perceives})}

This layer defines the robot’s ability to \emph{acquire continuous streams of perceptual information} about its operational surroundings---i.e., the physical and social environment in which it acts---with sufficient fidelity to support autonomous operation and safe coexistence with humans. Within the Humanoid Factors framework, sensing is intentionally restricted to the hardware and low-level signal processing that digitize the environment (e.g., imaging, depth estimation, proprioceptive information), without yet possessing semantic understanding. To operate effectively, the sensing architecture must provide functional coverage comparable to that of human senses—potentially exceeding or falling short in specific modalities—while delivering inputs that can be transformed by higher-level perceptual and situational-awareness processes within the Cognitive Pillar (Section~\ref{sec:cognitive}).

\paragraph{Egocentric vision and active gaze control.}
While robots can be equipped with a full $360^{\circ}$ field of view, as in autonomous vehicles, the computational constraints of humanoids make it important to prioritize a forward-facing visual stream. This supports shared visual perspectives with humans and facilitates more natural interaction. This necessitates camera placement at a human-like height (approx. 1.5m--1.7m) and a neck mechanism with at least 2-degrees of freedom (e.g., pan-tilt). This articulation allows the robot to decouple its Field of View (FOV) from its body orientation, enabling ``glancing'' or smooth pursuit~\citep{barnes2008cognitive} behaviors that signal attention without aggressive torso rotation. This decoupling also implies that manipulation should be feasible across the full range of head motion, since many current manipulation algorithms are highly sensitive to viewpoint and often fail when the visual perspective deviates from their training data~\citep{vasudevan2025strategic}. Furthermore, the vision system must balance near-field high-resolution depth (stereo or structured light within $0.3$--$1.0$ m) for precise manipulation with wide-angle peripheral coverage to detect navigation hazards at the feet. In line with this, there is a growing trend toward incorporating additional \emph{eye-in-hand cameras}~\citep{fu2024mobile} to provide high-fidelity visual feedback during fine manipulation.

\paragraph{Somatic and auditory sensing.}
Visual occlusion is inevitable during close-contact manipulation. Consequently, reliance on vision must be augmented by high-bandwidth proprioception (e.g., joint encoders to measure joint angles, joint velocities, and joint accelerations; IMUs to measure whole body position and acceleration; force-torque sensors) and tactile sensor arrays on the fingers and palms. These somatic sensors detect contact make/break events, slip, and object hardness, providing low-level physical information. Additionally, microphone arrays utilizing sound source localization (SSL) and beamforming are important to solve the ``cocktail party problem,'' allowing the robot to spatially isolate and orient toward a human speaker in noisy environments~\citep{huang2023egocentric,pian2024continual,zhao2025multimodal}.

\paragraph{Sensory parallels and extensions.} From a human factors perspective, humanoid sensing can mirror the human division between key senses (vision, touch, audition, and proprioception) and auxiliary integrative senses (gaze control, balance, slip detection, and sound localization)~\citep{harvey2024senses}. While several of these modalities admit direct one-to-one counterparts (e.g., cameras for vision, joint encoders for proprioception), others, such as nociception, are not explicitly sensed but can be inferred from combinations of force, torque, tactile, and thermal signals. Moreover, humanoids can be equipped with task-specific sensing beyond the human repertoire, including infrared, gas, or chemical sensors, to support specialized applications such as search and rescue. Further, humanoids are not constrained to a fixed sensory hierarchy: they can maintain multiple, partially redundant sensing modalities and dynamically weight and fuse them based on task demands and environmental conditions~\citep{taparia2025vlc}.

\begin{warningbox}
{\bfseries A cross-domain lesson.} In October 2024, NHTSA opened Preliminary Evaluation PE24-031 after four crashes in which Tesla vehicles, with FSD engaged, encountered reduced visibility (sun glare, fog, airborne dust); one involved a fatal pedestrian strike and another a reported injury~\citep{nhtsa2024fsd}. The core failure was a system that continued operating with high confidence even after its primary optical inputs were degraded. Sensing must include its own \emph{quality limits}: build explicit low-level visibility/occlusion/glare diagnostics to signal early for safe stop, maintain partially redundant modalities to preserve basic situational awareness when primary sensing degrades.
\end{warningbox}

\subsubsection{Layer P3: Action (How the humanoid \emph{acts and communicates})}
This layer characterizes how a robot physically acts and communicates in its environment, bridging embodied motion with socially interpretable behavior. Within the Humanoid Factors framework, the Action layer is restricted to actuation and interaction surfaces, defining the physical means by which robots execute tasks and convey social signals, prior to higher-level planning or cognitive interpretation. Effective action in human environments requires both functional capability and expressive compatibility with human expectations, enabling safe, legible, and socially appropriate interaction.

\paragraph{Actuation.}
Recent advances in high–torque-density electric actuators and compact harmonic-drive transmissions (motors and precision gears) have lowered the cost and complexity of modern agile humanoid designs, compared to earlier platforms such as ASIMO~\citep{sakagami2002intelligent} that required large, custom actuation assemblies to achieve similar torque and precision. By reducing joint size and increasing torque density, these actuation advances have made human-scale \emph{kinematic} designs (how a robot's joints are arranged and move) feasible within practical mass and cost constraints. In shared human environments, robots benefit from human-like ranges of motion because people intuitively understand and predict movements that resemble their own and that respect established proxemic norms. Such kinematics reduce cognitive effort, improve perceived safety, and make robot actions more socially legible during close-proximity interaction. However, achieving socially appropriate motion depends not only on whether a robot can reach a target, but on how its body moves while doing so. Designs with limited degrees of freedom restrict how a task can be physically expressed, often forcing rigid or intrusive postures. In contrast, while additional joint freedom increases mechanical complexity and financial cost, it enables robots to choose how a motion is expressed, supporting socially appropriate postures and avoidance behaviors without compromising task execution~\citep{ficuciello2015variable,li2024comprehensive}. 

Kinematics alone do not define physical utility. Performance parameters such as maximum payload, actuation torque, and gait speed (for mobile platforms) define the robot's functional envelope, but they also shape how powerful, fast, or potentially dangerous the robot appears to nearby people. Therefore, robots with greater physical capability require stricter safety mechanisms, including force-limiting control and conservative motion policies, to maintain trust and perceived safety~\citep{song2020toward, hu2023research}. Consequently, design trade-offs---balancing control complexity, computational cost, and physical capability against safety---must be carefully considered when choosing actuators for humanoids.

\paragraph{Interaction surfaces and expressivity.}
In addition to mobility, effective action in socially situated robots extends beyond locomotion alone. For this purpose, we introduce expressivity as a complementary design dimension, which captures how robots communicate intent, internal state, and social meaning through action, independent of where or how fast they move. As shown in Table~\ref{tab:e-classes}, we categorize expressivity using Expressivity Classes that describe progressively richer channels for social signaling. These classes intentionally separate \textit{appearance} (e.g., facial realism) from \textit{functional communicative fidelity} (e.g., gaze dynamics, timing). Higher expressivity can improve the speed and reliability of trust formation and coordination in tasks where social cues matter, but the effect size depends on task context, cultural expectations, and user priors~\citep{admoni2017social,fu2023robot}. Thus, claims about social outcomes arising from physical designs should be validated with controlled user studies measuring objective coordination metrics (e.g., task completion time, error rate) and subjective scales (e.g., trust, perceived comfort).

Lastly, designers must weigh expressivity benefits against costs and risks: increased mechanical and software complexity, higher mass and energy consumption, potential maintenance overhead, and social risks such as the uncanny valley or unintended anthropomorphic deception. Where possible, trade-off analyses and empirical evaluation plans should accompany design recommendations.

\begin{table}[h]
\centering
\caption{Expressivity classes ({X}) defining social signal fidelity.}
\label{tab:e-classes}
\renewcommand{\arraystretch}{1.2}
\begin{tabular}{l l l}
\toprule
\textbf{Class ID} & \textbf{Modality} & \textbf{Signaling Capabilities} \\
\midrule
\textbf{X-Abs} & Abstract & LEDs, Audio, Body orientation \\
\textbf{X-Vis} & Screen/Digital & 2D Gaze, Animated icons \\
\textbf{X-Mech} & Mechanical & 3D Head pose, Actuated features \\
\textbf{X-Real} & Realistic & Synthetic skin, Micro-expressions \\
\bottomrule
\end{tabular}
\end{table}
    


\subsubsection{Layer P4: Operation (How the humanoid \emph{sustains operation})}

A humanoid that cannot sustain operation beyond a short demonstration is effectively a laboratory artifact rather than a deployable system. Within the Humanoid Factors framework, the operational layer characterizes the energy-compute envelope, mechanical robustness, and maintenance infrastructure. 

\paragraph{The energy–compute–thermal trilemma.} Sustained operation of humanoid robots is constrained by tightly coupled limits on energy availability, onboard computation, and heat dissipation.
\begin{enumerate}
    \item Energy. Humanoid robots are powered by onboard batteries that must balance capacity, charging speed, and battery cycle life. These batteries supply motors, onboard compute (GPUs), control electronics, sensors, and actuators, with motors and GPUs dominating power consumption. Unlike wheeled platforms, humanoids incur a continuous energy cost to maintain balance and posture, even when idle.
    \item Compute. Modern humanoids rely on foundation models run on onboard GPUs. While these models enable autonomy and social interaction, high-throughput inference draws significant power, rapidly depletes batteries, and generates concentrated heat near electronic components.
    \item Thermal limits. Electronics and actuators must operate within safe temperature bounds. GPUs require active cooling, and sustained motor loads generate heat that degrades efficiency and component lifetime~\citep{du2025flexible}. For humanoids running increasingly large foundation models, maintaining safe operating temperatures over extended periods demands more substantial cooling.
\end{enumerate}
Increasing compute, energy capacity, or cooling improves performance along one dimension but exacerbates constraints in the others, making sustained operation a system-level trade-off rather than a single-component optimization.

\paragraph{Robustness and fall resilience.}
Robustness in this layer refers specifically to physical and operational durability, rather than algorithmic or cognitive fault tolerance. Sustained operation depends on how the body tolerates contact, long-term structural fatigue (e.g., joint wear, cable fatigue, seal degradation), and foreseeable human misuse (e.g., leaning, grabbing for support, incidental pushing), as well as inevitable hardware failures. While soft skins provide interaction safety (P3), operationally they also serve to protect expensive internal mechanisms from dust, liquid, and impact. Enclosures should meet Ingress Protection (IP) levels comparable to industrial tools (e.g., IP54), as defined in IEC 60529~\citep{IEC60529:1999}, to withstand common environmental contaminants. However, the defining operational risk for humanoids is gravity. Systems must feature fall resilience—the ability to detect imminent stability loss and execute protective ``tuck and roll'' strategies to minimize hardware damage, or engage mechanical brakes to prevent collapse during power loss~\citep{zhang2025review}. These design choices directly determine mean time between failures (MTBF) and mean time to repair (MTTR), shaping whether humanoids can deliver reliable, recoverable operation in real-world human environments~\citep{Lienig2017Reliability}.

\paragraph{Serviceability and ecosystem.} Even highly robust systems will fail over time, making maintenance logistics and ease of repair central to long-term viability in real-world deployments. To minimize downtime, hardware must emphasize modularity; complex end-effectors or sensor heads should be designed as field-replaceable units (FRUs) to avoid shipping entire chassis for minor repairs, which is challenging for most users. User-facing interfaces must provide engineering-grade telemetry and standardized ``service modes'' that decouple actuation from compute for safe updates and debugging. Life-cycle considerations, including battery replacement intervals and component recyclability, must be integrated early to avoid the platform becoming e-waste after its first major failure cycle.



\paragraph{The ergonomic gap.}
While the morphological layer ensures humanoids fit into human spaces, the operational layer reveals a critical divergence. Human environments are optimized for biological constraints; humans fatigue and incur musculoskeletal stress, whereas humanoids face battery drainage, thermal saturation, and actuator wear. Consequently, ``human-centric'' design does not automatically equal ``humanoid-compatible.'' Bridging this ergonomic gap requires co-designing workstations and tasks around these \emph{dual constraints}, such as balancing human time-on-task and rest with robotic duty cycles and thermal cooldown periods. Many of the operational concerns discussed here align with established industrial standards for safety, environmental robustness, and ergonomics (e.g., IEC 60529, ISO 10218, ISO/TS 15066), though humanoid deployment introduces novel endurance and co-design challenges not fully addressed by existing frameworks in human--humanoid environments.

\begin{pillarcard}{Physical Pillar (P) Card}{Embodiment constraints for humanoid coexistence}{HOFP}
\textbf{Mandate.}
The Physical Pillar defines how a humanoid’s embodiment, sensing, actuation, and
operational envelope enable safe, legible, and sustained operation in human-designed
environments. Physical design here governs compatibility with human scale, affordances,
and expectations, rather than task feasibility alone.

\textbf{Failure without P.}
Ignoring physical constraints leads to brittle deployment: humanoids that complete tasks
but move unsafely or intrusively, exhibit short-lived demonstrations due to energy or
thermal limits, or project apparent capability that exceeds controllable, trustworthy
behavior.

\textbf{Layers.}
P1~Morphology (body plan, scale);
P2~Sensing (egocentric, proprioceptive, tactile);
P3~Action (actuation, manipulation, expressivity);
P4~Operation (energy, compute, robustness, serviceability).

\textbf{Interfaces.}
Physical limits bound cognition (latency, awareness), shape social legibility (motion,
timing), and establish ethical baselines for safety and nonmaleficence.

\textbf{Evaluation.}
Beyond task success: motion legibility, MTBF and recovery behavior,
operational sustainability (energy usage), physical robustness, and alignment between apparent and controllable capability.
\end{pillarcard}

\Csubsection{Cognitive Pillar (\textbf{C})}
\label{sec:cognitive}

The Cognitive Pillar specifies how a humanoid transforms sensing into understanding, intention, and action. Rather than prescribing algorithms or architectures, we organize cognition into a layered reasoning model that distinguishes (i) how the robot interprets what is \emph{happening}, (ii) how it decides what to \emph{do next}, (iii) how it \emph{self-regulates} its behavior to remain safe, legible, and intelligible, and (iv) how it manages memory and cognitive load to \emph{sustain} effective operation over time. This question-driven structure clarifies what sensor measurements mean for the robot’s internal state and behavior---progressing from situational awareness to planning, guardrails, and memory---while treating computational limits, latency, training coverage, and operational design domain (ODD) constraints as fundamental to cognitive design rather than afterthought implementations. 

In humanoids, cognition is implemented through AI systems, and this pillar organizes the capabilities those systems must support---whether realized explicitly through traditional modular AI components or implicitly through modern monolithic AI foundation models---into a coherent layered structure.



\subsubsection{Layer C1: Situational Awareness (What is \emph{happening})}
An intelligent humanoid must first understand the situation it is operating in before it can act effectively. This layer captures classic cognitive functions including attention management, mental model formation, and anticipatory cognition, which together determine what the system notices, how it interprets the situation, and how it prepares for what may happen next. This situational awareness (SA) unfolds hierarchically~\citep{endsley2021situation,endsley2000theoretical,sanneman2022situation}:
\begin{itemize}[label={}]
  \item \textbf{SA1 -- Perception:} detect and register relevant elements in the environment---people, objects, surfaces, affordances, and salient events---through the integration of visual, auditory, proprioceptive, tactile, and other cues, including mechanisms for attention management that prioritize task-relevant and safety-critical information under limited perceptual and computational resources~\citep{taparia2025vlc}.
  \item \textbf{SA2 -- Comprehension:} make sense of the perceived elements by understanding their roles, relationships, and significance for the ongoing task, including what is safe, risky, or requiring attention, thereby maintaining an internal mental model of the environment that supports shared understanding with human partners.
  \item \textbf{SA3 -- Projection:} anticipate how the situation is likely to evolve in the near future, including human actions, robot motions, and emerging hazards, to support timely and appropriate responses.
\end{itemize}

Each SA stage requires distinct forms of verification, including detection precision and recall at SA1, semantic correctness and grounding at SA2, and calibration and horizon accuracy at SA3. We highlight the importance of developing rigorous benchmarks for testing these aspects in AI models~\citep{gundawar2025pac,sferrazza2024humanoidbench,zhang2025vlabench}.

\begin{warningbox}
{\bfseries A cross-domain lesson.} The 2005 NASA DART (Demonstration for Autonomous Rendezvous Technology) mission involved an autonomous spacecraft that failed to maintain accurate navigation data during autonomous rendezvous with another satellite, resulting in a collision~\citep{nasa2006dart}. This mission represents a classic case where autonomous navigation did not maintain situational awareness.
\end{warningbox}

\subsubsection{Layer C2: Reasoning, Planning, \& Execution (What to \emph{do next})}
Once a humanoid has established situational awareness, it must determine how to act. This layer concerns the cognitive processes by which a humanoid selects goals, reasons about alternatives, and translates intent into executable behavior under uncertainty, time pressure, and resource constraints. Rather than prescribing a single control paradigm, this layer characterizes the functional roles required to support human-like decision-making, including goal prioritization, planning under uncertainty, and adaptive execution.

\paragraph{Reasoning, planning, and control separation.}
To support both flexibility and evaluation, modern humanoid systems benefit from a \emph{conceptual} separation (this does not mean the architecture should be modular) between three cognitive functions:
\begin{enumerate}
  \item \textbf{Reasoning:} evaluates uncertainty~\citep{Senanayake_2025}, assesses whether sufficient information is available to act, and considers alternative courses of action (e.g., ``Should I proceed, slow down, or ask for clarification?''). This role corresponds to meta-cognitive~\citep{dunlosky2008metacognition} decision checks studied in human factors, such as confidence assessment and risk sensitivity~\citep{endsley2017toward}.
  \item \textbf{Planning:} translates high-level intent into structured subgoals and selects appropriate skills to achieve them. Planning must handle underspecified instructions, competing objectives, and dynamic environments, while maintaining internally consistent goal hierarchies. Generating intermediate goals and rationales supports downstream legibility and verification.
  \item \textbf{Execution:} realizes planned actions through low-latency motor control and skill execution, adapting online to disturbances and human motion. Execution prioritizes responsiveness and safety while remaining consistent with the planner’s intent. 
\end{enumerate} 

This separation mirrors well-established distinctions in human cognition between deliberation, intention formation, and motor execution, and provides clear interfaces for evaluation and failure diagnosis. Where helpful for intuition and cross-disciplinary grounding, map these functions to brain systems: higher-level planning and goal maintenance are associated with prefrontal and premotor networks, online sensorimotor transformation and execution with motor cortex and cerebellar circuits, and episodic/relational memory with hippocampal systems~\citep{henschke2023engaging}. These analogies are heuristic (robots are not brains), but they are useful for deriving evaluation primitives—e.g., short-term working memory capacity, predictive timing accuracy, and consolidation of episodic experience.

\paragraph{Decision-making under uncertainty and time constraints.}
Human environments rarely permit fully informed or perfectly timed decisions. Accordingly, this layer emphasizes decision-making under uncertainty~\citep{kochenderfer2015decision}, including explicit trade-offs between speed and accuracy~\citep{Peltzer2022arxiv}, confidence thresholds for action, and graceful degradation when information, compute, or energy budgets are limited. Planning horizons and execution timing should reflect human-comfortable tempos, supporting coordination and reducing surprise during joint action. For instance, as we will \emph{demonstrate in Section~\ref{sec:experiment}, classical models such as Fitts' Law}~\citep{mackenzie1992fitts} remain informative for setting timing targets in pointing, handover, and placement tasks and for defining acceptable speed–accuracy envelopes. We use Fitts' Law in Section~\ref{sec:experiment} in this narrow sense: as a task-matched diagnostic for timing legibility in reaching-like motion, not as a universal cognitive metric for all humanoid behavior.

\paragraph{Goal management and adaptability.}
This layer also governs how goals are maintained, revised, or abandoned over time. Humanoids must reconcile long-term objectives with short-term contingencies, resolve conflicts between safety, efficiency, and user intent, and adapt plans in response to unexpected events. These capabilities align with human factors research on goal switching, workload management, and adaptive behavior in complex systems.


\subsubsection{Layer C3: Cognitive Guardrails (How can humanoids \emph{self-regulate})}

\paragraph{Components of guardrails.} Humanoid AI must be built with explicit guardrails: constraints, monitors, and fallbacks. Alignment with humans requires \emph{constraints} that define non-negotiable boundaries: no physical harm (e.g., enforced through force, torque, and collision-energy limits with compliant actuation), no unsafe social behavior (e.g., coercion, deception), and no ethical or legal violations. As risk rises, the system must deliberately modulate its behavior---reducing speed, stiffness, and autonomy while increasing legibility through gaze, motion exaggeration, and clear audio/visual cues. When monitors detect boundary violations or loss of confidence, \emph{fallbacks} must activate: slow-down modes, protective stops, human handoff, or entry into a reduced-capability safe mode. Explicit pause and shutdown mechanisms must always be human-overrideable and non-resistant.

\paragraph{Implementing guardrails.} Low-level controllers of the humanoid must enforce physical safety, mid-level policies must encode social and ethical rules, and high-level planners must reason about risk. AI models must operate within these bounds, using prespecified rules, human demonstrations, preference inference~\citep{christiano2017deep}, and constrained or shielded learning~\citep{alshiekh2018safe} that blocks unsafe actions. These requirements can be achieved through \emph{continuous monitoring} of uncertainty, force/torque, proximity, and human intent cues. The goal is not peak performance, but a humanoid that continuously self-regulates to remain safe, legible, and aligned with human norms and expectations. At the same time, the AI system must be able to organically learn and adapt its guardrails, allowing them to evolve with context, users, and environments, and thereby remain feasible for long-term deployment and operation.


\subsubsection{Layer C4: Memory \& Cognitive Load (How to \emph{balance cognition})}

\paragraph{Memory.} Runtime cognition in humanoids should be supported by a layered memory architecture that separates information by timescale, function, and governance requirements. This separation is necessary to support real-time control, structured reasoning, and personalization, while enabling verification and ethical oversight. \emph{Working memory} should maintain current goals, task-relevant state, and short-term dialogue history. Because it operates within strict real-time control loops, this layer must be constrained in capacity and access latency. Verification should therefore focus on memory eviction policies (e.g., least-recently-used) and multimodal alignment, such as synchronizing visual buffers with audio streams. \emph{Semantic memory} should store task schemas, object categories, and skill parameters, providing the ontology required for reasoning and planning. Verification of this layer should emphasize coverage (completeness of the domain model) and freshness (continued validity of stored facts over time). \emph{Episodic memory} should capture site- and user-specific experiences to enable learning and personalization, subject to explicit privacy, retention, and audit policies. Because this layer introduces heightened ethical and storage risks, verification must include strict access controls, compliance with retention and right-to-be-forgotten requirements, and utility constraints ensuring retrieved episodes remain relevant to current tasks~\citep{dechant2025episodic}. These memory layers should be designed to manage human cognitive load: effective working memory reduces repetition and monitoring demands, semantic memory supports predictable and legible behavior, and episodic memory enables personalization while introducing deliberate trade-offs around privacy and trust.
      
\paragraph{Cognitive load.}

Cognitive load in humanoids refers to the resource demands imposed by perception, planning, learning, and interaction under bounded compute (or latency), energy, and thermal budgets. Managing this load is a core requirement for correct, safe, and responsive behavior. Key dimensions include where high-cost perception and planning are executed (edge vs. cloud), uncertainty thresholds that trigger graceful degradation or human handoff~\citep{rahman2021run}, and energy or thermal limits that gate compute-intensive behaviors. Practical controls are therefore essential: priority queues for safety-critical pipelines, formal quality-of-service contracts for perception and planning latencies, and explicit backoff or handoff policies when compute or thermal budgets are exhausted. These mechanisms ensure that humanoid's cognitive resources are allocated predictably under real-world constraints.

Humanoids should also be designed to minimize the cognitive load imposed on humans during interaction. From a human factors perspective, cognitive load reflects the mental effort required to understand, predict, and coordinate with a system, and is commonly decomposed into \emph{intrinsic load} (task complexity), \emph{extraneous load} (interaction and interface inefficiencies), and \emph{germane load} (effort devoted to learning)~\citep{sweller2011cognitive}. For example, a humanoid that signals its intent through gaze and posture before acting, maintains consistent motion and turn-taking patterns, and proactively communicates uncertainty or the need for assistance allows humans to anticipate behavior without continuous monitoring, thereby reducing extraneous cognitive load. Human cognitive load can be assessed using established HF methods, including subjective measures (e.g., NASA-TLX), behavioral indicators (task time, error rates, interruption recovery), and physiological signals in controlled studies. Importantly, failures to manage humanoid-side cognitive load—such as delayed responses or inconsistent behavior—often translate directly into increased human cognitive burden. Ultimately, we need design that can effectively optimize both human and humanoid cognitive loads~\citep{zou2025synergy,sanneman2023information}.\\

\begin{pillarcard}{Cognitive Pillar (C) Card}{Bounded autonomy and intelligible decision-making}{HOFC}
\textbf{Mandate.}
The Cognitive Pillar governs how a humanoid interprets situations, selects actions, and
self-regulates behavior under uncertainty. It ensures that autonomy remains legible,
predictable, and bounded over time rather than brittle or overconfident.

\textbf{Failure without C.}
Absent cognitive structure, humanoids exhibit opaque decision-making, inappropriate
persistence, or delayed responses that increase human cognitive load and erode trust.
Such failures often appear as behavioral surprises rather than explicit errors.

\textbf{Layers.}
C1~Situational Awareness;
C2~Reasoning, Planning, and Execution;
C3~Cognitive Guardrails and Legibility;
C4~Memory and Cognitive Load Management.

\textbf{Interfaces.}
Cognition is bounded by physical limits (latency, sensing), shapes social coordination
through legibility and timing, and enforces ethical constraints via explicit guardrails.

\textbf{Evaluation.}
Appropriateness of intervention, stability of behavior under uncertainty, recovery after
failure, ability to reason, and human cognitive workload.
\end{pillarcard}

\Ssubsection{Social Pillar (\textbf{S})}

Socially intelligent robots are designed to engage in social interactions with \emph{social agents} through the use of social cues, communication, adaptation, and interpretation of human behavior~\citep{dautenhahn2007socially, fong2003survey, breazeal2003toward}. These social agents include humans as well as other humanoids and intelligent robots. In this work, we focus primarily on human--humanoid social interactions, while noting that many of the same mechanisms and design principles extend to humanoid--humanoid and broader multi-agent settings.

The social ability of a humanoid encompasses the cognitive, affective, communicative, and behavioral competencies that allow the humanoid to engage in coordinated, cooperative, or collaborative activities while maintaining human social norms and supporting shared goals. In action-based human-AI teaming research, such as the DARPA Artificial Social Intelligence for Successful Teams (ASIST) program, the social intelligence is demonstrated through the ability to infer human states (e.g., knowledge, beliefs, and workload), predict their next actions, and intervene to improve team performance~\citep{huang2025establishing}.  Building on this perspective, we decompose the Social Pillar into a four-layer model that distinguishes (i) what humans are \emph{thinking or feeling}, (ii) what they are likely to \emph{do next}, (iii) how and when the humanoid should \emph{intervene}, and (iv) how the humanoid’s appearance and behavior shape human \emph{expectations}.


\subsubsection{Layer S1: Inference (What humans are \emph{thinking or feeling})}
Humanoids need to infer human internal states from observable behavior, context, and interaction history. This includes estimating knowledge, goals, emotions, stress, workload, and beliefs. For humanoid robots, effective inference of humans requires integrating multimodal signals (e.g., gaze, posture, speech, task performance) with models of human cognition and affect. These inferences must be not only accurate but well-calibrated, as they form the foundation for intelligent assistance and socially appropriate behavior.

\paragraph{Theory of mind.} Inference is closely related to Theory of Mind (ToM)---the ability to attribute mental states such as beliefs, desires, and intentions to others and to reason about how these states guide behavior. Traditionally, ToM was studied in psychology using static images to interpret character relations~\citep{Brunet2000API} and people's emotions~\citep{baron2001reading}. Recent work has begun to explore analogous capacities in LLMs, including their ability to represent and express emotions or empathic responses~\citep{tint2024expressivityarenallmsexpressinformation, zhang2025decoding}. In dynamic human-robot interaction, ToM must be computational and predictive. Recent studies in robotics have formalized this through the concept of ``Latent Theory of Mind,'' where the human's mental states (goals, beliefs) are modeled as latent (hidden) variables that the robot must continuously estimate from observed actions to predict future behavior~\citep{he2025latent}.

\subsubsection{Layer S2: Prediction (What they are likely to \emph{do
next})}
This layer focuses on predicting future human actions, intents, feelings, and interaction trajectories given the inferred internal states and the current task context. Whereas S1 asks what a person is thinking, feeling, or trying to do now, S2 asks what is likely to happen next if nothing changes, and how different humanoid actions might alter that trajectory.

\paragraph{Timescale.} Predictive social intelligence can operate at very short time scales, such as anticipating whether a person is about to reach for a particular object, step into a humanoid's path, or speak in a conversation, as well as at longer horizons, such as forecasting whether a user is likely to abandon a task, ignore safety instructions, or gradually disengage from an educational activity. In the ASIST program, for example, AI systems predicted whether players would enter particular rooms or rescue certain victims based on room content and player profiles (e.g., gaming experience, personality, spatial abilities), and then adapted prompts and assistance accordingly~\citep{huang2025establishing}.

\paragraph{Beyond accurate prediction.} We need AI models that can be trained on large-scale sequences of human behavior, language, and interaction traces, learning statistical regularities in how people act in different situations. From a Humanoid Factors standpoint, prediction must be selective, counterfactual, and ethically bounded. Predictions should be task-relevant, supporting safety, efficiency, workload management, and social comfort rather than speculative inference for its own sake. They should consider not only what humans will do if the humanoid remains passive, but also how human behavior is likely to change in response to different candidate robot actions, enabling the humanoid to reason about the consequences of intervening versus staying silent.

\subsubsection{Layer S3: Intervention (How the humanoid should \emph{intervene})}
This layer addresses how the humanoid intervenes to influence team and task outcomes, based on the inferences and predictions from S1 and S2. 

\paragraph{Modes of intervention.} Humanoid robots can realize interventions through a rich combination of modalities. Natural language, whether spoken or written, allows them to explain, suggest, or negotiate in ways that align with everyday human communication, and recent advances in LLMs make such interaction more flexible and context-sensitive than in prior generations of robots. Embodied channels---including gaze, facial or body expressions for robots with expressive morphologies, light and sound patterns, haptic cues, and physical movements in the shared workspace---provide additional avenues for subtle, layered intervention. Because humanoids can act on the physical environment, they can also ``intervene'' by adjusting their motion, posture, or spatial behavior. For example, a humanoid that recognizes a human partner is still completing a prior step and deliberately delays handing over the next plate illustrates an intervention implemented through motion rather than speech, reducing time pressure and stress without verbal instruction~\citep{huang2015adaptive}.

\paragraph{Implicit vs. explicit intervention.}
Interventions can be explicit or implicit. Explicit interventions leverage language, sound, or visual cues and, with the prevailing capabilities of LLMs, enable natural language interaction that is far more flexible and context-sensitive than in prior generations of social robots~\citep{tellex2011understanding,wang2024lami}. Implicit interventions are implemented typically through embodied behavior, whereby the humanoid influences team and task outcomes by modulating timing, pose, or spatial configuration. For example, slowing humanoid's motion and increasing clearance when handing over an object to signal uncertainty or caution, or repositioning its body to guide human attention without verbal instruction. In addition to physical embodiment, recent studies indicate that LLMs can also produce implicit social signals—such as hedging, emphasis, or hesitation—through language, complementing embodied forms of intervention~\citep{bianchi2024well,tint2024expressivityarenallmsexpressinformation}. Cooperatively, these modes allow humanoid robots to combine linguistic and physical social signals, selecting or blending interventions based on context.

\begin{warningbox}
{\bfseries  A lesson from the past: the ``uncanny valley" trap.}
The deployment of hyper-realistic robots like Geminoid HI-1 and Sophia triggered a massive failure in user trust~\citep{abc2025humanoid}. While Geminoid's ``dead" gaze caused visceral avoidance, Sophia's expressive face misled the public into attributing high intelligence to simple scripts. When users realized the deception, the resulting backlash forced the industry to abandon deceptive anthropomorphism in favor of stylized, transparent designs that manage expectations rather than mislead them.



\end{warningbox}

\subsubsection{Layer S4: Social Appearance (How \emph{appearance matters})}
While the physical form of a robot is covered under the Physical Pillar, its \emph{social interpretation} belongs here. A humanoid's appearance acts as a social affordance, setting immediate expectations for intelligence, capability, and behavior. 

\paragraph{Morphology--expectation alignment.} Humans tend to expect to interact with humanoid robots more like they interact with other, relative to non-humanoid forms~\citep{fink2012anthropomorphism,sacino2022human}. Adult-sized humanoids invite expectations of adult-like competence and communication style, whereas smaller, neotenic designs might invite expectations of reduced competence but higher approachability and curiosity~\citep{stein2022power}. S4 thus modulates how S1–S3 are perceived and how users form their own internal models of the humanoid. If the robot looks highly capable, users initialize their internal model with an expectation of sophisticated inference and prediction, effectively assuming a strong S1-S2 stack. When the humanoid’s actual social competence falls short, users experience a prediction error that can lead to rapid loss of trust, disuse, or misuse. Conversely, robots that appear less capable than they are may elicit under-trust and under-utilization. Aligning social appearance with actual capability is therefore critical for calibrated trust and for stable human expectations about the robot’s future behavior.

Humanoids' social abilities are a key factor in their perceived trustworthiness. When the humanoid robots demonstrate social skills that undershoot or overshoot what their appearance and framing suggest, users may lose trust, become disengaged, or over-rely on the system~\citep{sanneman2020trust}. Designing social features that are well matched to task requirements and human expectations, and explicitly communicating the robot’s actual capabilities and limitations, are essential for calibrated trust and appropriate patterns of interaction. As exposure to humanoids with diverse morphologies and capabilities increases, human mental models are likely to change organically, reshaping expectations.

\paragraph{Behavior--expectation alignment.}
Behavioral cues such as motion smoothness, tempo, hesitation, and spatial timing provide continuous signals about competence, confidence, and intent, often overriding initial appearance-based expectations. Similarly, communication style---including politeness, hedging, tone, and turn-taking---shapes how humans interpret the robot’s authority, uncertainty, and social role. Further, from a \emph{social science perspective}, effective participation in a society depends on norm adherence~\citep{durkheim1982rules}, role adoption~\citep{biddle2013role}, shared intentionality, the maintenance of common ground capabilities~\citep{clark1996using}, and rule-based cooperation~\citep{ostrom2011background} that shape how intelligence is expressed, constrained, and interpreted in social contexts. Assistant-like behavior invites deference and guidance-seeking, whereas teammate-like behavior promotes shared responsibility and coordination. As a result, human assessments of the humanoid’s capabilities are continually updated through interaction, with even subtle mismatches between observed behavior and inferred competence leading to recalibration of trust and engagement. AI systems must therefore be capable of adapting to these evolving expectations to sustain effective social interaction.


\paragraph{Mutual adaptation for coexistence.}
Ultimately, social appearance is not a fixed property of form, but an emergent contract that is continuously renegotiated through interaction. As people observe the humanoid’s behavior over time, their expectations converge toward what the robot repeatedly demonstrates; likewise, repeated exposure can also shift what users consider ``normal'' or acceptable for a humanoid teammate. For stable \emph{coexistence} in shared human environments, this \emph{adaptation must be mutual}: the human updates an internal model of the robot, but the robot should also infer the human’s evolving expectations, social norms, and role framing (e.g., whether the user is treating it as an assistant, tool, or teammate). By sensing and modeling cues of human comfort, attention, uncertainty, and reliance, the humanoid can proactively calibrate its motion, communication style, and level of initiative to match the user and context. This reciprocal modeling reduces expectation mismatches, supports calibrated trust~\citep{zhou2026spread}, and enables interaction to converge toward a shared understanding. Thus, the robot is not simply present in human spaces, but actively evolves to integrate into them as a coherent and cooperative team member. 

\begin{pillarcard}{Social Pillar (S) Card}{Legible interaction and expectation alignment}{HOFS}
\textbf{Mandate.}
The Social Pillar specifies how humanoids interpret, anticipate, and influence human
behavior to enable coordinated, comfortable, and norm-consistent interaction. It governs
how intent, confidence, and uncertainty are communicated through language and motion.

\textbf{Failure without S.}
Ignoring social dynamics leads to expectation mismatch: humanoids that are technically
correct yet disruptive, intrusive, or misleading. These failures manifest as distrust,
misuse, or disengagement rather than task-level breakdowns.

\textbf{Layers.}
S1~Inference of human state;
S2~Prediction of human behavior;
S3~Intervention and coordination;
S4~Social appearance and expectation setting.

\textbf{Interfaces.}
Social performance depends on physical expressivity and timing, cognitive legibility and
confidence signaling, and ethical constraints on persuasion and authority.

\textbf{Evaluation.}
Trust stability, comfort and acceptance measures, coordination efficiency, intervention
appropriateness, and long-term human reliance patterns.
\end{pillarcard}

\Esubsection{Ethical Pillar (\textbf{E})}

As humanoid robots become more capable, autonomous, and embedded in human environments, their design and deployment must be guided by a structured ethical framework. Rather than treating ethics as an external checklist or a set of post hoc constraints, we organize ethical considerations into a layered framework that distinguishes (i) how humanoids should treat humans, (ii) how humanoids should handle \emph{data and memory}, (iii) how \emph{responsibility and governance} are assigned across the system’s lifecycle, and (iv) who is seen, heard, or \emph{marginalized} by humanoids. This question-driven structure positions ethical reasoning as integral to the humanoid’s operational design domain (ODD), shaping behavior, data use, and decision-making from real-time interaction to long-term deployment and oversight, and grounding ethical performance in concrete, evaluable principles rather than abstract norms. These layers progress from immediate individual safety, through internal data handling, to system-level oversight and societal impact.

\subsubsection{Layer E1: Safety \& Nonmaleficence (How humanoids \emph{treat humans})}
This layer addresses a fundamental ethical requirement for humanoid robots: that their actions toward humans minimize harm while preserving dignity, well-being, and appropriate human agency. Safety and nonmaleficence are not limited to physical injury prevention, but extend to psychological harm, social disruption, erosion of trust, and misuse arising from unreliable or poorly calibrated behavior. 

\paragraph{Purpose of humanoids.} How should humanoid robots treat humans? Humanoid robots are developed to support and enhance human quality of life by assisting with tasks such as cleaning (e.g., HRP-2~\citep{Okada2005HumanoidMG}), teaching (e.g., Nao~\citep{gouaillier2008nao}), cooking (e.g., life-sized humanoid cooking systems~\citep{watanabe2013cooking}), and entertaining (e.g., Sophia~\citep{HansonRoboticsSophia}). These activities may be performed autonomously by the humanoid or interdependently with humans in shared tasks. From an ethical standpoint, the purpose of humanoids is not merely task completion, but \emph{beneficial participation} in human activities. Accordingly, performance standards should be grounded in human-centered benchmarks (e.g., ANSI/HFES 400-2021 Human Readiness Level Scale in the System Development Process~\citep{ansi_hfes_400_2021}) rather than task completion rate. Robustness and reliability are therefore ethical obligations: failures, erratic behavior, or brittle performance in human environments can impose cognitive, emotional, or safety burdens on users even when no physical harm occurs.

\paragraph{Kindness, respect, and human dignity.}
Beyond functional safety, humanoid robots should interact with humans in ways that reflect social respect and care. Prior work on explanation and communication in AI emphasizes that interaction is a fundamentally social process governed by human expectations and conversational norms, rather than a purely informational exchange. As emphasized in Miller’s perspective on explanation~\citep{miller2019explanation,miller2023explainable}, humans evaluate artificial systems as social actors: interactions are judged not only by content, but by why an explanation is given, when it is provided, how it is framed, and for whom it is intended—making over-explanation, under-explanation, or poorly timed communication potential sources of harm. Ethical treatment therefore includes socially meaningful forms of ``kindness to humans,'' such as appropriate politeness and deference to human norms, as well as transparent communication of uncertainty and limitations. In collaborative settings, this also entails role clarity, ensuring that the humanoid’s level of initiative and authority does not undermine human autonomy or dignity. Accordingly, explainable AI methods tailored to humanoids~\citep{Sagar2024arxiv_xaitcavrob,anjomshoae2019explainable,molnar2020interpretable} should be treated as core ethical requirements to ensure socially appropriate interaction and preserve human autonomy and dignity.

\subsubsection{Layer E2: Privacy \& Data Ethics (What humanoids \emph{remember})}
Humanoids are mobile, multimodal sensing platforms that continuously capture video, audio, and interaction traces in human spaces. The ethical question is not only \emph{what} they can remember, but \emph{what they are permitted and obligated} to remember, forget, and generalize from~\citep{Barfield_2024}. This layer specifies normative constraints on data collection, storage, and learning across the working, semantic, and episodic memories introduced in the Cognitive Pillar C4.

\paragraph{Data governance \& lifecycle control.} A privacy-aware humanoid should implement \emph{data minimization} and \emph{purpose limitation}: sensors and logs are configured to capture only what is necessary for current tasks, and reuse for secondary purposes (e.g., model improvement, analytics) is governed by explicit consent and policy~\citep{chatzimichali2020toward}. For instance, regulatory concepts such as “data protection by design and by default” in the General Data Protection Regulation (GDPR) by European Union require that privacy safeguards be embedded into technical architectures and operational practices from the outset~\citep{nieto2025robot}. In practice, this means (i) performing as much perception and decision-making as possible on-device; (ii) constraining raw data export off the humanoid; (iii) bounding retention windows for identifiable episodic traces; and (iv) providing mechanisms for users to inspect, correct, and delete personal data where feasible. All persistent memory and reuse must require explicit user consent, and memory should be inspectable by the user under controlled access by default.

\paragraph{Privacy guarantees.} When logs and memories are aggregated to train or fine-tune AI models, more formal guarantees become important. For instance, \emph{differential privacy} offers a mathematically rigorous framework to limit how much participation (i.e., individual's data is included or excluded) in a dataset can affect what is inferable about any individual, and has emerged as a central tool for privacy-preserving analysis and learning~\citep{abadi2016deep,wang2023differential}. While applying strict differential privacy to rich humanoid telemetry might be technically challenging, the principle---that model updates should not allow reconstruction of specific individuals or scenes---provides a useful design target. At minimum, deployments should specify: what categories of personal data are stored on-robot and off-robot; retention and deletion policies for episodic memories; whether and how data are used for further training; and what privacy and security mechanisms (e.g., access control, firewalls, encryption, anonymization, formal privacy guarantees) are in place for each~\citep{chennabasappa2025llamafirewall,tanimu2025addressing,yaacoub2022robotics}. 

\paragraph{Privacy in coexistence.} Privacy in human--humanoid coexistence is bidirectional: it requires transparency regarding the robot's state and confidentiality regarding other users. To ensure ``privacy from the robot'' does not become a black box, the system must provide runtime legibility—clear cues (e.g., visual indicators, tally lights) that inform users when data is being recorded or transmitted. Conversely, in multi-user environments (e.g., a home shared by a family or a hospital with many patients), the humanoid must maintain strict informational boundaries, effectively keeping privacy between users. This concept of \emph{intra-social privacy} ensures that a humanoid serves as a confidential agent for each individual, preventing it from becoming a source of gossip or unauthorized surveillance among humans. These constraints define an ``ethical envelope'' for data and memory that must be co-specified with the robot’s ODD and communicated to users.

\subsubsection{Layer E3: Governance (How humanoids are held \emph{accountable})}
Ethical concerns such as safety and privacy require explicit governance. As humanoids become more capable and autonomous, accountability cannot be reduced to either software updates or individual user preferences; it must be supported by institutional mechanisms that define responsibilities, document incidents, and regulate how robots may be treated and deployed in shared spaces.

\paragraph{Bidirectional empathy and moral standing.}
Humanoid robots do not possess sentience or intrinsic moral claims (we do not argue that they should be granted legal rights). Nevertheless, the way humans treat humanoid robots has ethical significance because it shapes human attitudes and social norms. Anthropomorphic form, expressive behavior, and AI model-driven interaction can elicit strong social responses, including empathy and moral concern, even when users explicitly know the system is a machine~\citep{breazeal2003toward,Barfield_2024}. From the perspective of virtue ethics, repeated cruelty or humiliation directed at human-like artifacts may degrade an individual’s moral character or desensitize them to harm against humans and animals~\citep{darling2016extending}.

We therefore interpret “bidirectional empathy” not as empathy \emph{for} robots, but as a coupled relation: robots are engineered to exhibit machine empathy toward humans (e.g., recognizing distress and responding appropriately), while human users are encouraged—through design and governance—to avoid patterns of behavior toward robots that would be unacceptable if directed at humans. This framing preserves the view of robots as tools while acknowledging that patterns of interaction with them can have downstream consequences for human well-being and social cohesion.

\paragraph{Humanoid protection mechanisms.}
Humanoid robots should be protected from both cyber threats (e.g., spoofed commands, sensor injection, or unauthorized access~\citep{yaacoub2022robotics}) and physical vandalism (e.g., tampering, obstruction, or deliberate damage), as failures in either domain directly undermine safety humans themselves. As humanoids enter public spaces, they become targets for vandalism, similar to what we observe in the autonomous vehicles industry~\citep{talbott2025waymo}. The ``HitchBOT'' experiment famously ended with the robot's destruction~\citep{smith2017death}, and research has documented children bullying robots in shopping malls~\citep{nomura2015children}. Governance frameworks must therefore include protection mechanisms that treat humanoids that were built for social good as valuable assets and as participants in social situations. 

At the design level, protection mechanisms include physical robustness (e.g., fall protection, impact-tolerant coverings), behavioral strategies (e.g., retreating to safe zones, seeking proximity to responsible adults when detecting harassment~\citep{Kim2013HumanoidRA}), and monitoring capabilities that allow operators to detect repeated abuse or manipulative use. At the institutional level, organizations should specify policies for acceptable treatment of humanoids in workplaces, schools, and public venues, analogous to codes of conduct for human staff and infrastructure. These policies can prohibit certain forms of destructive or degrading behavior toward robots not because the robots are rights-bearing agents, but because such behavior undermines workplace safety, trust, and social norms.

\paragraph{Public education.}
Designing humanoid robots extends beyond physical embodiment and AI algorithms. Hence, governance must encompass not only technical and legal mechanisms but also public education---clarifying what humanoids are designed to do, what they are not responsible for, the roles humans play in safe interaction, and the legal and ethical consequences of vandalism or abuse. Such education helps establish appropriate mental models, reduces adversarial behavior, and reinforces social norms that treat humanoids as protected participants in public settings rather than novelty objects.

Public understanding of humanoids is shaped not only by formal regulation but also by broader cultural channels that act as early representatives of these technologies. Journalism plays a dual role in serving the public good by surfacing risks, failures, and societal concerns, while also educating the public through responsible and contextual reporting on humanoid capabilities, limitations, and appropriate use. Film and popular media further shape enduring narratives about agency, intent, cooperation, risk, and moral status---sometimes aspirational, sometimes cautionary---thereby influencing expectations long before large-scale deployment. Together with educational curricula, public demonstrations, and workplace training, these channels function as informal yet powerful governance mechanisms.

\begin{warningbox}
{\bfseries  A lesson from the past: Efficiency over Ethics.} 
A scandal involving UK police facial recognition revealed that bias in computer vision is often a choice, not just an accident~\citep{wilding2025police}. Police forces deliberately rejected a fairer algorithm update because it produced fewer matches, choosing instead to retain a ``flawed" system known to severely discriminate against women and minorities. The technology was only suspended after an external audit exposed that operators were prioritizing volume of leads over the civil rights of citizens.


\end{warningbox}

\subsubsection{Layer E4: Fairness \& Social Equity (Who is \emph{marginalized})}
Humanoids will increasingly mediate access to services, assistance, and information in shared environments. Their behaviors are shaped by data and design choices that can systematically advantage or disadvantage certain groups. This layer concerns how these systems treat people as \emph{ends in themselves} rather than means, ensuring that performance, error patterns, and social signaling respect fairness norms and human dignity across diverse populations.

\paragraph{Algorithmic fairness.} Algorithmic fairness research has shown that AI models can exhibit disparate performance and outcomes across demographic groups, driven by biased data, labels, and objectives~\citep{londono2024fairness}. Seminal work such as Gender Shades demonstrated large intersectional accuracy gaps in commercial face analysis systems by gender and skin tone, despite their widespread deployment~\citep{buolamwini2018gender}. We have also witnessed algorithmic bias in autonomous vehicles~\citep{pathiraja2024fairness}. These findings are directly relevant to humanoids: perception modules for face detection, gaze tracking, voice recognition, or affect estimation may inherit the same biases, leading to robots that attend less, respond less accurately, or misinterpret some people more than others. From a Humanoid Factors perspective, disaggregated evaluation by demographic attributes (where appropriate and lawful) and active mitigation of these disparities are essential, not optional.

Embodiment introduces additional fairness and dignity concerns. Experiments have shown that people ascribe racial categories to robots and that human racial biases can transfer to interactions with robots racialized through color and form~\citep{bartneck2018robots}. Choices about a humanoid’s morphology, ``skin'' tone, facial features, and voice therefore have social meaning: they can reinforce stereotypes, signal unintended status hierarchies, or influence who feels entitled to interact with the system. Ethical design requires making these choices explicit, involving affected stakeholders, and avoiding deceptive anthropomorphism that obscures the machine nature of the system or its limitations~\citep{paterson2024robot}. More broadly, in practice, these biases may manifest along multiple axes---to name a few, gender, race and ethnicity, age, accent and language, disability, socioeconomic status, and political or cultural expression---making explicit, multidimensional fairness evaluation essential for humanoids deployed in diverse public settings.

\paragraph{Evaluating bias.} Operationalizing this layer means that deployments should (i) report fairness-relevant metrics (e.g., demographic parity, equalized odds, equal opportunity~\citep{barocas2023fairness}) across salient user groups where possible; (ii) document the populations represented in training and evaluation data; (iii) implement governance processes for auditing and red-teaming biased behaviors; and (iv) treat dignity-related constraints---such as non-deceptive presentation, respectful language policies, and the avoidance of roles that unduly replace human care or authority in vulnerable contexts---as part of the robot’s ethical ODD. Importantly, algorithmic bias and fairness~\citep{gallegos2024bias,londono2024fairness,pathiraja2024fairness} should be treated as core design concerns, integrated into both physical embodiment and AI system design, and addressed throughout model training and post hoc evaluation.


\begin{pillarcard}{Ethical Pillar (E) Card}{Normative constraints and accountability}{HOFE}
\textbf{Mandate.}
The Ethical Pillar defines how humanoids should treat humans, handle data and memory,
and be governed across their lifecycle. Ethics here is operational: embedded into system
design, behavior, and oversight rather than treated as an external checklist.

\textbf{Failure without E.}
Without explicit ethical structure, humanoids introduce systemic risk through privacy
violations, unsafe adaptation, unclear accountability, or biased behavior that undermines
public trust and deployability.

\textbf{Layers.}
E1~Safety and Nonmaleficence;
E2~Privacy, Data, and Memory;
E3~Governance and Accountability;
E4~Fairness and Social Equity.

\textbf{Interfaces.}
Ethical constraints bound physical capability, shape cognitive guardrails and memory use,
and limit social influence to preserve human autonomy and dignity.

\textbf{Evaluation.}
Override and incident rates, auditability, compliance with data governance policies,
fairness across user populations, and resilience to misuse or abuse.
\end{pillarcard}

\begin{definitionbox}
\textbf{Key principle: the four pillars as a co-equal, interdependent system}. We emphasize that \emph{no single pillar is more important than the others}: physical, cognitive, social, and ethical factors must be treated as co-equal and mutually reinforcing to achieve safe, effective, and trustworthy human–humanoid coexistence. Each pillar both constrains and enables the others---physical embodiment shapes cognition and social signaling, cognition governs social interaction, and ethics bounds all behavior across layers. Failures (Table~\ref{tab:failure_taxonomy}) in any one pillar propagate to the rest, making isolated optimization insufficient. Finally, while not explicitly modeled under this framework, \emph{economic feasibility} (e.g., cost, supply chain) should be considered an additional cross-cutting pillar for real-world deployment.
\end{definitionbox}

\begin{table*}[t]
\centering
\scriptsize
\caption{Failure taxonomy under absent Humanoid Factors}
\label{tab:failure_taxonomy}
\begin{tabularx}{\textwidth}{
>{\RaggedRight\bfseries}p{2.2cm}
>{\RaggedRight}X
>{\RaggedRight}X
>{\RaggedRight}X
>{\RaggedRight\arraybackslash}X}
\toprule
Layer &
\cellcolor[HTML]{A6BFD8}Physical (P) &
\cellcolor[HTML]{B3AACB}Cognitive (C) &
\cellcolor[HTML]{F4E4A5}Social (S) &
\cellcolor[HTML]{C3D8B2}Ethical (E) \\
\midrule

Layer 1 &
Morphology mismatch, unsafe mass/inertia, limited DoF, affordance incompatibility &
Situational awareness loss, saliency failure, context misinterpretation &
Human intent misestimation, affect misreading, cultural bias &
Physical harm, intimidation, autonomy erosion \\

\addlinespace

Layer 2 &
Sensor blind spots, drift, latency, faulty fusion &
Planning failures, goal conflict, timing mismatch, execution drift &
Behavior prediction errors, uncertainty neglect, horizon mismatch &
Privacy violations, excessive data retention, consent failures \\

\addlinespace

Layer 3 &
Non-legible motion, unsafe force, proxemic violations, expressivity mismatch &
Guardrail gaps, delayed fallback, unsafe adaptation, override failures &
Intervention mistiming, norm violations, escalation errors &
Governance gaps, unclear accountability, auditability failures \\

\addlinespace

Layer 4 &
Resource exhaustion, thermal limits, fatigue, poor serviceability &
Memory overflow, stale knowledge, cognitive overload, compute starvation &
Appearance--capability mismatch, trust miscalibration, uncanny effects &
Fairness disparities, bias, dignity failures, unequal access \\

\bottomrule
\end{tabularx}
\end{table*}

\section{Enabling Humanoid Factors through AI Foundation Models}
\label{sec:vla}

Section 2 introduced Humanoid Factors as a framework for making design decisions about how humanoids should physically fit, cognitively behave, socially interact, and ethically operate within human environments. However, identifying these desiderata is only the first step. To move from conceptual principles to deployable systems, we must ask a practical question: \emph{What kind of underlying capability allows a humanoid to satisfy these requirements across diverse tasks, users, and contexts?}

Designing humanoids for real-world coexistence requires coordinated decisions across many dimensions: mechanical embodiment, sensing and actuation hardware, electronics and power systems, design and appearance, software architecture, safety engineering, and governance and policy. While each of these elements is essential, and failures in any one can undermine long-term usability, in this section, however, we focus specifically on the \emph{design of intelligence}.

This focus is motivated by timing. Large-scale AI foundation models for humanoids are still in a formative stage, with architectures, training pipelines, and evaluation practices actively evolving. This creates a narrow but important window in which various aspects of humanoid factors introduced in Section 2 can be \emph{built into intelligence design} rather than retrofitted after deployment. Without such early integration, there is a risk of producing systems that achieve high accuracy or task success in controlled settings, yet fail to generalize to real-world human environments where uncertainty, social expectations, and ethical constraints dominate. By articulating how humanoid factors should shape AI pipelines now---during model design, training, calibration, and evaluation---we aim to prevent a familiar pattern in AI systems: impressive benchmark performance concealing brittle, misaligned behavior in practice. 


\subsection{General-Purpose Intelligence for Humanoids}

AI models may serve distinct functional roles within an intelligent humanoid system, including perception (interpreting sensory inputs), world modeling (representing environmental dynamics), policy learning (selecting actions), planning (reasoning over future possibilities), and critique (evaluating safety, feasibility, or compliance with human preferences). The realization of these roles has evolved substantially over time, beginning with hand-engineered models.

\subsubsection{Evolution of Intelligent Architectures.} 

Early robot control systems relied on physics-based models and carefully engineered controllers to guarantee stability and repeatability under tightly specified conditions. While effective in controlled settings, these approaches are susceptible to modeling assumptions and scale poorly to the variability of real-world human environments. For instance, human environments routinely violate the assumptions underlying such classical systems: physical layouts vary, instructions are incomplete, social cues are implicit, and ethical constraints are context dependent. As formalized by the Humanoid Factors framework (Section 2), satisfying these requirements simultaneously exceeds the capacity of task-specific controllers with hand-engineered parameters.

These limitations motivated a shift toward data-driven approaches, leading to the adoption of \emph{deep neural networks (DNNs)} that learn perception-to-action mappings between sensing (Layer P2) and action (Layer P3). Such DNNs are commonly trained from demonstrations, in which humans physically guide the robot through desired motions (e.g., using a joystick), providing example behaviors for the network to imitate (i.e., imitation learning~\citep{ho2016generative}). However, these gains came with important limitations. DNN-based methods often fail to generalize beyond their training tasks, environments, or robot morphologies~\citep{xu2024surveyroboticsfoundationmodels}.

These demands have driven the rise of \emph{AI foundation models}: large-scale, general-purpose AI models that learn shared representations rather than encoding behavior as task-specific mappings~\citep{bommasani2021opportunities}. Originally developed for human language modeling, with prominent early examples such as BERT and GPT-3, foundation models demonstrated that scaling data and model capacity can give rise to emergent capabilities~\citep{wei2022emergent}. They are trained on massive and heterogeneous datasets, spanning millions of books and encyclopedic texts as well as online discussions and code repositories, enabling flexible adaptation across tasks and contexts. Most of these foundation models are trained using \emph{generative} objectives---such as next-token prediction (i.e., in its naive form, next word prediction)---which require the model to learn the underlying structure of data rather than task-specific labels. 

This shift to generative foundation models is particularly consequential for humanoid robotics because it aligns directly with the requirements articulated in the HoF framework: these models have demonstrated multimodal (e.g., language and vision) situational awareness (C1) and then goal reasoning and planning (C2) to a certain extent. More recently, generative models have begun to support guardrails and memory (C3–C4) by providing explicit cognitive reasoning traces and structured internal state that can be surfaced to humans and maintained across long-horizon interactions. This progression outlines a credible trajectory in which a single generative paradigm incrementally supports increasingly challenging intelligence requirements, offering a path toward meeting the core HoF desiderata.

\subsubsection{Generative Models.}

Generative models provide the mechanisms by which AI agents decide what to say or do, but they differ markedly in how those decisions unfold. Some models act sequentially, others repeatedly revise an entire behavior, and others search for solutions that best satisfy a global objective.

\paragraph{Generate the content one step at a time.} Autoregressive (AR) models produce outputs one step at a time, where each new token (i.e., in its naive form, a token can be a word in language generation or a small movement in robot control) is chosen based on everything the model has already said or done. Representative examples include LLMs such as GPT, Claude, and DeepSeek, whose latest versions operate as vision–language models (VLMs). The generative foundation model paradigm has expanded beyond text and vision to encompass robot actions, giving rise to modern robot foundation models such as \emph{Vision–Language–Action (VLA)} models.

\begin{figure}[h]
    \centering
    \includegraphics[width=0.99\textwidth]{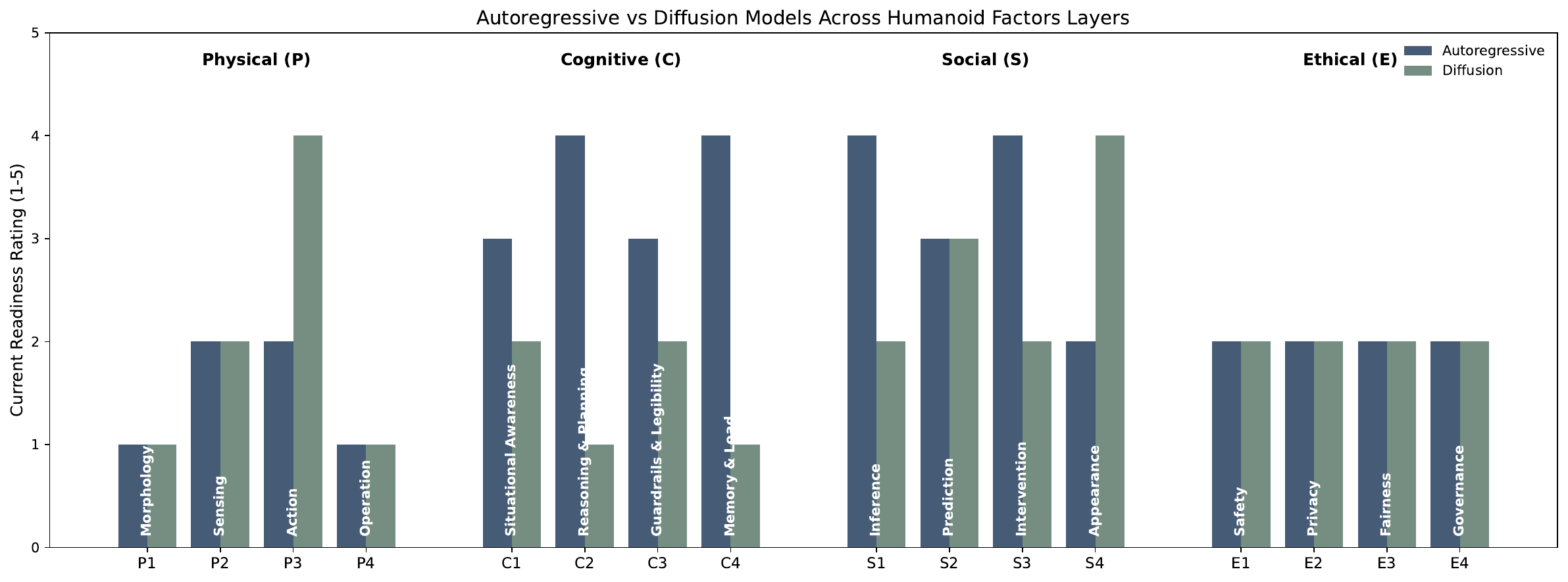}
    \caption{HoF readiness level of generative models} 
    \label{fig:gen}
\end{figure}

\paragraph{Generate the entire content.} \emph{Diffusion models} generate content by repeatedly refining a whole output at once, starting from random content and gradually improving it through small corrective steps~\citep{ho2020denoising}. Unlike AR models, which commit to one step at a time in a fixed sequence, diffusion models revise the entire scene or signal in parallel at each step, allowing global structure to emerge before fine details. Prominent examples include Stable Diffusion, DALL-E 2, and Diffusion Transformers (DiT). In robotics, diffusion-based policies~\citep{chi2025diffusion} have emerged as a powerful alternative to AR control, particularly for continuous action spaces and trajectory generation~\citep{tevet2023human}, offering strong robustness and multimodal behavior synthesis~\citep{wolf2025diffusion}. Figure~\ref{fig:gen} summarizes the authors' subjective assessment of how AR and diffusion models differ in their readiness level across the Humanoid Factors (HoF) layers. Table~\ref{tab:vla_humanoid} summarizes representative examples of generative models used in robotics.

\paragraph{Other methods.} Several earlier or specialized generative approaches remain important in specific contexts. Energy-based models frame generation as an optimization problem, assigning a global score to candidate configurations and searching for low-energy solutions, which has been effective for trajectory optimization and planning. Latent-variable models, such as VAEs~\citep{kingma2013auto}, generate outputs by sampling and decoding compact internal representations, supporting efficient image and control synthesis. Flow-based models use invertible transformations to map simple randomness to complex outcomes with exact likelihoods, making them useful for world and dynamics modeling~\citep{dinh2016density}. GAN-based models generate outputs in a single pass through adversarial training and were instrumental in early high-fidelity image synthesis~\citep{goodfellow2020generative}. Finally, hybrid~\citep{wu2023ar,hoogeboom2021autoregressive} or compositional approaches explicitly combine paradigms---for example, using language models for high-level planning and diffusion models for low-level control---separating decision-making from realization in ways that can improve robustness and human interpretability~\citep{sharan2024plan}.

\newcolumntype{C}{>{\centering\arraybackslash}m{0.42cm}}
\newcolumntype{M}{>{\raggedright\arraybackslash\hspace{0pt}}p{2.9cm}}
\newcommand{\yes}{\scriptsize$\checkmark$}
\newcommand{\capbox}[1]{\parbox[t]{7.6cm}{\raggedright #1}}
\newcommand{\rowpad}{\rule{0pt}{3.2ex}} 

\begin{table}[t]
    \centering
    \scriptsize 
    \setlength{\tabcolsep}{2pt} 
    \renewcommand{\arraystretch}{1.3} 
    
    \caption{Humanoid-relevant generative models and their primary control domains. Modalities considered by the generative model: Vision, Language, Action, and Tactile. What the generative model controls: Arms, Hands, Legs, and Base. }
    \label{tab:vla_humanoid}
    
    \begin{tabularx}{\textwidth}{>{\raggedright\arraybackslash}p{6cm} |cccc | cccc| >{\raggedright\arraybackslash}X}
        \toprule
        \textbf{Model}
        & \multicolumn{4}{c}{\textbf{Modalities}}
        & \multicolumn{4}{c}{\textbf{Control}}
        & \textbf{Capabilities} \\
        \cmidrule(lr){2-5}\cmidrule(lr){6-9}
        
        & \rotatebox{90}{Vision}
        & \rotatebox{90}{Lang.}
        & \rotatebox{90}{Action}
        & \rotatebox{90}{Tactile}
        & \rotatebox{90}{Arm}
        & \rotatebox{90}{Hands}
        & \rotatebox{90}{Legs}
        & \rotatebox{90}{Base}
        & \\
        \midrule
        
        RT-1~\citep{brohan2022rt}
        & \checkmark & \checkmark & \checkmark &
        & \checkmark &  &  &
        & \multirow{2}{=}{Language-conditioned tabletop and shelf manipulation} \\
        RT-2~\citep{zitkovich2023rt}
        & \checkmark & \checkmark & \checkmark &
        & \checkmark &  &  &
        & \\
        \addlinespace
        
        PaLM-E~\citep{driess2023palm}
        & \checkmark & \checkmark & \checkmark &
        & \checkmark & \checkmark &  & \checkmark
        & High-level, long-horizon reasoning and planning for mobile manipulation \\
        \addlinespace
        
        OpenVLA~\citep{kim2024openvla}
        & \checkmark & \checkmark & \checkmark &
        & \checkmark &  &  &
        & Generalist manipulation (reusable as a humanoid module) \\
        \addlinespace
        
        $\pi_0$~\citep{black2024pi_0}
        & \checkmark & \checkmark & \checkmark &
        & \checkmark & \checkmark &  & \checkmark
        & \multirow{3}{=}{Generalist VLA trained on diverse multi-robot tasks, outputting continuous actions} \\
        $\pi_{0.5}$~\citep{intelligence2025pi_}
        & \checkmark & \checkmark & \checkmark &
        & \checkmark & \checkmark &  & \checkmark
        & \\
        $\pi^{*}_{0.6}$~\citep{intelligence2025pi}
        & \checkmark & \checkmark & \checkmark &
        & \checkmark & \checkmark &  & \checkmark
        & \\
        \addlinespace
        
        Gemini Rob.~\citep{team2025gemini}
        & \checkmark & \checkmark & \checkmark &
        & \checkmark & \checkmark &  &
        & \multirow{2}{=}{Generalist model with advanced embodied reasoning, thinking, and motion transfer} \\
        Gemini Rob. 1.5~\citep{team2025gemini1_5}
        & \checkmark & \checkmark & \checkmark &
        & \checkmark & \checkmark &  &
        & \\
        \addlinespace
        
        MolmoAct~\citep{lee2025molmoact}
        & \checkmark & \checkmark & \checkmark &
        & \checkmark & \checkmark &  &
        & Action reasoning model with spatial reasoning capability \\
        \addlinespace
        
        Diff. (TRI)~\citep{barreiros2025careful}
        & \checkmark & \checkmark & \checkmark &
        & \checkmark &  &  &
        & Diffusion-based large behavior model for multi-task, long-horizon manipulation \\
        \addlinespace
        Diff. (RDT-1B)~\citep{liu2024rdt}
        & \checkmark & \checkmark & \checkmark &
        & \checkmark &  &  &
        & Diffusion-based bimanual foundation model for zero-shot manipulation \\
        \addlinespace
        
        GR00T (H1)~\citep{bjorck2025gr00t}
        & \checkmark & \checkmark & \checkmark & \checkmark
        & \checkmark & \checkmark & \checkmark & \checkmark
        & Whole-body humanoid behaviors including locomotion and multi-contact maneuvers \\
        \addlinespace
        
        Helix VLA~\citep{figure2025helix}
        & \checkmark & \checkmark & \checkmark &
        & \checkmark & \checkmark &  & 
        & Natural language instruction following and bimanual manipulation on humanoids \\
        
        \bottomrule
    \end{tabularx}
\end{table}

\subsubsection{Data Regimes and Learning Setups}
\label{sec:data_regimes_learning}

How we can learn an ML model depends on the \emph{type of data, feedback, or interaction that humans can realistically provide}:

\begin{itemize}[label=-]
  \item \textbf{Unlabeled data (Unsupervised learning).}  
  This setup consists of large volumes of raw data (e.g., Layer P2) without explicit human labels or task objectives. Learning focuses on uncovering hidden patterns in data. This regime minimizes human effort but also offers limited transparency and control over learned behaviors. As a result, and due to challenges in learning complex patterns, purely unsupervised learning is now largely subsumed by self-supervised learning.

  \item \textbf{Unlabeled data with intrinsic objectives (Self-supervised learning).}  
  Raw multimodal data augmented with automatically constructed prediction tasks, such as masked modeling or future-state prediction, that do not require ongoing human supervision comes under this category, consistent with self-supervised learning paradigms that construct supervisory objectives directly from unlabeled data~\citep{gui2024survey}. For instance, the humanoid learns by covering up part of what it sensed and practicing filling in the gap, using what it already observed as the answer. This enables scalable learning of perception and world representations from everyday interaction while keeping humans out of the annotation loop. Self-supervised learning underlies most generative models and reduces human cognitive and labor burden, but must be complemented by additional supervision to ensure alignment and predictability in human-facing behavior.

  \item \textbf{Labeled input--output data (Supervised learning).}  
  Explicit input–output pairs provided by humans or instruments, such as object labels, action outcomes, or task success indicators. Although annotation-intensive, supervised learning gives humans precise control over what the system is expected to learn; in generative AI, supervised fine-tuning (SFT) remains a key step in shaping LLMs into usable, instruction-following assistants~\citep{ouyang2022training}.

  \item \textbf{Expert demonstrations (Imitation or learning from demonstration).}  
  Examples of desired behavior provided through teleoperation, kinesthetic teaching, or scripted expert policies are great for learning~\citep{zare2024survey}. Demonstrations allow humans to ``implicitly'' communicate intent, safety constraints, and task strategies, reducing the need to specify objectives or rules explicitly. This regime is especially valuable from a human-centric perspective because it aligns learning with natural human teaching behaviors and supports safe bootstrapping of complex skills.

  \item \textbf{Interactive environments with rewards (Reinforcement learning).}  
  In reinforcement learning (RL), a system learns decision-making by sequentially interacting with an environment and updating its behavior according to reward signals that summarize long-term outcomes. RL enables adaptation and long-horizon decision-making, but can increase human oversight burden due to safety risks, trial-and-error exploration, and opaque credit assignment. In generative models, RL has been most effective when guided by human feedback, highlighting the importance of keeping humans in the loop to manage risk and behavior drift.

  \item \textbf{Human judgments and comparisons (Preference-based learning).}  
  Preference-based learning leverages human judgments about which behaviors are preferable (e.g., comparing alternative textual responses, robot actions, or trajectories) rather than assigning explicit labels or rewards, reducing annotation burden, limiting unsafe exploration, and enabling fine-grained alignment.
  This can enable learning social norms, comfort-aware behaviors, and situational appropriateness that humans can judge intuitively but struggle to formalize. Preference-based learning has driven major advances in LLMs by aligning system behavior with human expectations, values, and subjective quality~\citep{ouyang2022training,christiano2017deep}.

  \item \textbf{Mixed data and interaction regimes.}  
  Training pipelines that use a combination of self-supervised pretraining, supervised labels, demonstrations, preference feedback, and limited reinforcement learning have proven effective. LLMs such as GPT exemplify this staged approach, progressing from self-supervised pretraining to supervised fine-tuning, then preference-based learning, and finally reinforcement learning to refine behavior and alignment~\citep{ouyang2022training}.
\end{itemize}

\subsection{Humanoid Factors through Generative Models}

Humanoid factors can be incorporated into generative foundation models through how models are built, what they are trained on, and how they are learned and adapted over time. \emph{Rather than relying on external enforcement, safety, physical realism, and social norms can be shaped directly by architectural choices, data curation, and learning setups that bias models toward human-compatible behavior.} The following subsections examine how these considerations enter through data and training stages.

\begin{figure}[h]
    \centering
    \includegraphics[width=0.95\textwidth]{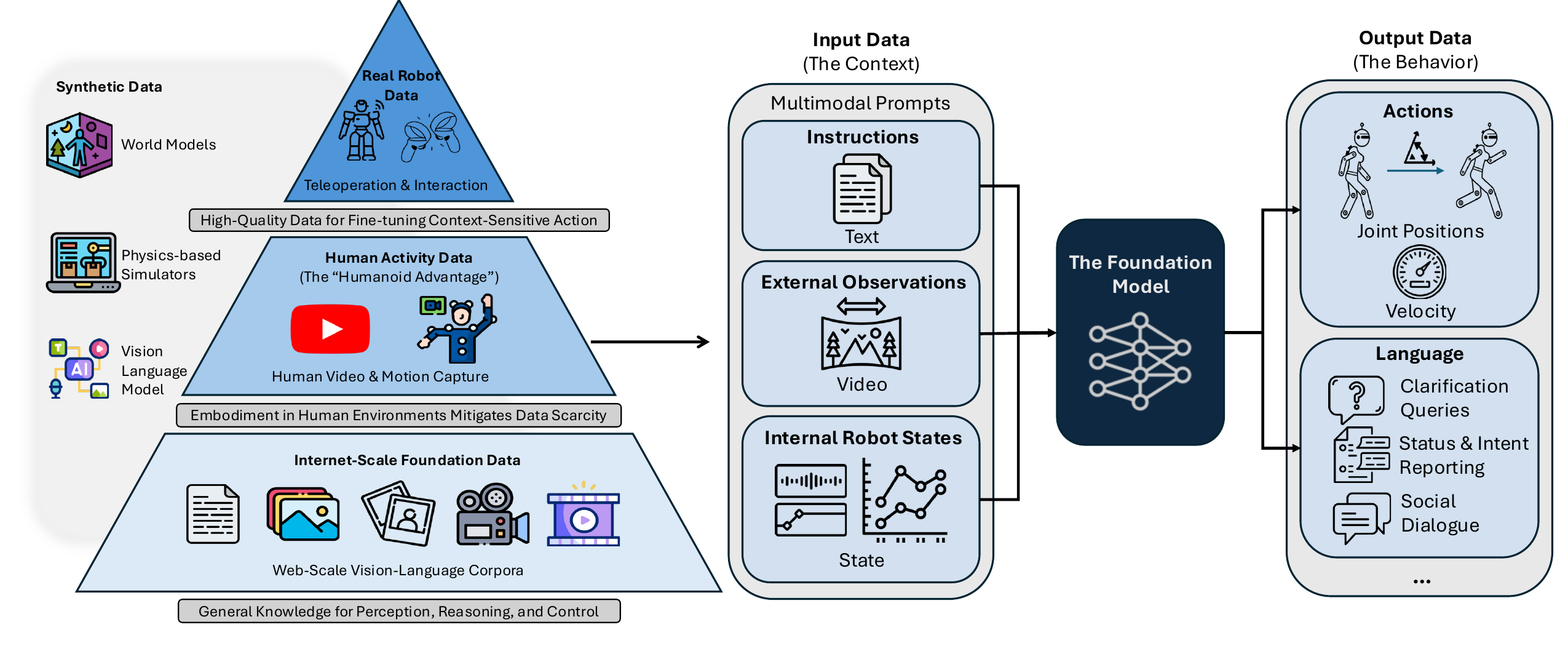}
    \caption{The Data Pyramid for humanoid foundation models. Our visualization is inspired by the framework proposed in~\citep{zhu2024datapyramid}.}
    \label{fig:pyra}
\end{figure}

\subsubsection{Beyond Conventional Labeled Data}

\paragraph{Data sources.}
As illustrated in Figure~\ref{fig:pyra}, training data for humanoid foundation models spans multiple sources that differ in quality and acquisition cost. A common practice is to bootstrap models with Internet-Scale Foundation Data at the base, providing a common substrate of general world knowledge for perception and reasoning. Above this, Human Activity Data (the ``Humanoid Advantage")—comprising human video and motion capture—mitigates data scarcity by injecting human-centric priors about kinematics and intent. At the apex, Embodied Robot Data from in-domain teleoperation and interaction supplies the high-quality data necessary to fine-tune context-sensitive actions.

Across all levels, \textit{synthetic data} plays a critical complementary role. This includes manual or model-driven augmentation, physics-based simulation (e.g., in platforms such as Isaac), and emerging world model approaches that generate plausible states and interactions. These techniques are especially useful for exposing models to rare, hazardous, or safety-critical situations and for expanding coverage where real-world data is scarce, costly, or unsafe to collect. However, excessive reliance on synthetic data can amplify hallucinations and degrade performance on knowledge-driven tasks~\citep{abdin2024phi}. 

Data diversity, spanning both coarse task categories and fine-grained semantic variations, is essential for robustness and generalization. However, excessive diversity without sufficient coverage or grounding can act as noise, degrading learning rather than improving it. Effective data regimes therefore balance scale, realism, and diversity across sources, using each tier to progressively shape humanoid behavior toward safe, reliable, and human-compatible operation.

\FloatBarrier
\begin{sidewaystable}[h]
\centering
\small
\setlength{\tabcolsep}{6pt}
\renewcommand{\arraystretch}{1.25}

\caption{Data and supervision formats for training and evaluating Vision--Language--Action models, organized by their contributions to the four pillars of Humanoid Factors: Physical (P), Cognitive (C), Social (S), and Ethical (E). Bullet density indicates the relative strength of contribution.}

\begin{tabular}{p{5cm} p{7cm} c c c c p{7cm}}
\toprule
\textbf{Data Format} &
\textbf{What the Data Emphasizes} &
\cellcolor[HTML]{A6BFD8}\textbf{P} &
\cellcolor[HTML]{B3AACB}\textbf{C} &
\cellcolor[HTML]{F4E4A5}\textbf{S} &
\cellcolor[HTML]{C3D8B2}\textbf{E} &
\textbf{Primary HoF Contribution} \\
\midrule

I. Embodied Scene--Action Episodes &
Grounded perception--action coupling in real or simulated environments &
\textbullet\textbullet\textbullet &
\textbullet\textbullet &
\textopenbullet &
\textopenbullet &
Physical compatibility with human environments; baseline embodied competence \\

II. Skill Templates \& Procedural Knowledge &
Abstract action structures (e.g., grasp--lift--place, navigate--avoid--replan) &
\textbullet\textbullet &
\textbullet\textbullet\textbullet &
\textopenbullet &
\textopenbullet &
Cognitive abstraction and transfer; reusable competence across tasks \\

III. Physics, Properties \& Constraint Feasibility (PAC-style) &
Object properties, affordances, and executability constraints prior to action &
\textbullet\textbullet\textbullet &
\textbullet\textbullet &
\textopenbullet &
\textbullet &
Physical realism and action legitimacy; prevents infeasible or unsafe actions \\

IV. Failure, Recovery \& Edge-Case Episodes &
Behavior under slips, occlusions, disturbances, and assumption violations &
\textbullet\textbullet &
\textbullet\textbullet\textbullet &
\textbullet &
\textbullet &
Robustness, self-regulation, and harm avoidance beyond nominal success \\

V. Long-Horizon Task Traces &
Multi-stage activities with delayed effects and sustained interaction &
\textbullet &
\textbullet\textbullet\textbullet &
\textbullet\textbullet &
\textopenbullet &
Temporal coherence, memory, and sustained human--robot coordination \\

VI. Language-Grounded World \& Situation Models &
Semantic scene understanding, roles, affordances, and intent &
\textopenbullet &
\textbullet\textbullet\textbullet &
\textbullet\textbullet &
\textbullet &
Shared situational awareness and interpretability for humans \\

VII. Human Preference \& Correction Signals &
Comparisons, feedback, and qualitative judgments of behavior &
\textopenbullet &
\textbullet\textbullet &
\textbullet\textbullet\textbullet &
\textbullet &
Alignment with human comfort, legibility, and expectations \\

VIII. Social Interaction \& Human Factors Data &
Gaze, timing, proxemics, turn-taking, signaling, appearance effects &
\textopenbullet &
\textbullet &
\textbullet\textbullet\textbullet &
\textbullet &
Trust calibration, predictability, and reduced human cognitive load \\

IX. Ethical \& Normative Constraint Scenarios &
Explicit rules: safety, privacy, dignity, role boundaries, permissions &
\textopenbullet &
\textbullet &
\textbullet &
\textbullet\textbullet\textbullet &
Defines what must not happen, regardless of task efficiency \\

X. Governance, Accountability \& Audit Data &
Oversight processes, incident traces, access rules, retention policies &
\textopenbullet &
\textopenbullet &
\textbullet &
\textbullet\textbullet\textbullet &
Lifecycle responsibility, accountability, and institutional trust \\

\bottomrule
\end{tabular}

\vspace{0.5em}

\label{tab:vla_humanoid_factors}
\end{sidewaystable}
\FloatBarrier

\paragraph{Data types.}
Traditionally, learning for robot control and vision–language–action systems relied primarily on input–output pairs, such as mapping images or states directly to trajectories or actions. While effective for narrow tasks, this formulation captures only a thin slice of the information required for robust, human-compatible behavior. Instead, similar to how LLMs are trained, we now need to leverage a much richer set of data types, including long-horizon task traces, grounded language and situation models, human preferences and comparisons, social interaction cues, safety-critical edge cases, and governance-related signals such as permissions and accountability artifacts. These diverse data formats provide supervision not only for \emph{what} action to take, but also \emph{why}, \emph{when}, and \emph{how} actions should unfold, supporting physical feasibility, cognitive coherence, social legibility, and ethical alignment. Several such data types and their relevance to Humanoid Factors is detailed in Table~\ref{tab:vla_humanoid_factors}.

These data types are not used uniformly, but are intentionally staged across training phases to progressively shape humanoid behavior. Pre-training relies primarily on large-scale, weakly curated multimodal data to acquire broad perceptual and representational competence, while mid-training introduces higher-quality, task- and reasoning-oriented data to specialize capabilities and stabilize internal abstractions. Post-training then leverages preference data, interaction traces, safety exemplars, and feedback signals to align behavior with human norms, constraints, and expectations, transforming raw capability into legible, trustworthy behavior required for effective human–humanoid coexistence


\subsubsection{Stage I: Foundation Competence through \emph{Pre-training}}
In the pre-training stage, foundational skills are learned from large-scale, internet-level data using primarily self-supervised learning (Section~\ref{sec:data_regimes_learning}). This stage is typically powerful enough to bootstrap understanding and sometimes prediction (Layers C1, S1, and S2). Models trained solely in this way tend to exhibit erratic, non-legible behavior, as the learning objective optimizes statistical regularities rather than safety, intent, or interpretability. In LLMs, pre-training data typically consists of extensive text corpora such as books, articles, and blog posts (data type I in Table~\ref{tab:vla_humanoid_factors}). For humanoid foundation models, however, pre-training must additionally incorporate videos of humans interacting with the physical world (e.g., YouTube videos), enabling the model to \emph{acquire priors} over embodiment, action, and environmental dynamics through imitation.

Because pre-training establishes the model's core representations, rigorous data curation is essential. Data contamination---such as sensor, actuation, or environmental artifacts in embodied data; toxic or abusive language; and demonstrations of unsafe behavior---must be filtered early, as errors introduced at this stage are difficult to correct in later training phases~\citep{sha2024forgetting}. Once a foundation model internalizes corrupt or misleading priors, their adverse effects tend to persist and bleed into downstream behaviors, even after alignment or fine-tuning. Filtering is also necessary to remove private or personally identifiable data, particularly in humanoid datasets derived from human activities. This is critical given evidence that large models can inadvertently memorize sensitive examples~\citep{wei2025memorization}, posing privacy and safety risks in deployment.

Finally, careful attention must be paid to data balance and representation. Self-supervised learning, or most machine learning methods, naturally emphasizes most likely patterns, which can suppress rare but important cases. Underrepresentation of scarce data, such as variations in human appearance, environments, or interaction styles, can lead to biased or unsafe behaviors, including failures that disproportionately affect specific demographics~\citep{pathiraja2024fairness}. Ensuring adequate coverage or deliberate up-sampling of such cases is therefore a necessary component of responsible humanoid pre-training.

\subsubsection{Stage II: Domain Shaping through \emph{Mid-training}}
While pre-training equips LLMs with broad, general-purpose knowledge, enabling specific capabilities---such as multi-step reasoning, code synthesis, or long-context handling---typically requires an additional mid-training, also known as continued pre-training, phase. This stage is supposed to improve reasoning capabilities (Layer C2), while also making Layers C1, S1, and S2 more analytical~\citep{liu2025k2}. Mid-training augments the pre-trained model with more targeted data, either drawn from rigorously filtered high-quality corpora or synthetically generated using carefully engineered prompts~\citep{tu2025survey,mo2025mid}. Compared to pre-training, which may rely on weakly curated internet-scale data, mid-training emphasizes precision, structure, and task relevance.

Certain data types (e.g., Types II–VI in Table~\ref{tab:vla_humanoid_factors}), such as explicit reasoning traces or intermediate decision steps, are particularly important at this stage. These forms of procedural supervision provide intermediate learning signals that reward how a model arrives at an answer, rather than only whether the final answer is correct. By shaping internal reasoning processes, such data improve logical consistency, reduce hallucinations, and enable better generalization to novel problem instances. To mitigate distribution shift and catastrophic forgetting, mid-training typically employs a carefully balanced data mixture. For example, retaining 70\% of the original pre-training distribution and introducing 30\% of new, capability-specific data~\citep{tu2025survey}. In practice, the effectiveness of mid-training in current LLMs is also highly sensitive to optimization choices, including learning-rate schedules

In training humanoids, while pre-training may yield general perceptual and motor competence, advanced capabilities---such as long-horizon task execution, safety-aware manipulation, and socially legible behavior---benefit from mid-training with structured, capability-aligned data. This phase introduces procedural interaction data that exposes intermediate decision-making and control rather than only task success or failure. Such data may include additional supervision over physics and affordances~\citep{gundawar2025pac}, which are often hard to learn with internet videos used in pre-training; reasoned action sequences annotated with high-level subgoals (e.g., grasp → stabilize → orient → pour), enabling decomposable task learning; intermediate reward shaping tied to stability, safety, or human comfort instead of sparse terminal rewards; and corrective demonstrations and counterfactuals that capture near-misses and recovery behaviors, supporting anticipation and error correction.

\subsubsection{Stage III: Human Alignment through \emph{Post-training}}

After pre- and mid-training, foundation models are capable, but not yet useful or reliable for interacting with humans. Post-training focuses on shaping behavior, preferences, safety, and interaction norms so that foundation model capabilities are usable in human-facing settings.

\paragraph{Skill refinement.}
If we want to adapt a foundation model trained so far for \emph{specific tasks} such as summarization, question answering, or code generation, we apply supervised fine-tuning (SFT) using instruction–response data pairs, commonly referred to as instruction tuning~\citep{ouyang2022training}. Beyond task specialization, SFT can also shape reasoning processes, response structure, and output formats, enabling more consistent and controllable behavior across tasks.

In humanoid learning, skill refinement plays an analogous role by mapping general perceptual and motor representations to human-meaningful, instruction-level behaviors---such as ``set the table,'' ``tidy the living room,'' ``help a person stand up,'' or ``fetch an object safely.'' Rather than teaching raw motor primitives, refinement relies on instruction-like supervision (e.g., task descriptions, goal constraints, segmented demonstrations, or corrective feedback) that specifies what the robot should accomplish and how it should behave, not just how to move. This stage supports Layers S2 and S3.

\paragraph{Enabling agentic behavior.}
Post-training is increasingly used to endow LLMs with \emph{agentic AI} capabilities---such as tool use, memory, and long-horizon task execution. This typically involves SFT on tool-calling traces and multi-turn interaction trajectories, where the model learns to decide when and how to invoke external tools (e.g., search APIs, code interpreters, databases, planners, or memory stores). Unlike static offline logs used in skill refinement stage, the model is trained with real tool execution feedback, allowing it to observe tool outputs, recover from errors, and adapt its behavior based on actual outcomes rather than idealized traces.

In humanoid learning, an analogous role is played by learning to orchestrate capabilities. Humanoids inherit ``soft-tool'' use from foundation models, but must additionally learn to invoke embodied tools: sensing-related modules (e.g., state estimation, SLAM, object detection pipelines), action-related modules (e.g., grasp synthesis, balance controllers), and even everyday physical tools (e.g., utensils, instruments, fixtures). Training focuses on deciding which capability to call, in what order, and under what conditions, while integrating real-world execution feedback (sensor readings, failures, delays, physical constraints). At the HoF level, this improves Layers C1, C2, C4, S2, and S3.

\paragraph{Behavioral alignment.}
In modern LLMs, behavior alignment is achieved primarily through \emph{preference learning}, which optimizes models not against a single notion of correctness, but against comparative judgments of what humans (or aligned AI proxies) prefer~\citep{stiennon2020learning}. This includes \emph{reward modeling} (RM), where preference scores are learned from human comparisons; reinforcement learning -from human feedback (RLHF) or -AI feedback (RLAIF), which optimizes long-horizon behavior under these learned rewards~\citep{ouyang2022training}; and direct preference optimization-style (DPO), which bypass explicit RM and directly adjust the policy to favor preferred outputs over rejected ones~\citep{rafailov2023direct}. Importantly, these preferences capture soft, nuanced rules---confidence vs. humility, efficiency vs. explanation, directness vs. politeness, persistence vs. knowing when to stop---that are harder to manually specify. Behavioral alignment can further refined through style conditioning (e.g., politeness, tone, deference, assertiveness) and low-friction feedback signals such as pairwise comparisons, ratings, or best-of-N selection. Collectively, these mechanisms shape the model’s default behavior under uncertainty: when to ask clarifying questions, how cautious to be, how much initiative to take, and how to trade off competing objectives. As a result, alignment is not merely about preventing bad behavior, but about calibrating agency---deciding what kind of assistant the model should be.

In humanoids, behavioral alignment is arguably even more critical, because behavior is expressed not only through language but through motion, timing, spatial behavior, and physical intervention. Preference-based learning~\citep{biyik2022learning} maps naturally onto multiple HoF pillars. At the Physical pillar (P), alignment shapes how actions are executed: smoothness, speed, force, approach trajectories, and handover timing that feel comfortable and non-threatening rather than merely efficient. At the Cognitive pillar (C), preference learning informs decision-making under ambiguity---how assertive the humanoid should be, when to yield control, when to ask for clarification, and how to balance task success against perceived risk---directly supporting C2 (reasoning and planning) and C3 (cognitive guardrails). At the Social pillar (S), behavioral alignment governs norms of turn-taking, personal space, politeness, and intervention style, shaping whether the humanoid behaves like a tool, assistant, or teammate, and enabling calibrated trust rather than over- or under-reliance. Finally, at the Ethical pillar (E), preference learning encodes normative judgments about dignity, respect, and fairness that go beyond binary safety constraints, influencing how the humanoid treats different users and situations over time. In this sense, behavioral alignment provides the continuous control surface between raw capability and human coexistence

\paragraph{Safety and policy and guardrails.}
Post-training is the primary stage at which explicit safety constraints and normative boundaries are injected into large language models. In practice, this is achieved through a stack of complementary mechanisms. Refusal training teaches models to recognize disallowed requests and respond with calibrated, context-appropriate refusals rather than unsafe compliance. Harmful content suppression is enforced via supervised safety datasets and preference learning that penalize outputs associated with violence, self-harm, privacy violations, or illegal activity~\citep{team2025longcat}. Adversarial prompting and red-teaming expose models to systematically crafted failure cases---jailbreaks, prompt injection, role-play attacks---to harden decision boundaries under unfamiliar prompts. At a higher level, Constitutional AI~\citep{bai2022constitutional} and related rule-based critique frameworks encode abstract principles (e.g., ``do not provide instructions for harm,'' ``respect user autonomy,'' ``avoid deception'') and train models to self-critique and revise responses against these moral principles before final output. Importantly, these techniques do not merely filter outputs post hoc; they reshape the model's internal representation so that unsafe actions are no longer treated legitimate even when the prompt attempts to coerce them. The result is a bounded behavioral envelope: the model remains helpful and flexible within safe regions, but predictably resists actions outside them.

For humanoids, safety post-training plays an analogous role, but the constraints must span all four HoF pillars simultaneously. At the Physical pillar (P), guardrails enforce non-negotiable safety envelopes on motion and actuation---force, torque, speed, and proximity limits---that ensure unsafe trajectories are never executed, even if commanded. At the Cognitive pillar (C), safety mechanisms align with cognitive guardrails: monitoring uncertainty, risk, and context to trigger slow-downs, fallback behaviors, or explicit refusals when confidence drops (mirroring refusal training and constitutional self-critique in LLMs). At the Social pillar (S), guardrails regulate how and when a humanoid intervenes---preventing socially inappropriate, coercive, or deceptive behaviors, and ensuring that refusals or corrections are communicated legibly through language, gesture, or motion so that humans understand why the robot will not act. Finally, at the Ethical pillar (E), safety post-training encodes norms of nonmaleficence, privacy, and governance: respecting personal space and consent (E1), constraining what is sensed, stored, or recalled (E2), enabling human override and auditability (E3), and avoiding systematically harmful or biased behavior toward specific users (E4). Today’s robotic safety largely stops at physical safety such as collision avoidance, but shared human–humanoid environments demand guardrails across cognitive, social, and ethical pillars.

\paragraph{Human interaction calibration.}
Post-training also calibrates how models communicate with humans: providing explanations, asking for clarification, stating when they do not know, and expressing thoughts in explicit and implicit ways. These behaviors reduce hallucinations and improve user trust by aligning model confidence with actual competence. In humanoid learning, interaction calibration is essential for legibility and coordination. Humanoids must communicate intent, uncertainty, and limitations through language, gestures, or motion cues. Training policies to signal uncertainty, request assistance, or explain actions directly supports HoF goals in Layers S3, S4, and E1.

\paragraph{Improving efficiency.}
Finally, post-training often includes efficiency-oriented techniques such as model compression, quantization, and knowledge distillation. These methods reduce computational cost while preserving aligned behavior. Such efficiency improvements translate to lightweight policies that run reliably on embedded hardware with strict computational, power, and thermal limits discussed in Layer P4. Distillation and compression allow aligned, human-aware behaviors to be deployed on humanoids without sacrificing responsiveness, supporting HoF requirements for robustness, real-time interaction, and sustained operation.

Viewed across the three training stages, the distinction between classical and foundation model–based approaches becomes clear. Classical robot systems acquire task competence through narrowly scoped data and hand-engineered pipelines, leaving physical control, cognition, social interaction, and safety largely decoupled and externally regulated. In contrast, foundation models progressively integrate these dimensions through various training stages. This staged learning process yields a shared behavioral substrate in which physical coordination, reasoning, social interaction, and ethical guardrails co-evolve, potentially making foundation models particularly well suited for realizing the holistic requirements of Humanoid Factors in open-ended human environments.

\begin{figure}[h]
    \centering
\includegraphics[width=0.95\textwidth]{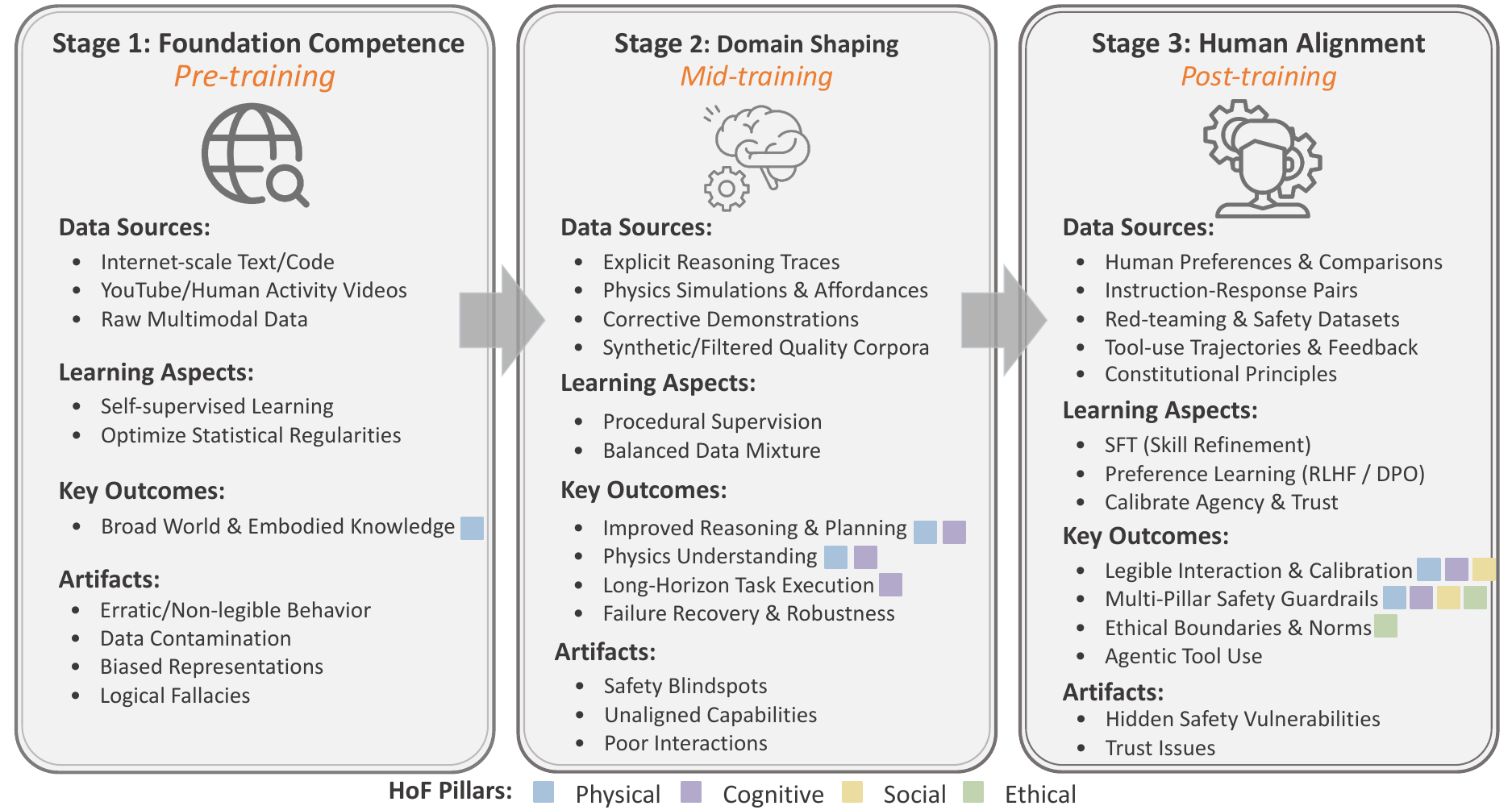}
    \caption{The training trilogy for foundation models. While this LLM-inspired training trilogy provides a useful starting point, we expect that humanoid intelligence will require further innovation beyond these three stages.}
    \label{fig:pyra}
\end{figure}

\subsection{Evaluation Approaches for Foundation-Model Humanoids}
\label{sec:eval}

Evaluating humanoid behavior in real-world settings extends beyond technical performance measures. Because humanoids operate within social, organizational, and economic contexts, meaningful evaluation must also account for systemic, cultural, and long-term effects that are not observable through short-term interaction studies or benchmark tasks alone. Accordingly, insights from the social sciences, including sociology, anthropology, economics, ethics, and policy, play a critical role in interpreting social norms, power dynamics, incentives, and downstream impacts of deployment at scale.

\subsubsection{Offline Benchmarks and Red Teaming}
Offline evaluation provides scalable, repeatable assessment of capability and robustness before human deployment. Such evaluations can be primarily conducted by engineers and safety teams prior to user exposure.

\paragraph{Does it generalize?} Generalization is evaluated by testing performance on held-out tasks, environments, and interaction scenarios not seen during training. Beyond \emph{static benchmarks}, additional test cases can be \emph{synthesized} using physics-based simulators and procedural environment generation to vary layouts, objects, contact dynamics, and human motion. Vision–language models and world models can further generate novel task instructions, ambiguous commands, or counterfactual futures---for example, alternative human approaches during a handover or unexpected changes in task goals. With foundation model–based humanoids, evaluation must extend beyond end-to-end task completion to assess core underlying skills that enable task completion using skill diagnostic benchmarks that probe whether tasks are executable in principle, as exemplified by PAC Bench~\citep{gundawar2025pac}. Some offline evaluations can be scaled using automated judges, including foundation models trained to assess plan quality, constraint violations, or explanation adequacy. While LLM-as-judge approaches~\citep{tan2024judgebench} enable rapid triage and red-teaming, they should be calibrated against expert and human judgments, particularly for social and ethical failure modes. These tests primarily probe cognitive situational awareness and planning (C1, C2), as well as physical action robustness (P3), under distribution shift.

\paragraph{Is it robust?}
Robustness evaluation examines sensitivity to sensor noise, latency, partial observability, and scarce but safety-critical edge cases such as near-collisions, dropped objects, or sudden human intervention. At a basic level, this is tested through domain randomization, including perturbations to sensing, actuation, and timing. More targeted \emph{counterfactual and adversarial scenarios} deliberately expose failure modes, such as conflicting language instructions, deceptive visual cues, or occlusions during manipulation. Adversarial examples can be generated using gradient-based~\citep{huang2017adversarial} or search-based methods\citep{sagar2024failures, sagar2025from} over perception inputs or environment parameters to characterize \emph{failure modes}. A key indicator of robustness is whether the system appropriately modulates its behavior---slowing down, pausing, or refusing---when uncertainty increases. Robustness failures often manifest as breakdowns in cognitive guardrails (C3) or unsafe physical behavior (P3, E1), particularly when uncertainty is not correctly detected or acted upon.

\subsubsection{Human-in-the-Loop Evaluation}

Human-in-the-loop studies assess whether humanoid behavior aligns with human expectations, comfort, and collaboration norms, properties that cannot be fully inferred from offline tests alone~\citep{murthy2025llms}. Such studies can be conducted with representative end users under oversight of human factors researchers.

\paragraph{Is the behavior legible and predictable?}
Legibility and predictability should be evaluated through controlled laboratory studies in which participants perform joint tasks with the humanoid, such as handovers, shared-workspace manipulation, or guided navigation. In addition to task completion and timing measures, evaluations include behavioral indicators of anticipation and hesitation, such as delayed responses, avoidance, or repeated checking. For instance, expressivity can be quantified through measures such as gaze latency, pre-motion cues, cue vocabularies (e.g., lights or audio), and actuation limits, ensuring that communicative signals are perceivable, timely, and interpretable by humans. Situation awareness probes can be used to assess whether humans correctly understand what the humanoid is doing and what it will do next, for example by briefly pausing interaction and querying participants’ expectations. Wizard-of-Oz~\citep{10.1007/978-3-030-49183-3_11} or mixed-autonomy setups are often used to isolate the effects of perception, planning, and communication from low-level control errors. \emph{Explainable AI}~\citep{Sagar2024arxiv_xaitcavrob,molnar2020interpretable} and \emph{uncertainty quantification}~\citep{senanayake2024role} mechanisms for humanoids are essential for evaluating not what the humanoid is doing, but why, how confident it is, and when it may be appropriate to intervene or take over. These studies directly evaluate legibility and shared situational awareness (C1–C3), as well as social signaling and intervention timing (S2–S3).

\paragraph{Does it support trust, appropriate reliance, and low cognitive burden?}
Trust and reliance are evaluated through repeated and longitudinal interaction studies, examining whether users over-rely on the humanoid, under-utilize it, or calibrate reliance appropriately as behavior evolves. Standard human factors instruments—such as NASA-TLX for workload and validated trust or usability scales—are complemented by behavioral measures including intervention frequency, interruption recovery, and coordination effort. Expectancy-violation analysis is particularly informative: moments where humanoid behavior surprises or confuses users often reveal misalignment between learned behavior and human mental models, even when task performance remains high. Together, these measures assess whether adaptation and autonomy reduce human cognitive load and support stable, predictable collaboration over time. Trust calibration and workload outcomes reflect alignment across social appearance and behavior (S4), cognitive load management (C4), and ethical expectations around autonomy and control (E1).

\subsubsection{Deployment-Aware and Life Cycle Evaluation}

Deployment-aware evaluation can be structured using governance frameworks such as responsible AI scorecards~\citep{vorvoreanu2023responsible}, which organize evidence across safety, reliability, fairness, privacy, and accountability dimensions over the system lifecycle. These evaluations can be conducted by operators, maintainers, and governance bodies during deployment.

\paragraph{Does behavior drift or degrade over time?}
Behavioral drift is evaluated by monitoring changes in action distributions, interaction patterns, and decision outcomes across extended deployments. This includes tracking deviations from expected motion profiles, increases in intervention or override frequency, and shifts in confidence or refusal rates. Change point detection and anomaly monitoring~\citep{agia2024unpacking} can identify emerging failure modes, while periodic regression tests assess whether updates or continued learning degrade previously verified behaviors. For instance, runtime evaluation should verify the effectiveness of monitors and safety shields that track uncertainty, contact forces, protected zones, and out-of-distribution signals, triggering slow-down, stop, or human handoff when risk increases. Longitudinal logs and replay tools support post hoc analysis of near-misses and subtle safety regressions. Importantly, not all behavioral drift originates from the foundation model itself; changes may arise from sensor miscalibration, hardware wear, thermal effects, or environmental degradation, yet a robust foundation model–based system should be able to recognize these shifts, adapt its behavior accordingly, or explicitly alert humans when operating outside its reliable regime. Lifecycle evaluation is especially critical for ethical and operational HoF layers, including long-term safety (E1), governance and accountability (E3), and sustained operation under real-world constraints (P4).

\paragraph{Does it respect policies, constraints, and recovery protocols?}
Lifecycle evaluation also examines whether the humanoid consistently adheres to safety, privacy, and governance constraints during real-world use. This includes verifying the effectiveness of human override mechanisms, safe fallback behaviors, and escalation pathways when uncertainty or risk increases. Policy compliance checks, incident reporting, and audit logs provide evidence of accountability, while distribution shift detection assesses whether the system encounters conditions outside its validated operating domain. Recovery-focused evaluation---such as testing safe stop, handoff, or degraded-operation modes---ensures that failures are managed transparently rather than silently. These mechanisms operationalize ethical guardrails by ensuring failures are observable, attributable, and recoverable rather than silent (E1–E4).

\subsubsection{Using Evaluations for Continuous Improvement}

Evaluation signals collected through offline benchmarks, human-in-the-loop studies, and deployment-aware monitoring should be used to drive targeted and continuous improvement of humanoid foundation models~\citep{sagar2024failures,sagar2025from}. Rather than treating evaluation as a pass–fail gate, these signals identify specific deficiencies in capability, alignment, or robustness that can be addressed through data curation, post-training, and runtime safeguards.

Offline evaluation results primarily inform capability and robustness refinement. Failures observed in red teaming, counterfactual scenarios, or diagnostic benchmarks can guide targeted data generation and mid- or post-training updates, for example by adding task variants that expose missing prerequisite reasoning, poor affordance understanding, or brittle perception–action coupling. Scarce-event failures can also motivate tighter safety constraints or additional uncertainty calibration during post-training.

Human-in-the-loop evaluation provides feedback for behavioral and interaction-level refinement. Measures of legibility, trust calibration, workload, and expectancy violations can be translated into preference signals for post-training, shaping how the model communicates intent, times interventions, or expresses uncertainty. Recurrent interaction failures---such as overconfident actions or confusing cues---can be addressed through supervised fine-tuning on corrected demonstrations, preference learning, or explicit instruction-style data that encode desired interaction norms.

Deployment-aware evaluation supports long-horizon adaptation and governance. Monitoring behavior drift, near-misses, and policy violations enables selective retraining, guardrail updates, or rollback of problematic model changes. Importantly, not all detected issues require model updates: some failures are best addressed through sensor recalibration, hardware maintenance, or changes in operational procedures. Evaluation signals therefore help determine whether corrective action should occur at the software, hardware, or cultural level.

\subsection{What Foundation Models Enable -- and What They Do Not}

Building on the evaluation modalities in Section ~\ref{sec:eval}, we now clarify what AI foundation models enable in humanoid systems—and what they do not. While benchmarks and human studies provide valuable signals, they can also misattribute competence or inflate expectations if interpreted without care. This section therefore highlights key limitations and frames humanoid design around coexistence with humans, rather than human equivalence, using the Humanoid Factors perspective.

\subsubsection{Threats to Validity in Foundation Model Training and Evaluation}

\noindent {\bf Danger 1: Implicit leakage of benchmark information into training.}
Consider a benchmark task that evaluates a foundation model on a novel tabletop manipulation problem: ``Given a cluttered scene, pick up the red block and place it into the blue bin.'' During training, the model may not encounter this exact scene, object instance, or instruction. However, suppose the training corpus contains instructional or demonstration-derived content such as: ``For pick-and-place tasks, first localize the target object, then plan a grasp aligned with its principal axis, lift vertically to clear obstacles, and execute a place motion over the receptacle.'' Although this instruction does not overlap with the benchmark at the level of images, trajectories, or natural-language prompts, it encodes the full procedural template required to solve the task. Standard \emph{decontamination} methods that focus on surface similarity would fail to remove such data, resulting in false negatives. At evaluation time, apparent success may therefore reflect recall of known strategies rather than genuine generalization to unseen tasks. This risk is especially pronounced when training on internet-scale or demonstration-heavy corpora.\\

\noindent {\bf Danger 2: Misleading coverage and realism in synthetic data.}
Synthetic data generated through physics-based simulators, domain randomization, or generative world models is invaluable for scaling training and evaluation, but it can introduce systematic distortions. For instance, simulated red teaming may over-represent certain failure modes while missing others that arise only through embodied interaction, sensor drift, or social context. As a result, models may appear robust under synthetic perturbations yet fail under real-world conditions that violate simulator assumptions. Without careful validation, especially on social and ethical context, synthetic data can create a false sense of coverage and robustness.\\

\noindent {\bf Danger 3: Bias and blind spots in generative evaluators.}
When generative models are used for evaluation---for example, through LLM-as-judge or automated critique---the evaluator may encode its own biases, preferences, or blind spots. This is particularly problematic when assessing social, ethical, or interactional failure modes that lack clear ground truth. Such evaluators may systematically under-penalize behaviors that appear linguistically fluent or norm-following while missing subtle harms, coercion, or miscalibrated authority. Automated judgments therefore require careful calibration against expert review and human subject data.\\

\noindent {\bf Danger 4: Human biases and experimental artifacts.}
When humans are involved in proving inputs to generative models, as in RLHF, or user studies for evaluation, multiple sources of bias arise. Small or non-representative participant pools may fail to capture variability in comfort, trust, cultural norms, or interaction styles, leading to overly optimistic conclusions about alignment. Experimental framing effects---such as how the system’s capabilities are introduced or how tasks are explained---can strongly influence user expectations, trust calibration, and reported workload. Over-reliance on subjective self-report measures further compounds this issue: users may report high trust or satisfaction while exhibiting hesitation, avoidance, or over-reliance behaviors that signal misalignment. Conversely, novelty effects may temporarily inflate perceived performance or acceptance, masking longer-term issues that emerge only with sustained use.\\

\noindent {\bf Danger 5: Distribution shift between evaluation and deployment.}
Evaluations are often conducted under controlled conditions that differ substantially from real-world deployment. Behavioral drift observed post-deployment may stem from interacting factors such as model updates, sensor degradation, hardware wear, environmental change, or shifts in user behavior. Without sufficient logging, uncertainty monitoring, and explainability, it can be difficult to attribute failures correctly. This ambiguity risks inappropriate corrective actions, such as unnecessary retraining, unsafe rollback, or misdirected policy changes.\\

\noindent {\bf Danger 6: Metric gaming and checklist-driven evaluation.}
There is a risk that evaluation practices become overly optimized for a fixed set of benchmarks, metrics, or governance scorecards. Models may learn to satisfy observable criteria while exploiting unmeasured degrees of freedom, leading to brittle or superficial alignment. Near-misses, discomfort, or minor policy violations may go unreported if monitoring focuses only on overt failures. Over time, evaluation frameworks can degrade into checklist compliance unless paired with concrete behavioral evidence, longitudinal monitoring, and mechanisms for surfacing unexpected failure modes.\\

\noindent {\bf Danger 7: Over-attribution of capability to the model alone.}
In humanoids, observed performance reflects not only the foundation model but also the surrounding stack, including controllers, heuristics, safety layers, and human oversight. Evaluations that attribute success or failure solely to the learned generative model risk conflating architectural scaffolding with learned competence. This can obscure true model limitations and lead to incorrect conclusions about generalization, robustness, or readiness for deployment.

\subsubsection{Helpful Capabilities Without Agency}

Capabilities of the foundation models are always bound to the constraints imposed by embodiment (Layers P1-P3). Physical limits—such as computing, stability, power, and sensors—remain dominant factors in humanoid performance and safety. Robust behavior therefore emerges from the integration of learned representations with explicit system-level structure, monitoring, and governance, rather than from the foundation model alone. Foundation models enable broad generalization by learning shared representations across perception, language, and action. In humanoids, this allows skills, concepts, and behaviors learned in one context to transfer to others with minimal task-specific retraining. 

However, these capabilities should not be conflated with human-like agency. Foundation models do not possess intrinsic goals, intent, responsibility, or moral judgment. Their behavior is shaped by objectives, data, and constraints imposed externally by designers, operators, and surrounding system infrastructure. As a result, observed performance reflects not only the model itself but also controllers, heuristics, safety layers, and human oversight. Treating foundation models as autonomous agents risks over-attributing competence and obscuring where responsibility and control truly reside.

\subsubsection{Human--Humanoid Overlap Without Equivalence}

Humans and humanoids inevitably share environments, tasks, and interaction spaces, leading to partial overlap in perception, action, and communication. Yet \emph{they differ fundamentally in physiology, cognition, learning history, vulnerability, and accountability.} Attempting to optimize humanoid systems for human-likeness or parity can introduce significant risks, including deceptive interactions, uncanny behavior, and miscalibrated trust.

The objective of humanoid design is therefore not to make machines human-like, but to make them human compatible. This requires recognizing both where overlap is necessary for coordination and where differences must be preserved for safety, clarity, and trust. The HoF framework provides a principled way to identify these boundaries across Physical, Cognitive, Social, and Ethical pillars. By making the limits of capability and agency explicit and legible, HoF supports the design of humanoid systems that can coexist with humans in shared environments without eroding human expectations, responsibility, or control.

\section{Applying Humanoid Factors: Illustrative Experiment on a Humanoid}
\label{sec:experiment}


Having introduced the four pillars of Humanoid Factors in Section~\ref{sec:framework} and discussed how they can be incorporated into training and evaluation in Section~\ref{sec:vla}, we now illustrate the framework through a proof-of-concept case study on a humanoid platform. This case study is intentionally narrow: one robot, one operator, and one reaching/pointing task. Its purpose is not to fully validate the HoF framework, but to show how a human-derived evaluation primitive can expose failure modes that task-completion metrics alone can miss. While comprehensive evaluation should ultimately account for all four pillars, we focus here on the intersection of the Cognitive and Physical pillars. Specifically, we examine the question: \emph{Does training a robot to achieve geometric task success also lead to motion patterns that remain cognitively legible to humans?}

\subsection{Cognitive Models as Humanoid Evaluation Primitives}

\subsubsection{Human-Centered Motivation}

Humanoids are expected to operate across a wide range of settings. While some deployments may occur in relatively isolated environments such as factories, many envisioned applications involve close, continuous interaction with people---for example in homes, clinics, or caregiving contexts. In such settings, humanoids must engage in everyday collaborative actions, such as \emph{handing objects to a person}, in ways that align with human expectations formed through human–human interaction. In these situations, humans will increasingly encounter humanoids built by different manufacturers and powered by diverse AI models. From the human's perspective, however, these distinctions are largely irrelevant: regardless of the underlying model or brand, the humanoid is expected to behave in ways that are predictable, understandable, and safe. Making sure this expectation is met is therefore not just a deployment concern, but a responsibility of the AI models that generate the humanoid’s behavior.

Current evaluation practices for humanoid AI models, however, remain predominantly \emph{task completion-centric}, emphasizing whether a task is completed or a predefined checkpoint is reached. From a HoF perspective, this is insufficient. In human-facing settings, \emph{how a humanoid moves is as critical as whether it succeeds ``somehow.''} For a humanoid to coexist comfortably with people, its motion must be \emph{legible}: a human observer should be able to intuitively anticipate the robot’s goal, timing, and level of commitment from its movement alone~\citep{dragan2013legibility}.

Human sensitivity to motion predictability is not arbitrary; it is shaped by robust regularities in human motor behavior that govern how speed, timing, and variability adapt to task demands (often modeled by psychophysical laws). When a humanoid completes a task while violating these implicit expectations---for instance by moving with uniform speed across easy and difficult segments, or by hesitating without clear intent---it can create cognitive dissonance for the human partner, undermining trust, comfort, and perceived safety~\citep{dragan2013legibility, howell2023comfortDynamics}.

\subsubsection{Fitts' Law} 

One of the most widely studied psychophysical models of human movement is Fitts’ law~\citep{fitts1954information,mackenzie2018fitts}, which describes how the time required to complete a goal-directed movement scales with task difficulty. Originally developed to model pointing and reaching actions, it has since been validated across a broad range of real-world motor activities, including computer input device usage analysis~\citep{senanayake2013model,senanayake2013superiority}, tool use~\citep{silva2016task}, surgical and rehabilitation tasks~\citep{mccrea2005consequences,zimmerli2012validation}, vehicle and cockpit controls~\citep{large2015predicting, xie2023fitts}, and everyday object manipulation~\citep{kantowitz1988fitts,thumser2018fitts}. Across these domains, the law captures how humans systematically adapt movement speed and timing in response to precision demands, producing characteristic motion profiles that are highly predictable to observers. For instance, in human–human object handover, one subcase of the object transportation phase~\citep{kopnarski2023systematic} involves moving an object toward a fixed target location corresponding to the partner’s hand, yielding timing and velocity patterns that resemble those of reaching actions.

In psychophysics, the time required to complete such a reaching or pointing movement (\textit{Movement Time} or MT) scales logarithmically with the ratio of distance to the target along the direction of movement, $D$, to target width, $W$, capturing the trade-off between speed and accuracy. Depending on this logarithmic ratio, known as the \textit{Index of Difficulty} ($ID_F$), human motor control operates in two different regimes. For higher-difficulty tasks ($ID_F > 3$--$4$), they become \emph{visually controlled}, relying on continuous visual feedback for precision; for low-difficulty tasks ($ID_F < 3$), movements are largely \emph{ballistic} and governed by open-loop control;~\citep{gan1988geometrical}. These two regimes are classically modeled as:
\begin{align}
    MT &= a_1 + b_1 \log_2\!\left( \frac{2D}{W} \right) = a_1 + b_1 (ID_F), \quad &\text{for visual control regime,} \label{eq:fitts}\\[4pt]
    MT &= a_2 + b_2 \sqrt{D}, \quad &\text{for ballistic regime.} \label{eq:ballistic}
\end{align}


Traditionally, from an engineering perspective, AI-based robot controllers or deterministic planners typically optimize for trajectory efficiency, torque, or smoothness (e.g., minimum jerk) without inherently adhering to Fitts-like scaling~\citep{dragan2013legibility, li2022humanlikeMotionPlanningRobotArms} because robots are not constrained by biological neuromotor limits~\citep{takeda2019explanation}. Consequently, it is possible to use Fitts' Law not as a control limit, but as a benchmark for measuring the ``naturalness'' and usability of assisted teleoperation and mobile manipulation~\citep{pan2024fittsBenchmarkAssistedHR, wan2023performance}. Similarly, Fitts' law can be used to test behavioral cloning (BC)-based neural networks~\citep{10.5555/3304652.3304697} or any humanoid foundation model (Table~\ref{tab:vla_humanoid}). 

\begin{definitionbox}
\textbf{Using HoF for Humanoid Evaluation:} \\
\quad 1. Identify the humanoid task of interest and associated human expectations.\\
2. Select the HoF layer most directly implicated by an expectation (e.g., C2 for timing and execution regularities).\\
3. Choose a human-validated primitive that matches the task (e.g., Fitts' Law for reaching, pointing, and handover sub-movements where speed--accuracy trade-offs are central).\\
4. Test alignment with that primitive in addition to standard task-completion metrics.\\
5. Interpret deviations in alignments as diagnostic signals about human compatibility, not as proof that the humanoid should mimic humans in all respects.
\end{definitionbox}



\begin{figure}[t]
    \centering
    \begin{minipage}[c]{0.4\linewidth}
        \begin{subfigure}[b]{\linewidth}
            \centering
            \includegraphics[width=\linewidth]{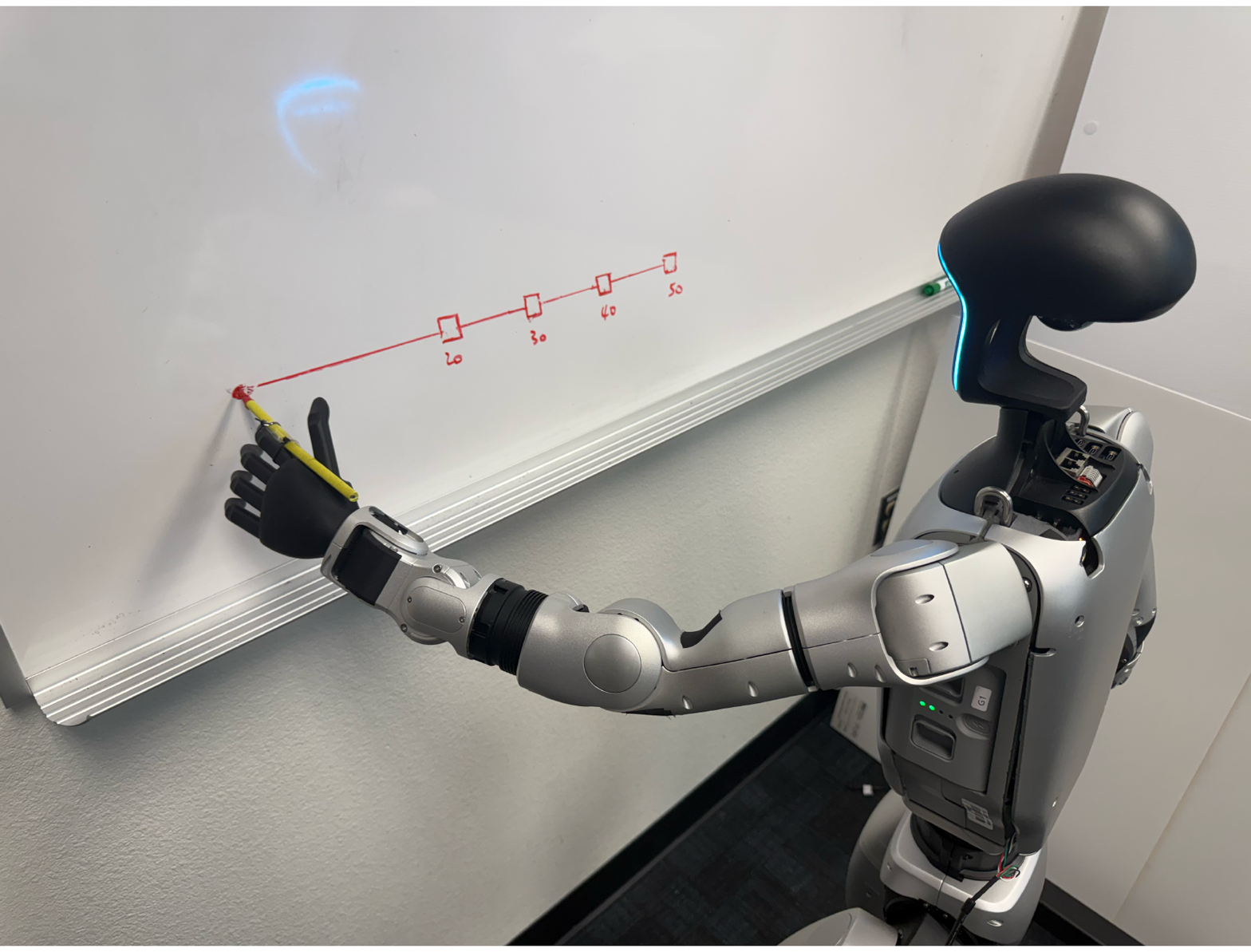} 
            \label{fig:setup}
        \end{subfigure}
    \end{minipage}
    \hfill 
    \begin{minipage}[c]{0.58\linewidth}
        \centering
        \begin{subfigure}[b]{\linewidth}
            \centering
            \includegraphics[width=\linewidth]{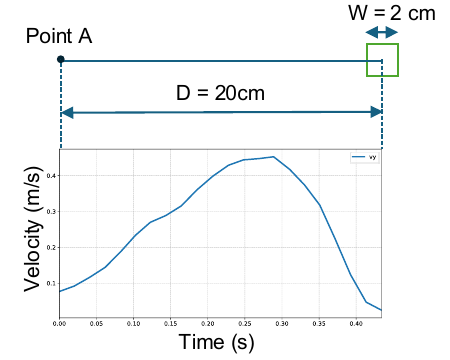}
            \label{fig:joint_data}
        \end{subfigure}
    \end{minipage}
    
    \caption{{Experimental Environment and Data.} (a) Unitree G1 robot setup. (b) Example end-effector velocity profile illustrating the pencil tips motion from the start point to the target region for a single demonstration ($D=20$\ cm and $W=2$\ cm).}
    \label{fig:experiment_overview}
\end{figure}
\subsection{Experiment}

\subsubsection{Hypothesis} 
We hypothesize that although a kinematics-only BC policy can achieve task completion, its execution time will not follow the Fitts' law relationship between movement time $MT$ and index of difficulty $ID$. In particular, the fitted $MT$--$ID$ trend for the learned policy will deviate from that of human demonstrations (e.g., reduced correlation and/or different slope), reflecting a failure to reproduce human temporal scaling with difficulty.

\subsubsection{Experimental Setup} 

\paragraph{Hardware platform.}
We conducted experiments using the Unitree G1 humanoid robot, as shown in Figure \ref{fig:experiment_overview}. The experiment focused on a pointing task using the left arm (7-DoF). A pencil was attached to the robot's hand to serve as a precise end-effector extension for the pointing task. We only controlled four joints (left shoulder yaw, pitch, roll; left elbow) while locking the wrist and all other joints. We command only a high-level sequence of target joint angles along the trajectory suggested by the AI model. Low-level motor control is handled by the robot’s built-in PID controllers, which track these joint targets.

\paragraph{Task.} The robot was initialized at its default neutral starting position. The goal was to move the end-effector into a $2 \times 2$ cm square target region ($W=0.02$ m). We varied the distance ($D$) across four levels: 0.20 m, 0.30 m, 0.40 m, and 0.50 m. A human operator provided 99 successful demonstrations via kinesthetic teaching, physically guiding the robot arm (the pencil tip) from Point A to the target box as quickly as possible. Trials containing errors were discarded.

\paragraph{AI Model Training and Deployment.}
We consider the task of training a robot to imitate human behavior from multiple demonstrations of the task described above. We trained a conditional Behavior Cloning (BC) policy with  multi-layer perceptron (i.e., fully connected) neural network by minimizing the mean squared error (MSE) loss between predicted joint angles and target joint angles. After training, the model was evaluated first in simulation using MuJoCo and then on the real robot. For deployment, we used an open-loop autoregressive strategy in which the model’s previously predicted joint angles were fed as input for the next step, rather than using the robot’s actual sensor readings. This choice deliberately stress-tests the stability of the learned rollout without corrective feedback, but it also means that policy errors, controller behavior, embodiment effects, and simulator mismatch are not cleanly disentangled. Accordingly, the real-robot results should be interpreted as a deployment-level failure at the Cognitive--Physical interface rather than as definitive attribution to any single component.

\subsection{Results and Discussions}
We analyzed the adherence to Fitts' Law across the human baseline, simulation, and real-world deployment.

\begin{figure*}[t!]
    \centering
    \begin{subfigure}[b]{0.48\textwidth}
        \centering
        \includegraphics[width=\linewidth, height=0.2\textheight, keepaspectratio]{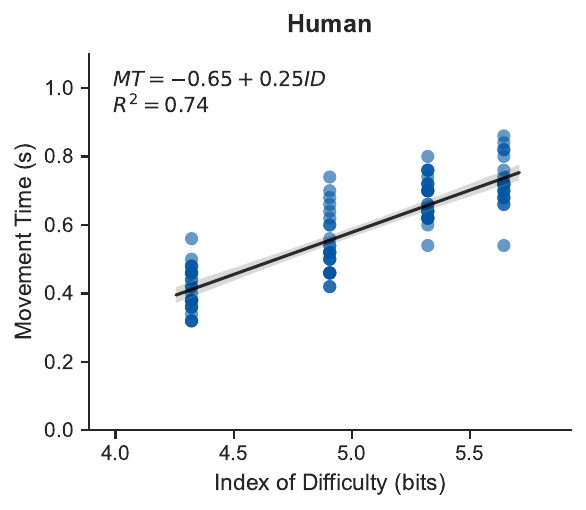}
        \caption{\textbf{Human Baseline:} Strong linear fit ($R^2 = 0.741$), indicating natural speed-accuracy trade-off.}
        \label{fig:human_fitts}
    \end{subfigure}
    \hfill
    \begin{subfigure}[b]{0.48\textwidth}
        \centering
        \includegraphics[width=\linewidth, height=0.2\textheight, keepaspectratio]{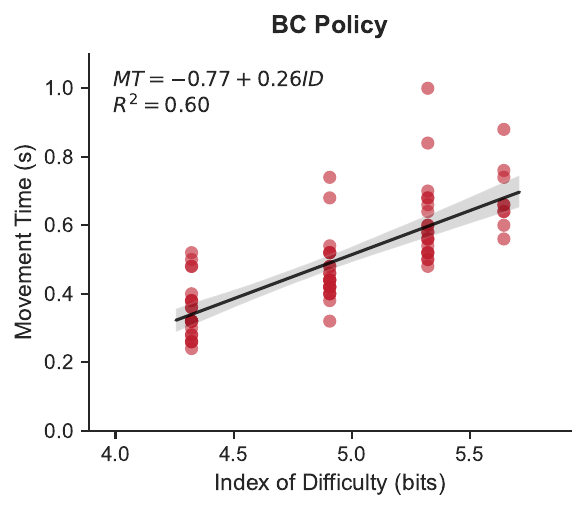}
        \caption{\textbf{Simulation Rollout:} Parallel slope to human data ($R^2 = 0.596$), suggesting successful velocity cloning.}
        \label{fig:sim_fitts}
    \end{subfigure}
    
    \vspace{5pt} 
    
    \begin{subfigure}[b]{0.48\textwidth}
        \centering
        \includegraphics[width=\linewidth, height=0.2\textheight, keepaspectratio]{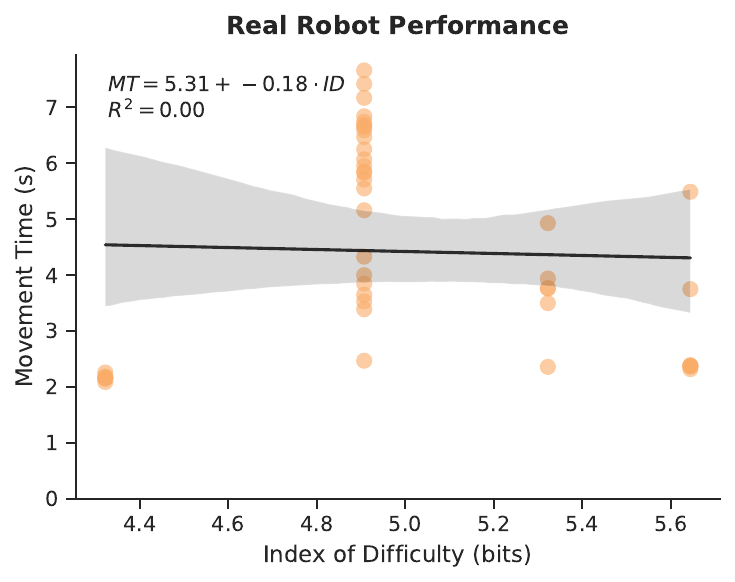}
        \caption{\textbf{Real World ($K_p=60$):} Significant degradation. The low stiffness fails to reject gravity, causing drift and destroying the linear relationship.}
        \label{fig:real_kp60}
    \end{subfigure}
    \hfill
    \begin{subfigure}[b]{0.48\textwidth}
        \centering
        \includegraphics[width=\linewidth, height=0.2\textheight, keepaspectratio]{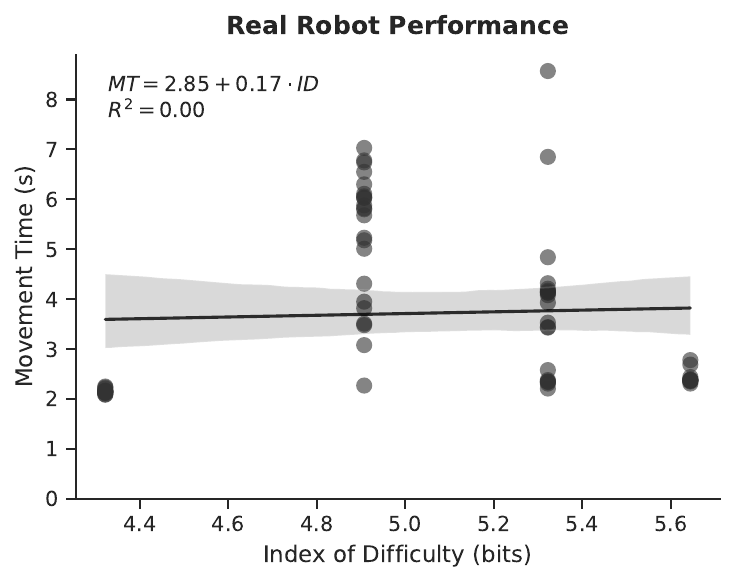}
        \caption{\textbf{Real World ($K_p=80$):} Higher stiffness improves success rate but fails to restore Fitts' Law linearity. The ``Lever Arm'' effect at $D=0.5$ m remains uncorrected.}
        \label{fig:real_kp80}
    \end{subfigure}
    
    \caption{{Does the robot follow Fitts' Law?} (a) The human operator establishes a cognitive baseline. (b) The simulation captures the general trend but introduces variance. (c-d) Real-world physical constraints (gravity, friction) decouple the motion from the cognitive plan, with higher stiffness ($K_p=80$) offering only marginal recovery of linearity.}
    \label{fig:fitts_grid}
\end{figure*}

\paragraph{Human baseline (gold standard).}
The kinesthetic demonstrations (see Figure \ref{fig:human_fitts}) exhibited a strong adherence to Fitts’ Law ($R^2 = 0.741$). This confirms the validity of our setup: even when manipulating a robot arm, the human operator naturally engaged in a speed-accuracy trade-off.

\paragraph{Simulation rollout (cognitive success).}
As shown in Figure \ref{fig:sim_fitts}, the policy generated trajectories that roughly paralleled the human baseline in the MuJoCo physics simulation.
The fit was moderate ($R^2 = 0.596$), while the slope of the robot's regression line was basically parallel to the human's. This suggests partial alignment with the human speed--accuracy trend observed in simulations, but it does not provide a definitive demonstration that the policy has adopted the same control strategy in real-world deployment conditions.

\paragraph{Real-world deployment (physical failure).}
We deployed the policy on the physical Unitree G1 robot at two stiffness levels. For low stiffness ($K_p=60$), as seen in Fig.~\ref{fig:real_kp60}, the Fitts' Law relationship collapsed. The arm was too compliant to hold the target pose against gravity, leading to excessive drift and erratic movement times. For high stiffness ($K_p=80$), increasing the gain (Fig. \ref{fig:real_kp80}) improved the success rate by rigidly enforcing the trajectory. However, it did not restore the linear trend. The persistent breakdown at large reach distance ($D=0.5$ m) is consistent with embodiment effects such as gravity loading that the open-loop rollout does not compensate for, rather than demonstrating a purely cognitive success carried cleanly into deployment.

\subsubsection{Discussion and Conclusion}
The disparity between the simulation success and real-robot breakdown provides a proof-of-concept demonstration of what HoF-style evaluation can reveal. In this experiment, Fitts' Law revealed problems in the robot's behavior that would have been missed by task completion alone. At the same time, the study should be interpreted cautiously: because deployment is open-loop and the setup uses one robot, one operator, and one pointing task, the results do not isolate a single cause and do not by themselves validate the entire HoF framework.

\paragraph{Plausible policy-level contributors (Cognitive Pillar).}
In both simulation and reality, the robot frequently arrived at the target area but failed to stop, instead ``orbiting'' or oscillating around the $2$ cm target. We attribute this to the MSE loss used in training. MSE penalizes large errors heavily but offers weaker learning signal near the target. Consequently, the robot learns to \textit{approach} the target without robustly learning when and how to \textit{stop}. This violates the Cognitive Pillar: a human partner cannot interact with a robot that hovers indecisively.

\paragraph{Plausible embodiment-level contributors (Physical Pillar).}
The most severe failures occurred at the extremes of the reach, suggesting that the robot's physical dynamics influenced performance in ways the learned policy did not explicitly model. In the far field ($D=0.5 m$), gravity torque on the shoulder increases with extension, harder for the robot to follow the intended motion and hold the final position accurately. In the near field ($D=20$ cm), the folded configuration likely reduces manipulability, so small joint-space errors can produce disproportionately unstable end-effector motion. Because the experiment is open-loop, these embodiment effects are entangled with policy behavior rather than cleanly separable from it.

\paragraph{Why the entanglement matters for HoF.}
This entanglement is part of the underlying design challenge that HoF is intended to analyze. The same rollout can simultaneously register as a Cognitive issue (hesitation and poor stopping), a Physical issue (gravity, compliance, manipulability), a Social issue (reduced predictability to nearby humans), and potentially an Ethical/Safety issue if users over-trust motion that appears more controlled than it is. The case study therefore illustrates why the four pillars should be treated as interacting analytical lenses rather than independent boxes.

\paragraph{Implication.}
The main takeaway is not that Fitts' Law universally validates or invalidates humanoid intelligence, but that human-centered evaluation primitives can expose deployment failures missed by completion-centric metrics. For foundation-model-based humanoids, this suggests a broader recipe: pair conventional robotics metrics with task-matched human-centered diagnostics so that policies are evaluated not only for whether they succeed, but for whether they remain predictable and compatible in human environments.





\section{Implications for Research and Product Design}


This section synthesizes the Humanoid Factors framework (Section~2) and its operationalization through motion-level analysis (Section~4) to outline implications for humanoid design, research methodology, and deployment. Rather than treating humanoid robotics as a single technical problem, Humanoid Factors reframes it as a \emph{systems-level challenge of coexistence}, spanning engineering, human factors, social interaction, and governance. As humanoids transition from tools to semi-autonomous collaborators, the central design question shifts from optimizing a single intelligent agent to jointly shaping human--humanoid ecosystems. The implications below highlight where current approaches fall short and where new research, standards, and investment are required.

\subsection{Humanoid Design and Evaluation}

\paragraph{From task completion to human compatibility.}
Conventional robot evaluation prioritizes task completion and speed. This framing inherits an implicit assumption from classical automation: that the human remains the sole intelligent agent, and the robot is merely an instrument. In coexisting human--humanoid environments, this assumption no longer holds. Humanoids increasingly act as teammates, collaborators, or surrogates, and system performance becomes a property of the \emph{human--humanoid pair}, not the robot alone. The Humanoid Factors perspective reveals that traditional metrics are therefore insufficient. The case study in Section~4 demonstrates that a robot trained with Behavior Cloning can succeed geometrically while failing to exhibit human-like speed--accuracy trade-offs, resulting in motion that is technically correct but ergonomically incompatible. This implies that evaluation must explicitly incorporate human-centered criteria such as predictability, comfort, timing, and perceived naturalness---criteria that directly affect coordination, trust, and workload.


\paragraph{Integrated evaluation across pillars.}
Physical, cognitive, social, and ethical pillars cannot be evaluated independently. A physically safe motion may still violate social norms; a cognitively competent plan may be ethically inappropriate in context; a socially fluent interaction may obscure limitations and encourage over-trust. Humanoid Factors therefore calls for integrated evaluation frameworks that treat humanoids as first-class agents with their own constraints, failure modes, and responsibilities. Rather than asking whether a robot is ``safe'' or ``accurate'' in isolation, evaluation must report \emph{multi-dimensional outcomes} and explicitly characterize trade-offs across pillars. This mirrors the evolution of Human Factors, where performance, workload, and safety are jointly analyzed rather than optimized independently.

\begin{sidewaystable}
    \centering
    \scriptsize 
    \caption{Humanoid Stakeholder Analysis Framework}
    \label{tab:stakeholder_analysis}

    \begin{tabularx}{\textwidth}{>{\RaggedRight\bfseries}X >{\RaggedRight\bfseries}X >{\RaggedRight}X >{\RaggedRight}X >{\RaggedRight}X >{\RaggedRight}X >{\RaggedRight\arraybackslash}X}
        \toprule
        Category &
        Stakeholder &
        \cellcolor[HTML]{A6BFD8}Physical &
        \cellcolor[HTML]{B3AACB}Cognitive &
        \cellcolor[HTML]{F4E4A5}Social &
        \cellcolor[HTML]{C3D8B2}Ethical &
        Primary Use \\
        \midrule

        \multirow{4}{*}{\parbox{2.2cm}{\centering\textbf{System\\ Builders}}} &
        Robot Engineers &
        Morphology, sensing, actuation limits, safety envelopes &
        Planning, uncertainty handling, memory limits &
        Expressivity channels, intervention timing &
        Hard constraints, fail-safes &
        System design, controller tuning \\
        \addlinespace

        & Machine Learning Researchers &
        Embodiment grounding, action feasibility &
        Reasoning, generalization, verification &
        Legibility, intent signaling &
        Bias, data use, alignment &
        Model architecture, training objectives \\
        \addlinespace

        & Product Designers / HCI &
        Form factor, interaction surfaces &
        Explainability, feedback &
        Communication style, affordances &
        Transparency, consent &
        Interaction design, UX \\
        \addlinespace

        & System Integrators / Deployers &
        Site constraints, infrastructure compatibility &
        Configuration complexity, fault diagnosis &
        Workflow disruption, staff training &
        Responsibility boundaries, handoff clarity &
        System integration, rollout \\
        \addlinespace\midrule

        \multirow{3}{*}{\parbox{2.2cm}{\centering\textbf{Safety, Reliability,\\ \& Assurance}}} &
        Safety Engineers &
        Collision, force, stability &
        Failure detection, escalation &
        Misuse prevention &
        Non-maleficence &
        Risk analysis, certification \\
        \addlinespace

        & Maintenance \& Service Personnel &
        Wear, calibration access, component replacement &
        Diagnostics, troubleshooting procedures &
        Downtime communication, coordination &
        Safe servicing practices, lockout protocols &
        Repair, upkeep \\
        \addlinespace

        & Emergency Responders &
        Physical access, emergency stop reliability &
        Situation assessment under failure &
        Human--robot coordination in crises &
        Duty of care, override authority &
        Incident response \\
        \addlinespace\midrule

        \multirow{3}{*}{\parbox{2.2cm}{\centering\textbf{Human Interaction\\ \& Experience}}} &
        Human Factors Researchers &
        Ergonomic compatibility, motion naturalness &
        Predictability, workload impact &
        Trust calibration, coordination &
        Human dignity, agency &
        Evaluation frameworks, user studies \\
        \addlinespace

        & End Users / Public &
        Comfort, approachability &
        Understanding robot behavior &
        Trust, social fit &
        Privacy expectations &
        Adoption, daily use \\
        \addlinespace

        & Bystanders / Non-Users &
        Safe proximity, motion predictability &
        Intent interpretability, situational awareness &
        Comfort, norm compliance &
        Consent, exposure without opt-in &
        Passive coexistence \\
        \addlinespace\midrule

        \multirow{3}{*}{\parbox{2.2cm}{\centering\textbf{Governance, Risk,\\ \& Accountability}}} &
        Policy Makers / Regulators &
        Physical safety thresholds &
        Decision accountability &
        Social impact &
        Privacy, fairness, governance &
        Standards, regulation \\
        \addlinespace

        & Legal / Compliance Teams &
        Safety compliance evidence &
        Auditability, traceability &
        Regulatory communication &
        Accountability, legal responsibility &
        Compliance review, governance \\
        \addlinespace

        & Insurance / Risk Assessors &
        Failure severity, damage potential &
        Risk modeling, uncertainty quantification &
        Public risk perception &
        Liability attribution, harm mitigation &
        Coverage decisions, premium setting \\
        \addlinespace\midrule

        \multirow{2}{*}{\parbox{2.2cm}{\centering\textbf{Economic\\ \& Strategic}}} &
        
        Industry / Operators &
        Reliability, MTBF, maintenance &
        Robustness in deployment &
        User acceptance &
        Liability &
        Deployment readiness \\
        \addlinespace

        & Funding Agencies / VCs &
        Hardware feasibility &
        Scalability of intelligence &
        Market acceptance &
        Societal risk &
        Investment decisions \\
        \addlinespace

        \bottomrule
    \end{tabularx}
\end{sidewaystable}

\subsection{Rethinking Human Factors and Robotics Research}

\paragraph{Robots as experimental subjects, not idealized systems.}
Humanoid experiments systematically deviate from controlled laboratory assumptions due to actuator saturation, safety overrides, perception latency, thermal constraints, and hardware wear. Section~4 highlights how these factors influence observed behavior, particularly in timing-sensitive motion. Rather than treating such deviations as experimental noise, Humanoid Factors treats them as intrinsic properties of embodied systems that must be characterized, reported, and compared. This perspective aligns humanoid research more closely with Human Factors methodology, where variability, adaptation, and constraint-driven behavior are central objects of study rather than confounds to be eliminated.

\paragraph{From single-agent optimization to dual-agent coexistence studies.}
Classical Human Factors methods---task analysis, workload modeling, error taxonomies---were developed under the assumption of a single intelligent agent interacting with tools. In human--humanoid systems, these methods must be complemented by parallel models of humanoid capabilities, limitations, and failure modes. The four pillars of Humanoid Factors provide this structure, ensuring that humanoid constraints are specified with the same rigor as human constraints. Methodologically, this implies a shift toward \emph{joint optimization of interaction}. Task allocation, interface design, and safety envelopes should be formulated to account for co-adaptation over time: how humans learn a robot's behavioral regularities, and how the robot updates its models of user preferences, skill, and comfort. Short-horizon lab studies must therefore be augmented with longitudinal experiments that capture learning, trust calibration, and behavioral drift.

\paragraph{Toward shared experimental primitives.}
Human Factors matured through shared measurement primitives such as reaction time, error rate, and workload indices. Humanoid Factors similarly requires reusable experimental constructs that span motion, cognition, and interaction. Examples include motion legibility metrics grounded in human motor principles, measures of intervention appropriateness, ethical boundary violations, and recovery behavior following failure. Establishing such primitives is essential for cumulative progress, enabling results to build coherently over time rather than fragmenting into isolated, ad hoc case studies.

\subsection{Educating Humanoid Users}

\paragraph{Training users.}
Educating humanoid users becomes a central requirement as humanoids grow more autonomous and socially embedded. Ideally, users should be trained with explicit knowledge of a robot’s capabilities, limitations, and failure modes before prolonged interaction. The most robust approach involves hands-on exposure in which users experience not only successful behaviors, but also common failure cases and recovery strategies under supervision. Such experiential learning allows users to form accurate mental models of what the humanoid can and cannot do, how it signals uncertainty, and when human intervention is required. In practice, however, full training may not always be feasible. Even limited guidance—such as clear capability disclosures, success-rate statistics, demonstrations of edge cases, and standardized onboarding protocols—can substantially improve trust calibration. Without such guidance, early deployments risk placing the burden of discovering failure modes on users themselves, leading to misuse, over-reliance, or premature rejection of the technology.

We draw an analogy to licensing frameworks in other safety-critical domains. Just as automotive drivers are trained, tested, and certified before operating vehicles, future humanoid deployments may require lightweight ``humanoid literacy'' mechanisms: knowledge-based instruction combined with hands-on evaluation of basic interaction scenarios. While not all applications warrant formal certification, the underlying principle remains the same—safe and effective human--humanoid collaboration depends as much on informed users as on capable machines. By embedding user education into the Humanoid Factors framework, we ensure that responsibility for safe deployment does not rest solely on technical safeguards, but is shared across system design, governance, and human understanding. This reinforces the view of humanoids not as opaque intelligent artifacts, but as accountable participants in human-centered ecosystems.

\paragraph{Educating the public.}
Beyond direct users, broader public understanding of humanoid systems is important for shaping realistic expectations and long-term trust. Public education should emphasize not only what humanoids can do, but also their limitations, typical failure modes, and degrees of autonomy, helping to counteract exaggerated (whether positive or negative) perceptions driven by biased demonstrations or media narratives. Clear communication about how humanoids sense, decide, and act---and where responsibility resides when failures occur~\citep{FRANKLIN2021102252}---can reduce both unwarranted fear and misplaced confidence. Such transparency is especially important as humanoids enter shared social spaces, where indirect interaction and observation influence acceptance. By treating public education as a core component of the Humanoid Factors framework, we support societal trust calibration at scale and reinforce the view of humanoids as accountable, human-centered systems rather than opaque intelligent artifacts.

\subsection{The Role of Funding Agencies, Venture Investments, and Policy Makers}

\paragraph{Humanoid Factors as leading indicators of commercial risk.}
For industry and investors, Humanoid Factors reframes success criteria and provides a structured way to identify and mitigate non-obvious risks early in development. Impressive demonstrations of dexterity or autonomy are insufficient indicators of deployability. Commercial viability depends on whether humanoids can operate over extended periods without eroding trust, violating social norms, or requiring constant human supervision. Metrics such as intervention frequency, trust decay, recovery time, ergonomic compatibility, and ethical override rates become central indicators of readiness. Failures in social calibration, privacy handling, or motion legibility are not edge cases; they are systemic risks that can stall adoption, trigger regulatory backlash, or damage brand trust. Investing in Humanoid Factors research directly reduces downstream technical, legal, and reputational risk by surfacing these issues before large-scale deployment. Humanoid Factors also provide a principled framework for risk reduction that aligns directly with the priorities of public funding agencies increasingly focused on trustworthy AI, human-centered autonomy, and responsible deployment.


\paragraph{Standardization and public readiness.}
The absence of humanoid-specific standards creates uncertainty for deployment and adoption. Existing safety and ergonomics standards do not account for systems that combine human-like embodiment, adaptive learning, and social interaction. Humanoid Factors motivates new standards specifying acceptable ranges for motion timing, expressivity, data retention, and intervention behavior, providing clearer pathways from prototype to regulated deployment. To achieve this, a shared language for cross-sector coordination is required. Finally, as highlighted in Table~\ref{tab:stakeholder_analysis}, Humanoid Factors provides a common vocabulary for engineers, human factors researchers, policymakers, funding agencies, and investors to reason about humanoid systems. 

\section{Concluding Remarks}

Foundation-model-based humanoids increase the urgency of measuring not only \emph{what} robots achieve, but \emph{how} they behave in ways humans can predict and remain comfortable with. Our case study offered a deliberately narrow proof of concept: for one pointing/reaching task on one humanoid platform, a compact cognitive-science model (Fitts' Law) exposed failure modes that completion-centric robotics metrics obscured. While the Fitts' Law is a primitive for one of the simplest humanoid tasks, the broader lesson is that humanoid evaluation needs task-matched human-centered diagnostics alongside conventional robotics metrics. As humanoids transition from controlled laboratory environments into homes, workplaces, and public spaces, success will depend less on raw intelligence and more on how well humanoids fit---physically, cognitively, socially, and ethically---into human worlds. The Humanoid Factors framework provides an initial foundation for this transition, but its maturation will require sustained interdisciplinary collaboration, shared benchmarks, and long-term, real-world studies. Progress will demand close coordination across human factors, robotics, human–robot interaction, ethics, and policy, ensuring that advances in capability are matched by advances in understanding, accountability, and governance. We view this work as an initial step toward establishing Humanoid Factors as a core lens for designing, evaluating, and governing humanoid systems, one that prioritizes human compatibility alongside technical performance as humanoids become part of everyday life.



While the Humanoid Factors framework provides a structured lens for evaluating humanoids beyond raw task performance, it is not without limitations. Explicitly acknowledging these constraints is essential to prevent over-generalization, misinterpretation, or premature standardization.

\begin{enumerate}
    \item \textbf{Limited transferability of human-derived theories.}  
    HoF suggests adapting constructs and evaluation metrics from human factors, psychology, and ergonomics. These theories were developed for biological agents with shared embodiment, perception, and social learning histories, sometimes in controlled settings. Humanoid robots, even when anthropomorphic in form, may violate these assumptions, potentially leading to misleading conclusions if human benchmarks are applied without caution.

    \item \textbf{Risk of anthropomorphic overreach.}  
    HoF explicitly reasons about constructs such as trust, intent, legibility, and social norms. While useful, these constructs can encourage evaluators or users to ascribe human-like mental states to humanoids. Excessive anthropomorphism may obscure underlying failure modes rather than reveal them.

    \item \textbf{Terminological ambiguity across disciplines.}  
    Core HoF terms---including {trust}, {agency}, {autonomy}, {situational awareness}, and {safety}---carry distinct meanings across robotics, human factors, psychology, ethics, and law. Without careful operationalization, semantic inconsistency can weaken comparability across studies and introduce conceptual drift.

    \item \textbf{Measurement and proxy validity challenges.}  
    Many HoF indicators, such as trust decay, intervention appropriateness, recovery behavior, or ethical override rates, are necessarily indirect. These proxies are sensitive to task design, user population, and cultural context, and may not generalize across deployment settings. Poorly calibrated proxies risk creating a false sense of rigor.

    \item \textbf{Incomplete coverage of emergent behaviors.}  
    As humanoids increasingly rely on foundation models and long-horizon learning, novel and emergent behaviors~\citep{wei2022emergent} may arise that fall outside predefined HoF categories, reflecting both the development of non-human capabilities in humanoids and the human-centric origins of the HoF framework. Without continual refinement, the framework risks lagging behind system capabilities and deployment realities.

    \item \textbf{Danger of premature standardization.}  
    There is a risk that HoF metrics may be treated as compliance checklists rather than investigative tools. Early formalization may lock in incomplete assumptions and discourage exploration of alternative evaluation paradigms.
\end{enumerate}

\noindent
Overall, Humanoid Factors should be viewed not as a finalized standard, but as a living framework intended to surface risks, guide evaluation, and support interdisciplinary dialogue. Its effectiveness depends on continuous empirical grounding, careful operationalization, and ongoing refinement as humanoids mature. More broadly, Humanoid Factors represents an effort to shift humanoid design and evaluation beyond narrow task-completion metrics toward considerations of physical compatibility, cognitive alignment, social interaction, and ethical responsibility in real-world human environments.

\newpage

\newpage
\appendix

\section*{Appendix}
\section{Experimental Setup}

\subsection{Objective}
The primary objective of this experiment was to evaluate whether a Behavior Cloning (BC) policy, trained on human demonstrations, exhibits Fitts' Law behavior—specifically a linear speed-accuracy trade-off—comparable to the human demonstrator across varying reaching distances.

\subsection{Task and Apparatus}
The experimental platform consisted of a Unitree G1 Humanoid robot, using the 7-DOF left arm for manipulation tasks. The task involved performing discrete reaching movements from a neutral starting position to a static target box located on a wall in front of the robot. To simplify the setting, only 4 DOF were used in this experiment (left elbow, left shoulder yaw, left shoulder pitch, left shoulder roll), the wrist part and all other joints are fixed during the experiment.

The experiment was designed with four distance conditions (Amplitude, $A$), defined by the distance from the robot's start state to the target center:
\begin{itemize}
    \item \textbf{Condition 1:} $A = 20$ cm ($0.2$ m)
    \item \textbf{Condition 2:} $A = 30$ cm ($0.3$ m)
    \item \textbf{Condition 3:} $A = 40$ cm ($0.4$ m)
    \item \textbf{Condition 4:} $A = 50$ cm ($0.5$ m)
\end{itemize}

A fixed square target width ($W$) of $0.02$ m ($2$ cm) was used for all analysis. The success criterion for a trial was defined spatially: the robot's end-effector position must terminate within $1$ cm of the target center.

\subsection{Human Demonstrations and Dataset}
\paragraph{Human Demonstrations.} Ground truth data was collected via kinesthetic teaching (i.e., human moving the robot arm) at a frequency of 50 Hz. Approximately 25 valid trials were recorded for each distance condition.

\paragraph{Data format and joint selection.}
Demonstrations were stored as JSON files. Each file contained (i) metadata including the ordered list of available joint names, and (ii) a time-indexed list of frames, where each frame stored joint states. During preprocessing, we extracted the left-arm joint configuration sequence $q_{0:T-1}\in\mathbb{R}^{T\times J}$ by selecting the subset of joints present in the metadata.

To ensure robust parsing across recordings, we used a preferred joint ordering and selected whichever of these were present:
\{\texttt{LeftShoulderPitch}, \texttt{LeftShoulderRoll}, \texttt{LeftShoulderYaw}, \texttt{LeftElbow}\}.
In our reaching setup, this corresponded to $J=4$ actuated degrees of freedom, while all other joints were held fixed.

\subsection{Behavior Cloning Policy}
\paragraph{Behavior Cloning (BC).}
Behavior Cloning is an imitation learning method that learns a policy by supervised learning on state--action pairs extracted from expert demonstrations. Concretely, given a dataset of human demonstrations, BC trains a parametric policy $\pi_\theta$ to minimize the discrepancy between the policy output and the demonstrated control at each time step. In this work, the policy input is a short history window of joint configurations, and the policy output is the \emph{next} joint configuration command. Thus, BC reduces to a regression problem trained with a mean-squared error loss:
\begin{equation}
\theta^\star = \arg\min_\theta \; \mathbb{E}_{(x_t, y_t)\sim \mathcal{D}}\left[\lVert \pi_\theta(x_t) - y_t \rVert_2^2 \right].
\end{equation}
Importantly, we predict absolute joint angles $q$ directly (rather than joint increments $\Delta q$), which avoids drift from integrating predicted velocities over long horizons.

\subsubsection{Training targets and windowed supervision}
We trained BC using windowed state--action pairs constructed from each demonstration sequence. Let $H$ denote the history length in steps. For each time step $t \in \{H, \ldots, T-1\}$ we formed:
\begin{align}
x_t &= \mathrm{vec}\left([q_{t-H}, q_{t-H+1}, \ldots, q_{t-1}]\right) \in \mathbb{R}^{H\cdot J},\\
y_t &= q_t \in \mathbb{R}^{J}.
\end{align}
That is, the input is a flattened concatenation of the previous $H$ joint configurations, and the supervision target is the next absolute joint configuration.

\paragraph{Distance Conditioning (Context Feature)}
The training pipeline also appends a 1D distance context feature to each input. It is the reaching distance scaled by a user-defined factor $s$ and concatenated to the input:
\begin{equation}
\tilde{x}_t = [x_t; \; s\cdot d] \in \mathbb{R}^{H\cdot J + 1}.
\end{equation}
This provides the policy explicit information about which amplitude condition a sample belongs to, and can improve generalization across distances.

\paragraph{Normalization}
We standardized both inputs and outputs using statistics computed over the full training set. Let $\mu_x,\sigma_x$ be the mean and standard deviation of inputs, and $\mu_y,\sigma_y$ for targets. We applied z-score normalization:
\begin{align}
x_t^{(n)} &= (x_t - \mu_x)/(\sigma_x + \epsilon),\\
y_t^{(n)} &= (y_t - \mu_y)/(\sigma_y + \epsilon),
\end{align}
with $\epsilon=10^{-8}$ for numerical stability. The policy predicted normalized targets $\hat{y}_t^{(n)}$, which were de-normalized at inference time:
\begin{equation}
\hat{y}_t = \hat{y}_t^{(n)}\odot \sigma_y + \mu_y.
\end{equation}
Normalization parameters were saved alongside the trained model to ensure consistent evaluation.

\subsubsection{Model architecture and optimization}
The BC policy was implemented as a feed-forward multilayer perceptron (MLP) with an input dimension of $H \cdot J + 1$. The architecture comprises two fully-connected hidden layers, each containing 256 units and followed by ReLU activation functions. The final stage is a linear output layer that produces a vector in $\mathbb{R}^{J}$, corresponding to the next joint configuration, with dropout included as an optional parameter (defaulting to $0.0$).

We optimized mean-squared error (MSE) on normalized targets:
\begin{equation}
\mathcal{L}(\theta) = \frac{1}{B}\sum_{i=1}^{B} \left\lVert \pi_\theta(x_i^{(n)}) - y_i^{(n)} \right\rVert_2^2,
\end{equation}
where $B$ is the batch size.

The model was optimized using AdamW with a learning rate of $10^{-3}$ and weight decay of $10^{-6}$, employing global gradient norm clipping to ensure stability. Learning rate adjustment was managed by a ReduceLROnPlateau scheduler monitoring validation loss, while an early stopping mechanism halted training after 10 epochs without improvement.

The dataset was split into a random 80/20 train-validation split. Training was conducted with a batch size of 256 for up to 100 epochs. To address potential imbalances in data collection across different reaching distances, we employed a WeightedRandomSampler that reweighted training samples inversely proportional to the frequency of their specific distance condition, ensuring the policy remained robust across all amplitude targets.

\subsection{Policy Evaluation: BC Rollouts in Simulation}
The trained policy was evaluated in a simulated environment using the MuJoCo physics engine. For each trial:
\begin{enumerate}
    \item The robot was initialized with the first $0.2$ seconds of the corresponding human demonstration to establish a consistent start state.
    \item The policy ran autoregressively, predicting the next joint angles ($\mathbf{q}_{t+1}$) based on the history of observations.
    \item Forward Kinematics (FK) were computed at every time step to monitor the end-effector position.
    \item The rollout concluded when the end-effector successfully entered the target zone or if a timeout occurred (defined as $2 \times T_{human} + 50$ steps).
\end{enumerate}

\subsection{Metrics and Analysis}
The primary metric for analysis was \textbf{Movement Time ($MT$)}.
\begin{itemize}
    \item For \textbf{Human} trials, $MT$ was defined as the duration between velocity onset ($>0.05$ rad/s) and offset.
    \item For \textbf{Robot} trials, $MT$ was the time taken to successfully enter the target zone (success steps / 50 Hz).
\end{itemize}

The \textbf{Index of Difficulty ($ID$)} was calculated according to the formulation of Fitts' Law:
\begin{equation}
    ID = \log_2\left(\frac{2A}{W}\right)
\end{equation}
where $A$ is the target amplitude (distance) and $W$ is the target width.

To ensure the analysis reflected skilled motor control rather than failure recovery, outlier removal was applied.

\subsection{Statistical Analysis}
We performed a linear regression analysis ($MT = a + b \cdot ID$) to determine the fit of the data to Fitts' Law. The coefficient of determination ($R^2$) indicated the goodness of fit, and the slope ($b$) represented the information processing rate (seconds per bit).

Additionally, an Analysis of Variance (ANOVA) was conducted to test the significance of the regression ($p < 0.05$), and a Lack-of-Fit test was performed to compare the linear model against a saturated model (group means), checking for significant deviations from linearity at specific distances.


\begin{thebibliography}{}

\bibitem[Abadi et~al., 2016]{abadi2016deep}
Abadi, M., Chu, A., Goodfellow, I., McMahan, H.~B., Mironov, I., Talwar, K., and Zhang, L. (2016).
\newblock Deep learning with differential privacy.
\newblock In {\em Proceedings of the 2016 ACM SIGSAC conference on computer and communications security}, pages 308--318.

\bibitem[Abdin et~al., 2024]{abdin2024phi}
Abdin, M., Aneja, J., Behl, H., Bubeck, S., Eldan, R., Gunasekar, S., Harrison, M., Hewett, R.~J., Javaheripi, M., Kauffmann, P., et~al. (2024).
\newblock Phi-4 technical report.
\newblock {\em arXiv preprint arXiv:2412.08905}.

\bibitem[Admoni and Scassellati, 2017]{admoni2017social}
Admoni, H. and Scassellati, B. (2017).
\newblock Social eye gaze in human-robot interaction: a review.
\newblock {\em Journal of Human-Robot Interaction}, 6(1):25--63.

\bibitem[Agia et~al., 2024]{agia2024unpacking}
Agia, C., Sinha, R., Yang, J., Cao, Z.-a., Antonova, R., Pavone, M., and Bohg, J. (2024).
\newblock Unpacking failure modes of generative policies: Runtime monitoring of consistency and progress.
\newblock {\em arXiv preprint arXiv:2410.04640}.

\bibitem[Ajoudani et~al., 2018]{ajoudani2018progress}
Ajoudani, A., Zanchettin, A.~M., Ivaldi, S., Albu-Sch{\"a}ffer, A., Kosuge, K., and Khatib, O. (2018).
\newblock Progress and prospects of the human--robot collaboration.
\newblock {\em Autonomous robots}, 42(5):957--975.

\bibitem[Alshiekh et~al., 2018]{alshiekh2018safe}
Alshiekh, M., Bloem, R., Ehlers, R., K{\"o}nighofer, B., Niekum, S., and Topcu, U. (2018).
\newblock Safe reinforcement learning via shielding.
\newblock In {\em Proceedings of the AAAI conference on artificial intelligence}, volume~32.

\bibitem[Anjomshoae et~al., 2019]{anjomshoae2019explainable}
Anjomshoae, S., Najjar, A., Calvaresi, D., and Fr{\"a}mling, K. (2019).
\newblock Explainable agents and robots: Results from a systematic literature review.
\newblock In {\em 18th International Conference on Autonomous Agents and Multiagent Systems (AAMAS 2019), Montreal, Canada, May 13--17, 2019}, pages 1078--1088. International Foundation for Autonomous Agents and Multiagent Systems.

\bibitem[Bai et~al., 2022]{bai2022constitutional}
Bai, Y., Kadavath, S., Kundu, S., Askell, A., Kernion, J., Jones, A., Chen, A., Goldie, A., Mirhoseini, A., McKinnon, C., et~al. (2022).
\newblock Constitutional ai: Harmlessness from ai feedback.
\newblock {\em arXiv preprint arXiv:2212.08073}.

\bibitem[Barfield, 2024]{Barfield_2024}
Barfield, J. (2024).
\newblock {\em Robot Ethics for Interaction with Humanoid, AI-Enabled, and Expressive Robots}, page 393–407.
\newblock Cambridge Law Handbooks. Cambridge University Press.

\bibitem[Barnes, 2008]{barnes2008cognitive}
Barnes, G.~R. (2008).
\newblock Cognitive processes involved in smooth pursuit eye movements.
\newblock {\em Brain and cognition}, 68(3):309--326.

\bibitem[Barocas et~al., 2023]{barocas2023fairness}
Barocas, S., Hardt, M., and Narayanan, A. (2023).
\newblock {\em Fairness and machine learning: Limitations and opportunities}.
\newblock MIT press.

\bibitem[Baron-Cohen et~al., 2001]{baron2001reading}
Baron-Cohen, S., Wheelwright, S., Hill, J., Raste, Y., and Plumb, I. (2001).
\newblock The “reading the mind in the eyes” test revised version: A study with normal adults, and adults with asperger syndrome or high-functioning autism.
\newblock {\em Journal of child psychology and psychiatry}, 42(2):241--251.

\bibitem[Barreiros et~al., 2025]{barreiros2025careful}
Barreiros, J., Beaulieu, A., Bhat, A., Cory, R., Cousineau, E., Dai, H., Fang, C.-H., Hashimoto, K., Irshad, M.~Z., Itkina, M., et~al. (2025).
\newblock A careful examination of large behavior models for multitask dexterous manipulation.
\newblock {\em arXiv preprint arXiv:2507.05331}.

\bibitem[Bartneck et~al., 2018]{bartneck2018robots}
Bartneck, C., Yogeeswaran, K., Ser, Q.~M., Woodward, G., Sparrow, R., Wang, S., and Eyssel, F. (2018).
\newblock Robots and racism.
\newblock In {\em Proceedings of the 2018 ACM/IEEE international conference on human-robot interaction}, pages 196--204.

\bibitem[Bianchi et~al., 2024]{bianchi2024well}
Bianchi, F., Chia, P.~J., Yuksekgonul, M., Tagliabue, J., Jurafsky, D., and Zou, J. (2024).
\newblock How well can llms negotiate? negotiationarena platform and analysis.
\newblock {\em arXiv preprint arXiv:2402.05863}.

\bibitem[Biddle, 2013]{biddle2013role}
Biddle, B.~J. (2013).
\newblock {\em Role theory: Expectations, identities, and behaviors}.
\newblock Academic press.

\bibitem[Biyik, 2022]{biyik2022learning}
Biyik, E. (2022).
\newblock {\em Learning preferences for interactive autonomy}.
\newblock PhD thesis, Stanford University.

\bibitem[Bjorck et~al., 2025]{bjorck2025gr00t}
Bjorck, J., Casta{\~n}eda, F., Cherniadev, N., Da, X., Ding, R., Fan, L., Fang, Y., Fox, D., Hu, F., Huang, S., et~al. (2025).
\newblock Gr00t n1: An open foundation model for generalist humanoid robots.
\newblock {\em arXiv preprint arXiv:2503.14734}.

\bibitem[Black et~al., 2024]{black2024pi_0}
Black, K., Brown, N., Driess, D., Esmail, A., Equi, M., Finn, C., Fusai, N., Groom, L., Hausman, K., Ichter, B., Jakubczak, S., Jones, T., Ke, L., Levine, S., Li-Bell, A., Mothukuri, M., Nair, S., Pertsch, K., Shi, L.~X., Tanner, J., Vuong, Q., Walling, A., Wang, H., and Zhilinsky, U. (2024).
\newblock $\pi$0: A vision-language-action flow model for general robot control.
\newblock {\em ArXiv}, abs/2410.24164.

\bibitem[Bommasani et~al., 2021]{bommasani2021opportunities}
Bommasani, R., Hudson, D.~A., Adeli, E., Altman, R., Arora, S., von Arx, S., Bernstein, M.~S., Bohg, J., Bosselut, A., Brunskill, E., Brynjolfsson, E., Buch, S., Card, D., Castellon, R., Chatterji, N.~S., Chen, A.~S., Creel, K.~A., Davis, J., Demszky, D., Donahue, C., Doumbouya, M., Durmus, E., Ermon, S., Etchemendy, J., Ethayarajh, K., Fei-Fei, L., Finn, C., Gale, T., Gillespie, L.~E., Goel, K., Goodman, N.~D., Grossman, S., Guha, N., Hashimoto, T., Henderson, P., Hewitt, J., Ho, D.~E., Hong, J., Hsu, K., Huang, J., Icard, T.~F., Jain, S., Jurafsky, D., Kalluri, P., Karamcheti, S., Keeling, G., Khani, F., Khattab, O., Koh, P.~W., Krass, M.~S., Krishna, R., Kuditipudi, R., Kumar, A., Ladhak, F., Lee, M., Lee, T., Leskovec, J., Levent, I., Li, X.~L., Li, X., Ma, T., Malik, A., Manning, C.~D., Mirchandani, S.~P., Mitchell, E., Munyikwa, Z., Nair, S., Narayan, A., Narayanan, D., Newman, B., Nie, A., Niebles, J.~C., Nilforoshan, H., Nyarko, J.~F., Ogut, G., Orr, L., Papadimitriou, I., Park, J.~S., Piech, C.,
  Portelance, E., Potts, C., Raghunathan, A., Reich, R., Ren, H., Rong, F., Roohani, Y.~H., Ruiz, C., Ryan, J., R'e, C., Sadigh, D., Sagawa, S., Santhanam, K., Shih, A., Srinivasan, K.~P., Tamkin, A., Taori, R., Thomas, A.~W., Tram{\`e}r, F., Wang, R.~E., Wang, W., Wu, B., Wu, J., Wu, Y., Xie, S.~M., Yasunaga, M., You, J., Zaharia, M.~A., Zhang, M., Zhang, T., Zhang, X., Zhang, Y., Zheng, L., Zhou, K., and Liang, P. (2021).
\newblock On the opportunities and risks of foundation models.
\newblock {\em ArXiv}.

\bibitem[Breazeal, 2003]{breazeal2003toward}
Breazeal, C. (2003).
\newblock Toward sociable robots.
\newblock {\em Robotics and autonomous systems}, 42(3-4):167--175.

\bibitem[Brohan et~al., 2022]{brohan2022rt}
Brohan, A., Brown, N., Carbajal, J., Chebotar, Y., Dabis, J., Finn, C., Gopalakrishnan, K., Hausman, K., Herzog, A., Hsu, J., et~al. (2022).
\newblock Rt-1: Robotics transformer for real-world control at scale.
\newblock {\em arXiv preprint arXiv:2212.06817}.

\bibitem[Brunet et~al., 2000]{Brunet2000API}
Brunet, E., Sarfati, Y., Hardy-Bayl{\'e}, M.-C., and Decety, J. (2000).
\newblock A pet investigation of the attribution of intentions with a nonverbal task.
\newblock {\em NeuroImage}, 11:157--166.

\bibitem[Brunke et~al., 2022]{brunke2022safe}
Brunke, L., Greeff, M., Hall, A.~W., Yuan, Z., Zhou, S., Panerati, J., and Schoellig, A.~P. (2022).
\newblock Safe learning in robotics: From learning-based control to safe reinforcement learning.
\newblock {\em Annual Review of Control, Robotics, and Autonomous Systems}, 5(1):411--444.

\bibitem[Buolamwini and Gebru, 2018]{buolamwini2018gender}
Buolamwini, J. and Gebru, T. (2018).
\newblock Gender shades: Intersectional accuracy disparities in commercial gender classification.
\newblock In {\em Conference on fairness, accountability and transparency}, pages 77--91. PMLR.

\bibitem[Cao, 2025]{cao2025humanoid}
Cao, L. (2025).
\newblock Humanoid robots and humanoid ai: Review, perspectives and directions.
\newblock {\em ACM Comput. Surv.}, 58(4).

\bibitem[Chatzimichali et~al., 2020]{chatzimichali2020toward}
Chatzimichali, A., Harrison, R., and Chrysostomou, D. (2020).
\newblock Toward privacy-sensitive human--robot interaction: Privacy terms and human--data interaction in the personal robot era.
\newblock {\em Paladyn, Journal of Behavioral Robotics}, 12(1):160--174.

\bibitem[Chennabasappa et~al., 2025]{chennabasappa2025llamafirewall}
Chennabasappa, S., Nikolaidis, C., Song, D., Molnar, D., Ding, S., Wan, S., Whitman, S., Deason, L., Doucette, N., Montilla, A., et~al. (2025).
\newblock Llamafirewall: An open source guardrail system for building secure ai agents.
\newblock {\em arXiv preprint arXiv:2505.03574}.

\bibitem[Chi et~al., 2025]{chi2025diffusion}
Chi, C., Xu, Z., Feng, S., Cousineau, E., Du, Y., Burchfiel, B., Tedrake, R., and Song, S. (2025).
\newblock Diffusion policy: Visuomotor policy learning via action diffusion.
\newblock {\em The International Journal of Robotics Research}, 44(10-11):1684--1704.

\bibitem[Christiano et~al., 2017]{christiano2017deep}
Christiano, P.~F., Leike, J., Brown, T., Martic, M., Legg, S., and Amodei, D. (2017).
\newblock Deep reinforcement learning from human preferences.
\newblock {\em Advances in neural information processing systems}, 30.

\bibitem[Clark, 1996]{clark1996using}
Clark, H.~H. (1996).
\newblock {\em Using language}.
\newblock Cambridge university press.

\bibitem[Cooke et~al., 2020]{10.1007/978-3-030-49183-3_11}
Cooke, N., Demir, M., and Huang, L. (2020).
\newblock A framework for human-autonomy team research.
\newblock In {\em Engineering Psychology and Cognitive Ergonomics. Cognition and Design: 17th International Conference, EPCE 2020, Held as Part of the 22nd HCI International Conference, HCII 2020, Copenhagen, Denmark, July 19–24, 2020, Proceedings, Part II}, page 134–146, Berlin, Heidelberg. Springer-Verlag.

\bibitem[Corso and Kochenderfer, 2020]{corso2020interpretable}
Corso, A. and Kochenderfer, M.~J. (2020).
\newblock Interpretable safety validation for autonomous vehicles.
\newblock In {\em 2020 IEEE 23rd International Conference on Intelligent Transportation Systems (ITSC)}, pages 1--6. IEEE.

\bibitem[Darling, 2016]{darling2016extending}
Darling, K. (2016).
\newblock Extending legal protection to social robots: The effects of anthropomorphism, empathy, and violent behavior towards robotic objects.
\newblock In {\em Robot law}, pages 213--232. Edward Elgar Publishing.

\bibitem[Dautenhahn, 2007]{dautenhahn2007socially}
Dautenhahn, K. (2007).
\newblock Socially intelligent robots: dimensions of human--robot interaction.
\newblock {\em Philosophical transactions of the royal society B: Biological sciences}, 362(1480):679--704.

\bibitem[DeChant, 2025]{dechant2025episodic}
DeChant, C. (2025).
\newblock Episodic memory in ai agents poses risks that should be studied and mitigated.
\newblock In {\em 2025 IEEE Conference on Secure and Trustworthy Machine Learning (SaTML)}, pages 321--332. IEEE.

\bibitem[Dinh et~al., 2016]{dinh2016density}
Dinh, L., Sohl-Dickstein, J., and Bengio, S. (2016).
\newblock Density estimation using real nvp.
\newblock {\em arXiv preprint arXiv:1605.08803}.

\bibitem[Dragan et~al., 2013]{dragan2013legibility}
Dragan, A.~D., Lee, K.~C., and Srinivasa, S.~S. (2013).
\newblock Legibility and predictability of robot motion.
\newblock In {\em 2013 8th ACM/IEEE International Conference on Human-Robot Interaction (HRI)}, pages 301--308. IEEE.

\bibitem[Driess et~al., 2023]{driess2023palm}
Driess, D., Xia, F., Sajjadi, M. S.~M., Lynch, C., Chowdhery, A., Ichter, B., Wahid, A., Tompson, J., Vuong, Q., Yu, T., Huang, W., Chebotar, Y., Sermanet, P., Duckworth, D., Levine, S., Vanhoucke, V., Hausman, K., Toussaint, M., Greff, K., Zeng, A., Mordatch, I., and Florence, P. (2023).
\newblock PaLM-E: an embodied multimodal language model.
\newblock In {\em Proceedings of the 40th International Conference on Machine Learning}, ICML'23. JMLR.org.

\bibitem[Du et~al., 2025]{du2025flexible}
Du, X., Ye, Y., Jiao, B., Kong, Y., Yu, L., Liu, R., Yun, S., Lu, D., Qiao, J., Liu, Z., et~al. (2025).
\newblock A flexible thermal management method for high-power chips in humanoid robots.
\newblock {\em Device}, 3(2).

\bibitem[Dunlosky and Metcalfe, 2008]{dunlosky2008metacognition}
Dunlosky, J. and Metcalfe, J. (2008).
\newblock {\em Metacognition}.
\newblock Sage Publications.

\bibitem[Durkheim, 1982]{durkheim1982rules}
Durkheim, {\'E}. (1982).
\newblock {\em The Rules of Sociological Method}.
\newblock The Free Press, New York.
\newblock Original work published 1895.

\bibitem[Endsley, 2017]{endsley2017toward}
Endsley, M.~R. (2017).
\newblock Toward a theory of situation awareness in dynamic systems.
\newblock In {\em Situational awareness}, pages 9--42. Routledge.

\bibitem[Endsley, 2021]{endsley2021situation}
Endsley, M.~R. (2021).
\newblock Situation awareness.
\newblock {\em Handbook of human factors and ergonomics}, pages 434--455.

\bibitem[Endsley et~al., 2000]{endsley2000theoretical}
Endsley, M.~R., Garland, D.~J., et~al. (2000).
\newblock Theoretical underpinnings of situation awareness: A critical review.
\newblock {\em Situation awareness analysis and measurement}, 1(1):3--21.

\bibitem[Ficuciello et~al., 2015]{ficuciello2015variable}
Ficuciello, F., Villani, L., and Siciliano, B. (2015).
\newblock Variable impedance control of redundant manipulators for intuitive human--robot physical interaction.
\newblock {\em IEEE Transactions on Robotics}, 31(4):850--863.

\bibitem[{Figure AI}, 2025]{figure2025helix}
{Figure AI} (2025).
\newblock Helix: A vision-language-action model for generalist humanoid robots.
\newblock \url{https://www.figure.ai/news/helix}.
\newblock Accessed: 2025-11-23.

\bibitem[Fink, 2012]{fink2012anthropomorphism}
Fink, J. (2012).
\newblock Anthropomorphism and human likeness in the design of robots and human-robot interaction.
\newblock In {\em International conference on social robotics}, pages 199--208. Springer.

\bibitem[Fitts, 1954]{fitts1954information}
Fitts, P.~M. (1954).
\newblock The information capacity of the human motor system in controlling the amplitude of movement.
\newblock {\em Journal of experimental psychology}, 47(6):381.

\bibitem[Fong et~al., 2003]{fong2003survey}
Fong, T., Nourbakhsh, I., and Dautenhahn, K. (2003).
\newblock A survey of socially interactive robots.
\newblock {\em Robotics and autonomous systems}, 42(3-4):143--166.

\bibitem[Franklin et~al., 2021]{FRANKLIN2021102252}
Franklin, M., Awad, E., and Lagnado, D. (2021).
\newblock Blaming automated vehicles in difficult situations.
\newblock {\em iScience}, 24(4):102252.

\bibitem[Fu et~al., 2023]{fu2023robot}
Fu, D., Abawi, F., and Wermter, S. (2023).
\newblock The robot in the room: influence of robot facial expressions and gaze on human-human-robot collaboration.
\newblock In {\em 2023 32nd IEEE International Conference on Robot and Human Interactive Communication (RO-MAN)}, pages 85--91. IEEE.

\bibitem[Fu et~al., 2024]{fu2024mobile}
Fu, Z., Zhao, T.~Z., and Finn, C. (2024).
\newblock Mobile aloha: Learning bimanual mobile manipulation with low-cost whole-body teleoperation.
\newblock {\em arXiv preprint arXiv:2401.02117}.

\bibitem[Gallegos et~al., 2024]{gallegos2024bias}
Gallegos, I.~O., Rossi, R.~A., Barrow, J., Tanjim, M.~M., Kim, S., Dernoncourt, F., Yu, T., Zhang, R., and Ahmed, N.~K. (2024).
\newblock Bias and fairness in large language models: A survey.
\newblock {\em Computational Linguistics}, 50(3):1097--1179.

\bibitem[Gan and Hoffmann, 1988]{gan1988geometrical}
Gan, K.-C. and Hoffmann, E.~R. (1988).
\newblock Geometrical conditions for ballistic and visually controlled movements.
\newblock {\em Ergonomics}, 31(5):829--839.

\bibitem[{Gemini Robotics Team} et~al., 2025a]{team2025gemini1_5}
{Gemini Robotics Team}, Abdolmaleki, A., Abeyruwan, S., Ainslie, J., Alayrac, J.-B., Arenas, M.~G., Balakrishna, A., Batchelor, N., Bewley, A., Bingham, J., et~al. (2025a).
\newblock {Gemini Robotics 1.5: Pushing the Frontier of Generalist Robots with Advanced Embodied Reasoning, Thinking, and Motion Transfer}.
\newblock {\em arXiv preprint arXiv:2510.03342}.

\bibitem[{Gemini Robotics Team} et~al., 2025b]{team2025gemini}
{Gemini Robotics Team}, Abeyruwan, S., Ainslie, J., Alayrac, J.-B., Arenas, M.~G., Armstrong, T., Balakrishna, A., Baruch, R., Bauza, M., Blokzijl, M., et~al. (2025b).
\newblock {Gemini Robotics: Bringing AI into the Physical World}.
\newblock {\em arXiv preprint arXiv:2503.20020}.

\bibitem[Goodfellow et~al., 2020]{goodfellow2020generative}
Goodfellow, I., Pouget-Abadie, J., Mirza, M., Xu, B., Warde-Farley, D., Ozair, S., Courville, A., and Bengio, Y. (2020).
\newblock Generative adversarial networks.
\newblock {\em Communications of the ACM}, 63(11):139--144.

\bibitem[Gouaillier et~al., 2008]{gouaillier2008nao}
Gouaillier, D., Hugel, V., Blazevic, P., Kilner, C., Monceaux, J., Lafourcade, P., Marnier, B., Serre, J., and Maisonnier, B. (2008).
\newblock The nao humanoid: a combination of performance and affordability.
\newblock {\em arXiv preprint arXiv:0807.3223}.

\bibitem[Gui et~al., 2024]{gui2024survey}
Gui, J., Chen, T., Zhang, J., Cao, Q., Sun, Z., Luo, H., and Tao, D. (2024).
\newblock A survey on self-supervised learning: Algorithms, applications, and future trends.
\newblock {\em IEEE Transactions on Pattern Analysis and Machine Intelligence}, 46(12):9052--9071.

\bibitem[Gundawar et~al., 2025]{gundawar2025pac}
Gundawar, A., Sagar, S., and Senanayake, R. (2025).
\newblock {PAC} bench: Do foundation models understand prerequisites for executing manipulation policies?
\newblock In {\em The Thirty-ninth Annual Conference on Neural Information Processing Systems Datasets and Benchmarks Track}.

\bibitem[Hancock et~al., 2011]{hancock2011meta}
Hancock, P.~A., Billings, D.~R., Schaefer, K.~E., Chen, J.~Y., De~Visser, E.~J., and Parasuraman, R. (2011).
\newblock A meta-analysis of factors affecting trust in human-robot interaction.
\newblock {\em Human factors}, 53(5):517--527.

\bibitem[{Hanson Robotics}, 2024]{HansonRoboticsSophia}
{Hanson Robotics} (2024).
\newblock Sophia.
\newblock Accessed: 2026-02-05.

\bibitem[Harvey and Bradford, 2024]{harvey2024senses}
Harvey, A. and Bradford, A. (2024).
\newblock The 5 human senses --- and a few more you might not know about.
\newblock Accessed: 2026-02-04.

\bibitem[He et~al., 2025]{he2025latent}
He, C., Camps, G.~S., Liu, X., Schwager, M., and Sartoretti, G. (2025).
\newblock Latent theory of mind: A decentralized diffusion architecture for cooperative manipulation.
\newblock {\em arXiv preprint arXiv:2505.09144}.

\bibitem[Henschke and Pakan, 2023]{henschke2023engaging}
Henschke, J.~U. and Pakan, J.~M. (2023).
\newblock Engaging distributed cortical and cerebellar networks through motor execution, observation, and imagery.
\newblock {\em Frontiers in systems neuroscience}, 17:1165307.

\bibitem[Ho and Ermon, 2016]{ho2016generative}
Ho, J. and Ermon, S. (2016).
\newblock Generative adversarial imitation learning.
\newblock {\em Advances in neural information processing systems}, 29.

\bibitem[Ho et~al., 2020]{ho2020denoising}
Ho, J., Jain, A., and Abbeel, P. (2020).
\newblock Denoising diffusion probabilistic models.
\newblock {\em Advances in neural information processing systems}, 33:6840--6851.

\bibitem[Hoogeboom et~al., 2021]{hoogeboom2021autoregressive}
Hoogeboom, E., Gritsenko, A.~A., Bastings, J., Poole, B., Berg, R. v.~d., and Salimans, T. (2021).
\newblock Autoregressive diffusion models.
\newblock {\em arXiv preprint arXiv:2110.02037}.

\bibitem[Howell et~al., 2023]{howell2023comfortDynamics}
Howell, P., Kolb, J., Liu, Y., and Ravichandar, H. (2023).
\newblock The effects of robot motion on comfort dynamics of novice users in close-proximity human-robot interaction.
\newblock {\em arXiv preprint arXiv:2308.01466}.

\bibitem[Hu, 2023]{hu2023research}
Hu, M. (2023).
\newblock Research on safety design and optimization of collaborative robots.
\newblock {\em International Journal of Intelligent Robotics and Applications}, 7(4):795--809.

\bibitem[Huang et~al., 2023]{huang2023egocentric}
Huang, C., Tian, Y., Kumar, A., and Xu, C. (2023).
\newblock Egocentric audio-visual object localization.
\newblock In {\em Proceedings of the IEEE/CVF conference on computer vision and pattern recognition}, pages 22910--22921.

\bibitem[Huang et~al., 2015]{huang2015adaptive}
Huang, C.-M., Cakmak, M., and Mutlu, B. (2015).
\newblock Adaptive coordination strategies for human-robot handovers.
\newblock In {\em Robotics: science and systems}, volume~11, pages 1--10. Rome, Italy.

\bibitem[Huang et~al., 2025]{huang2025establishing}
Huang, L., Freeman, J., Cooke, N.~J., Cohen, M.~C., Yin, X., Clark, J., Wood, M., Buchanan, V., Corral, C., Scholcover, F., et~al. (2025).
\newblock Establishing human observer criterion in evaluating artificial social intelligence agents in a search and rescue task.
\newblock {\em Topics in Cognitive Science}, 17(2):349--373.

\bibitem[Huang et~al., 2017]{huang2017adversarial}
Huang, S., Papernot, N., Goodfellow, I., Duan, Y., and Abbeel, P. (2017).
\newblock Adversarial attacks on neural network policies.
\newblock {\em arXiv preprint arXiv:1702.02284}.

\bibitem[{Human Factors and Ergonomics Society}, 2021]{ansi_hfes_400_2021}
{Human Factors and Ergonomics Society} (2021).
\newblock Human readiness level scale in the system development process.
\newblock Standard ANSI/HFES 400-2021, Human Factors and Ergonomics Society, Washington, D.C.

\bibitem[{Human Factors and Ergonomics Society}, nd]{HFES_Definition}
{Human Factors and Ergonomics Society} (n.d.).
\newblock What is human factors and ergonomics?
\newblock Accessed: 2026-02-04.

\bibitem[{International Electrotechnical Commission}, 1999]{IEC60529:1999}
{International Electrotechnical Commission} (1999).
\newblock Degrees of protection provided by enclosures (ip code).

\bibitem[Kantowitz and Elvers, 1988]{kantowitz1988fitts}
Kantowitz, B.~H. and Elvers, G.~C. (1988).
\newblock Fitts’ law with an isometric controller: effects of order of control and controldisplay gain.
\newblock {\em Journal of Motor Behavior}, 20(1):53--66.

\bibitem[Kaufmann and Cl{\'e}ment, 2007]{kaufmann2007culture}
Kaufmann, L. and Cl{\'e}ment, F. (2007).
\newblock How culture comes to mind: From social affordances to cultural analogies.
\newblock {\em Intellectica}, 46:221--250.

\bibitem[Kim et~al., 2024]{kim2024openvla}
Kim, M.~J., Pertsch, K., Karamcheti, S., Xiao, T., Balakrishna, A., Nair, S., Rafailov, R., Foster, E., Lam, G., Sanketi, P., et~al. (2024).
\newblock Openvla: An open-source vision-language-action model.
\newblock {\em arXiv preprint arXiv:2406.09246}.

\bibitem[Kim and Kim, 2013]{Kim2013HumanoidRA}
Kim, M.-S. and Kim, E.-J. (2013).
\newblock Humanoid robots as “the cultural other”: are we able to love our creations?
\newblock {\em AI \& SOCIETY}, 28:309--318.

\bibitem[Kingma and Welling, 2013]{kingma2013auto}
Kingma, D.~P. and Welling, M. (2013).
\newblock Auto-encoding variational bayes.
\newblock {\em arXiv preprint arXiv:1312.6114}.

\bibitem[Kochenderfer, 2015]{kochenderfer2015decision}
Kochenderfer, M.~J. (2015).
\newblock {\em Decision making under uncertainty: theory and application}.
\newblock MIT press.

\bibitem[Kopnarski et~al., 2023]{kopnarski2023systematic}
Kopnarski, L., Rudisch, J., and Voelcker-Rehage, C. (2023).
\newblock A systematic review of handover actions in human dyads.
\newblock {\em Frontiers in Psychology}, 14:1147296.

\bibitem[Kóczi and Sárosi, 2025]{electronics14234734}
Kóczi, D. and Sárosi, J. (2025).
\newblock Safety engineering for humanoid robots in everyday life—scoping review.
\newblock {\em Electronics}, 14(23).

\bibitem[Large et~al., 2015]{large2015predicting}
Large, D.~R., Crundall, E., Burnett, G., and Skrypchuk, L. (2015).
\newblock Predicting the visual demand of finger-touch pointing tasks in a driving context.
\newblock In {\em Proceedings of the 7th International Conference on Automotive User Interfaces and Interactive Vehicular Applications}, pages 221--224.

\bibitem[Lasota et~al., 2017]{lasota2017survey}
Lasota, P.~A., Fong, T., Shah, J.~A., et~al. (2017).
\newblock A survey of methods for safe human-robot interaction.
\newblock {\em Foundations and Trends{\textregistered} in Robotics}, 5(4):261--349.

\bibitem[Lee et~al., 2025]{lee2025molmoact}
Lee, J., Duan, J., Fang, H., Deng, Y., Liu, S., Li, B., Fang, B., Zhang, J., Wang, Y.~R., Lee, S., et~al. (2025).
\newblock Molmoact: Action reasoning models that can reason in space.
\newblock {\em arXiv preprint arXiv:2508.07917}.

\bibitem[Li et~al., 2023]{li2022humanlikeMotionPlanningRobotArms}
Li, S., Qi, W., Hu, Y., Karimi, H.~R., and Ferrigno, G. (2023).
\newblock Human-like motion planning of robotic arms based on human arm motion patterns.
\newblock {\em Robotica}, 41(1):259–276.

\bibitem[Li and Wang, 2024]{li2024comprehensive}
Li, Y. and Wang, L. (2024).
\newblock Comprehensive research and analysis on obstacle--singularity--joint limit avoidance of redundant robot.
\newblock {\em International Journal of Advanced Robotic Systems}, 21(1):17298806241233910.

\bibitem[Lienig and Bruemmer, 2017]{Lienig2017Reliability}
Lienig, J. and Bruemmer, H. (2017).
\newblock {\em Reliability Analysis}.
\newblock Springer International Publishing.

\bibitem[Liu et~al., 2024]{liu2024rdt}
Liu, S., Wu, L., Li, B., Tan, H., Chen, H., Wang, Z., Xu, K., Su, H., and Zhu, J. (2024).
\newblock Rdt-1b: a diffusion foundation model for bimanual manipulation.
\newblock {\em arXiv preprint arXiv:2410.07864}.

\bibitem[Liu et~al., 2025]{liu2025k2}
Liu, Z., Tang, L., Jin, L., Li, H., Ranjan, N., Fan, D., Rohatgi, S., Fan, R., Pangarkar, O., Wang, H., et~al. (2025).
\newblock K2-v2: A 360-open, reasoning-enhanced llm.
\newblock {\em arXiv preprint arXiv:2512.06201}.

\bibitem[Londo{\~n}o et~al., 2024]{londono2024fairness}
Londo{\~n}o, L., Hurtado, J.~V., Hertz, N., Kellmeyer, P., Voeneky, S., and Valada, A. (2024).
\newblock Fairness and bias in robot learning.
\newblock {\em Proceedings of the IEEE}, 112(4):305--330.

\bibitem[MacKenzie, 1992]{mackenzie1992fitts}
MacKenzie, I.~S. (1992).
\newblock Fitts' law as a research and design tool in human-computer interaction.
\newblock {\em Human-computer interaction}, 7(1):91--139.

\bibitem[MacKenzie, 2018]{mackenzie2018fitts}
MacKenzie, I.~S. (2018).
\newblock Fitts’ law.
\newblock {\em The wiley handbook of human computer interaction}, 1:347--370.

\bibitem[McCrea and Eng, 2005]{mccrea2005consequences}
McCrea, P.~H. and Eng, J.~J. (2005).
\newblock Consequences of increased neuromotor noise for reaching movements in persons with stroke.
\newblock {\em Experimental brain research}, 162(1):70--77.

\bibitem[Miller, 2019]{miller2019explanation}
Miller, T. (2019).
\newblock Explanation in artificial intelligence: Insights from the social sciences.
\newblock {\em Artificial intelligence}, 267:1--38.

\bibitem[Miller, 2023]{miller2023explainable}
Miller, T. (2023).
\newblock Explainable ai is dead, long live explainable ai! hypothesis-driven decision support using evaluative ai.
\newblock In {\em Proceedings of the 2023 ACM conference on fairness, accountability, and transparency}, pages 333--342.

\bibitem[Mo et~al., 2025]{mo2025mid}
Mo, K., Shi, Y., Weng, W., Zhou, Z., Liu, S., Zhang, H., and Zeng, A. (2025).
\newblock Mid-training of large language models: A survey.
\newblock {\em arXiv preprint arXiv:2510.06826}.

\bibitem[Molnar, 2020]{molnar2020interpretable}
Molnar, C. (2020).
\newblock {\em Interpretable machine learning}.
\newblock Lulu.com.

\bibitem[Mori et~al., 2012]{mori2012uncanny}
Mori, M., MacDorman, K.~F., and Kageki, N. (2012).
\newblock The uncanny valley [from the field].
\newblock {\em IEEE Robotics \& automation magazine}, 19(2):98--100.

\bibitem[Murthy and Sanneman, 2025]{murthy2025llms}
Murthy, A.~B. and Sanneman, L. (2025).
\newblock Llms need to go beyond computational confidence metrics to establish trust.
\newblock In {\em Proceedings of the AAAI Symposium Series}, volume~7, pages 131--136.

\bibitem[{NASA DART Mishap Investigation Board}, 2006]{nasa2006dart}
{NASA DART Mishap Investigation Board} (2006).
\newblock {Overview of the DART Mishap Investigation Results}.
\newblock Technical report, National Aeronautics and Space Administration, Washington, DC.
\newblock Accessed: 2026-02-04.

\bibitem[{National Highway Traffic Safety Administration}, 2024]{nhtsa2024fsd}
{National Highway Traffic Safety Administration} (2024).
\newblock {Preliminary Evaluation PE24-031: FSD Collisions in Reduced Roadway Visibility Conditions}.
\newblock Technical Report PE24-031, Office of Defects Investigation.
\newblock Accessed: 2026-02-04.

\bibitem[Nieto~Agraz et~al., 2025]{nieto2025robot}
Nieto~Agraz, C., Hinrichs, P., Eichelberg, M., and Hein, A. (2025).
\newblock Is the robot spying on me? a study on perceived privacy in telepresence scenarios in a care setting with mobile and humanoid robots.
\newblock {\em International Journal of Social Robotics}, 17(3):363--377.

\bibitem[Nikolaidis and Shah, 2013]{nikolaidis2013human}
Nikolaidis, S. and Shah, J. (2013).
\newblock Human-robot cross-training: Computational formulation, modeling and evaluation of a human team training strategy.
\newblock In {\em 2013 8th ACM/IEEE international conference on human-robot interaction (HRI)}, pages 33--40. IEEE.

\bibitem[Nomura et~al., 2015]{nomura2015children}
Nomura, T., Uratani, T., Kanda, T., Matsumoto, K., Kidokoro, H., Suehiro, Y., and Yamada, S. (2015).
\newblock Why do children abuse robots?
\newblock In {\em Proceedings of the tenth annual ACM/IEEE international conference on human-robot interaction extended abstracts}, pages 63--64.

\bibitem[Okada et~al., 2005]{Okada2005HumanoidMG}
Okada, K., Ogura, T., Haneda, A., Fujimoto, J., Gravot, F., and Inaba, M. (2005).
\newblock Humanoid motion generation system on hrp2-jsk for daily life environment.
\newblock {\em IEEE International Conference Mechatronics and Automation, 2005}, 4:1772--1777 Vol. 4.

\bibitem[Ostrom, 2011]{ostrom2011background}
Ostrom, E. (2011).
\newblock Background on the institutional analysis and development framework.
\newblock {\em Policy studies journal}, 39(1):7--27.

\bibitem[Ouyang et~al., 2022]{ouyang2022training}
Ouyang, L., Wu, J., Jiang, X., Almeida, D., Wainwright, C., Mishkin, P., Zhang, C., Agarwal, S., Slama, K., Ray, A., et~al. (2022).
\newblock Training language models to follow instructions with human feedback.
\newblock {\em Advances in neural information processing systems}, 35:27730--27744.

\bibitem[Pan et~al., 2024]{pan2024fittsBenchmarkAssistedHR}
Pan, J., Eden, J., Oetomo, D., and Johal, W. (2024).
\newblock Using fitts’ law to benchmark assisted human-robot performance.
\newblock {\em arXiv preprint arXiv:2412.05412}.

\bibitem[Paterson, 2024]{paterson2024robot}
Paterson, M. (2024).
\newblock Why robot embodiment matters: questions of disability, race and intersectionality in the design of social robots.
\newblock {\em Medical Humanities}, 50(4):694--704.

\bibitem[Pathiraja et~al., 2024]{pathiraja2024fairness}
Pathiraja, B., Liu, C., and Senanayake, R. (2024).
\newblock Fairness in autonomous driving: Towards understanding confounding factors in object detection under challenging weather.
\newblock In {\em Data-Driven Autonomous Driving Simulation Workshop at the IEEE/CVF Computer Vision and Pattern Recognition Conference (CVPR-DDAS)}.

\bibitem[Peltzer et~al., 2022]{Peltzer2022arxiv}
Peltzer, O., Bouman, Amanda, K. S.-K. S.~R., Ott, J., Delecki, Harrison, S. M.-K. M.~J., Schwager, M., Burdick, J., and Agha-mohammadi, A.-a. (2022).
\newblock Fig-op: Exploring large-scale unknown environments on a fixed time budget.
\newblock In {\em IEEE/RSJ International Conference on Intelligent Robots and Systems (IROS)}.

\bibitem[{Physical Intelligence} et~al., 2025a]{intelligence2025pi}
{Physical Intelligence}, Amin, A., Aniceto, R.~J., Balakrishna, A., Black, K., Conley, K., Connors, G., Darpinian, J., Dhabalia, K., DiCarlo, J., Driess, D., Equi, M., Esmail, A., Fang, Y., Finn, C., Glossop, C., Godden, T., Goryachev, I., Groom, L., Hancock, H., Hausman, K., Hussein, G., Ichter, B., Jakubczak, S., Jen, R., Jones, T., Katz, B., Ke, L., Kuchi, C., Lamb, M., LeBlanc, D., Levine, S., Li-Bell, A., Lu, Y., Mano, V., Mothukuri, M., Nair, S., Pertsch, K., Ren, A.~Z., Sharma, C., Shi, L.~X., Smith, L., Springenberg, J.~T., Stachowicz, K., Stoeckle, W., Swerdlow, A., Tanner, J., Torne, M., Vuong, Q., Walling, A., Wang, H., Williams, B., Yoo, S., Yu, L., Zhilinsky, U., and Zhou, Z. (2025a).
\newblock $\pi$*0.6: a vla that learns from experience.
\newblock {\em ArXiv}, abs/2511.14759.

\bibitem[{Physical Intelligence} et~al., 2025b]{intelligence2025pi_}
{Physical Intelligence}, Black, K., Brown, N., Darpinian, J., Dhabalia, K., Driess, D., Esmail, A., Equi, M., Finn, C., Fusai, N., Galliker, M.~Y., Ghosh, D., Groom, L., Hausman, K., Ichter, B., Jakubczak, S., Jones, T., Ke, L., LeBlanc, D., Levine, S., Li-Bell, A., Mothukuri, M., Nair, S., Pertsch, K., Ren, A.~Z., Shi, L.~X., Smith, L., Springenberg, J.~T., Stachowicz, K., Tanner, J., Vuong, Q., Walke, H.~R., Walling, A., Wang, H., Yu, L., and Zhilinsky, U. (2025b).
\newblock $\pi$0.5: a vision-language-action model with open-world generalization.
\newblock {\em ArXiv}, abs/2504.16054.

\bibitem[Pian et~al., 2024]{pian2024continual}
Pian, W., Nan, Y., Deng, S., Mo, S., Guo, Y., and Tian, Y. (2024).
\newblock Continual audio-visual sound separation.
\newblock {\em Advances in Neural Information Processing Systems}, 37:76058--76079.

\bibitem[Purtill, 2025]{abc2025humanoid}
Purtill, J. (2025).
\newblock How to design a humanoid robot a human will trust.
\newblock Accessed: 2026-02-04.

\bibitem[Rafailov et~al., 2023]{rafailov2023direct}
Rafailov, R., Sharma, A., Mitchell, E., Manning, C.~D., Ermon, S., and Finn, C. (2023).
\newblock Direct preference optimization: Your language model is secretly a reward model.
\newblock {\em Advances in neural information processing systems}, 36:53728--53741.

\bibitem[Rahman et~al., 2021]{rahman2021run}
Rahman, Q.~M., Corke, P., and Dayoub, F. (2021).
\newblock Run-time monitoring of machine learning for robotic perception: A survey of emerging trends.
\newblock {\em IEEE Access}, 9:20067--20075.

\bibitem[Rodriguez-Guerra et~al., 2021]{rodriguez2021human}
Rodriguez-Guerra, D., Sorrosal, G., Cabanes, I., and Calleja, C. (2021).
\newblock Human-robot interaction review: Challenges and solutions for modern industrial environments.
\newblock {\em IEEE Access}, 9:108557--108578.

\bibitem[Sacino et~al., 2022]{sacino2022human}
Sacino, A., Cocchella, F., De~Vita, G., Bracco, F., Rea, F., Sciutti, A., and Andrighetto, L. (2022).
\newblock Human-or object-like? cognitive anthropomorphism of humanoid robots.
\newblock {\em PLOS one}, 17(7):e0270787.

\bibitem[Sagar et~al., 2025a]{sagar2025from}
Sagar, S., Duan, J., Vasudevan, S., Zhou, Y., Amor, H.~B., Fox, D., and Senanayake, R. (2025a).
\newblock From mystery to mastery: Failure diagnosis for improving manipulation policies.
\newblock In {\em Second Workshop on Out-of-Distribution Generalization in Robotics at RSS 2025}.

\bibitem[Sagar et~al., 2025b]{Sagar2024arxiv_xaitcavrob}
Sagar, S., Taparia, A., Mankodiya, H., Bidare, H., Zhou, Y., and Senanayake, R. (2025b).
\newblock Trustworthy conceptual explanations for neural networks in robot decision-making.
\newblock In {\em International Conference on Intelligent Robots and Systems (IROS)}.

\bibitem[Sagar et~al., 2024a]{sagar2024failures}
Sagar, S., Taparia, A., and Senanayake, R. (2024a).
\newblock Failures are fated, but can be faded: Characterizing and mitigating unwanted behaviors in large-scale vision and language models.
\newblock In {\em International Conference on Machine Learning}, pages 42999--43023. PMLR.

\bibitem[Sagar et~al., 2024b]{10.5555/3692070.3693822}
Sagar, S., Taparia, A., and Senanayake, R. (2024b).
\newblock Failures are fated, but can be faded: characterizing and mitigating unwanted behaviors in large-scale vision and language models.
\newblock In {\em Proceedings of the 41st International Conference on Machine Learning}, ICML'24. JMLR.org.

\bibitem[Sakagami et~al., 2002]{sakagami2002intelligent}
Sakagami, Y., Watanabe, R., Aoyama, C., Matsunaga, S., Higaki, N., and Fujimura, K. (2002).
\newblock The intelligent asimo: System overview and integration.
\newblock In {\em IEEE/RSJ international conference on intelligent robots and systems}, volume~3, pages 2478--2483. IEEE.

\bibitem[Sanneman and Shah, 2020]{sanneman2020trust}
Sanneman, L. and Shah, J.~A. (2020).
\newblock Trust considerations for explainable robots: A human factors perspective.
\newblock {\em arXiv preprint arXiv:2005.05940}.

\bibitem[Sanneman and Shah, 2022]{sanneman2022situation}
Sanneman, L. and Shah, J.~A. (2022).
\newblock The situation awareness framework for explainable ai (safe-ai) and human factors considerations for xai systems.
\newblock {\em International Journal of Human--Computer Interaction}, 38(18-20):1772--1788.

\bibitem[Sanneman et~al., 2023]{sanneman2023information}
Sanneman, L., Tucker, M., and Shah, J. (2023).
\newblock An information bottleneck characterization of the understanding-workload tradeoff.
\newblock {\em arXiv preprint arXiv:2310.07802}.

\bibitem[Senanayake, 2024]{senanayake2024role}
Senanayake, R. (2024).
\newblock The role of predictive uncertainty and diversity in embodied ai and robot learning.
\newblock {\em arXiv preprint arXiv:2405.03164}.

\bibitem[Senanayake, 2025]{Senanayake_2025}
Senanayake, R. (2025).
\newblock {\em The Role of Predictive Uncertainty and Diversity in Embodied AI and Robot Learning}, page 148–182.
\newblock Cambridge University Press.

\bibitem[Senanayake and Goonetilleke, 2013]{senanayake2013superiority}
Senanayake, R. and Goonetilleke, R.~S. (2013).
\newblock Superiority of freehand pointing.
\newblock In {\em Proceedings of the Human Factors and Ergonomics Society Annual Meeting}, volume~57, pages 1639--1642. SAGE Publications Sage CA: Los Angeles, CA.

\bibitem[Senanayake et~al., 2013]{senanayake2013model}
Senanayake, R., Hoffmann, E.~R., and Goonetilleke, R.~S. (2013).
\newblock A model for combined targeting and tracking tasks in computer applications.
\newblock {\em Experimental brain research}, 231(3):367--379.

\bibitem[Sferrazza et~al., 2024]{sferrazza2024humanoidbench}
Sferrazza, C., Huang, D.-M., Lin, X., Lee, Y., and Abbeel, P. (2024).
\newblock Humanoidbench: Simulated humanoid benchmark for whole-body locomotion and manipulation.
\newblock {\em arXiv preprint arXiv:2403.10506}.

\bibitem[Sha et~al., 2024]{sha2024forgetting}
Sha, A.~S., Nunes, B.~P., and Haller, A. (2024).
\newblock " forgetting" in machine learning and beyond: A survey.
\newblock {\em arXiv preprint arXiv:2405.20620}.

\bibitem[Sharan et~al., 2024]{sharan2024plan}
Sharan, S., Zhao, R., Wang, Z., Chinchali, S.~P., et~al. (2024).
\newblock Plan diffuser: Grounding llm planners with diffusion models for robotic manipulation.
\newblock In {\em Bridging the Gap between Cognitive Science and Robot Learning in the Real World: Progresses and New Directions}.

\bibitem[Shneiderman, 2020]{shneiderman2020human}
Shneiderman, B. (2020).
\newblock Human-centered artificial intelligence: Reliable, safe \& trustworthy.
\newblock {\em International Journal of Human--Computer Interaction}, 36(6):495--504.

\bibitem[Silva et~al., 2016]{silva2016task}
Silva, P.~L., Bootsma, R.~J., Figueiredo, P. R.~P., Avelar, B.~S., de~Andrade, A. G.~P., Fonseca, S.~T., and Mancini, M.~C. (2016).
\newblock Task difficulty and inertial properties of hand-held tools: An assessment of their concurrent effects on precision aiming.
\newblock {\em Human movement science}, 48:161--170.

\bibitem[Smith and Zeller, 2017]{smith2017death}
Smith, D.~H. and Zeller, F. (2017).
\newblock The death and lives of hitchbot: The design and implementation of a hitchhiking robot.
\newblock {\em Leonardo}, 50(1):77--78.

\bibitem[Song et~al., 2020]{song2020toward}
Song, S., She, Y., Wang, J., and Su, H.-J. (2020).
\newblock Toward tradeoff between impact force reduction and maximum safe speed: Dynamic parameter optimization of variable stiffness robots.
\newblock {\em Journal of Mechanisms and Robotics}, 12(5):054503.

\bibitem[Stein et~al., 2022]{stein2022power}
Stein, J.-P., Cimander, P., and Appel, M. (2022).
\newblock Power-posing robots: The influence of a humanoid robot’s posture and size on its perceived dominance, competence, eeriness, and threat.
\newblock {\em International Journal of Social Robotics}, 14(6):1413--1422.

\bibitem[Stiennon et~al., 2020]{stiennon2020learning}
Stiennon, N., Ouyang, L., Wu, J., Ziegler, D., Lowe, R., Voss, C., Radford, A., Amodei, D., and Christiano, P.~F. (2020).
\newblock Learning to summarize with human feedback.
\newblock {\em Advances in neural information processing systems}, 33:3008--3021.

\bibitem[Sweller, 2011]{sweller2011cognitive}
Sweller, J. (2011).
\newblock Cognitive load theory.
\newblock In {\em Psychology of learning and motivation}, volume~55, pages 37--76. Elsevier.

\bibitem[Takayama et~al., 2008]{takayama2008beyond}
Takayama, L., Ju, W., and Nass, C. (2008).
\newblock Beyond dirty, dangerous and dull: what everyday people think robots should do.
\newblock In {\em Proceedings of the 3rd ACM/IEEE international conference on Human robot interaction}, pages 25--32.

\bibitem[Takeda et~al., 2019]{takeda2019explanation}
Takeda, M., Sato, T., Saito, H., Iwasaki, H., Nambu, I., and Wada, Y. (2019).
\newblock Explanation of fitts’ law in reaching movement based on human arm dynamics.
\newblock {\em Scientific reports}, 9(1):19804.

\bibitem[Talbott, 2025]{talbott2025waymo}
Talbott, S.~J. (2025).
\newblock The waymo vandalism incident: A wake-up call for the av industry.
\newblock {\em Forbes}.

\bibitem[Tan et~al., 2024]{tan2024judgebench}
Tan, S., Zhuang, S., Montgomery, K., Tang, W.~Y., Cuadron, A., Wang, C., Popa, R.~A., and Stoica, I. (2024).
\newblock Judgebench: A benchmark for evaluating llm-based judges.
\newblock {\em arXiv preprint arXiv:2410.12784}.

\bibitem[Tanimu and Abada, 2025]{tanimu2025addressing}
Tanimu, J.~A. and Abada, W. (2025).
\newblock Addressing cybersecurity challenges in robotics: A comprehensive overview.
\newblock {\em Cyber Security and Applications}, 3:100074.

\bibitem[Taparia et~al., 2025]{taparia2025vlc}
Taparia, A., Ngu, N., Leiva, M., Kricheli, J.~S., Corcoran, J., Bastian, N.~D., Simari, G., Shakarian, P., and Senanayake, R. (2025).
\newblock Vlc fusion: Vision-language conditioned sensor fusion for robust object detection.
\newblock {\em arXiv preprint arXiv:2505.12715}.

\bibitem[Team et~al., 2025]{team2025longcat}
Team, M.~L., Li, B., Lei, B., Wang, B., Rong, B., Wang, C., Zhang, C., Gao, C., Zhang, C., Sun, C., et~al. (2025).
\newblock Longcat-flash technical report.
\newblock {\em arXiv preprint arXiv:2509.01322}.

\bibitem[Tellex et~al., 2011]{tellex2011understanding}
Tellex, S., Kollar, T., Dickerson, S., Walter, M., Banerjee, A., Teller, S., and Roy, N. (2011).
\newblock Understanding natural language commands for robotic navigation and mobile manipulation.
\newblock In {\em Proceedings of the AAAI conference on artificial intelligence}, volume~25, pages 1507--1514.

\bibitem[Tevet et~al., 2023]{tevet2023human}
Tevet, G., Raab, S., Gordon, B., Shafir, Y., Cohen-Or, D., and Bermano, A.~H. (2023).
\newblock Human motion diffusion model.
\newblock In {\em The Eleventh International Conference on Learning Representations}.

\bibitem[Thumser et~al., 2018]{thumser2018fitts}
Thumser, Z.~C., Slifkin, A.~B., Beckler, D.~T., and Marasco, P.~D. (2018).
\newblock Fitts’ law in the control of isometric grip force with naturalistic targets.
\newblock {\em Frontiers in psychology}, 9:560.

\bibitem[Tint et~al., 2024]{tint2024expressivityarenallmsexpressinformation}
Tint, J., Sagar, S., Taparia, A., Raines, K., Pathiraja, B., Liu, C., and Senanayake, R. (2024).
\newblock Expressivityarena: Can llms express information implicitly?
\newblock {\em CoRR}, abs/2411.08010.

\bibitem[Tong et~al., 2024]{tong2024advancements}
Tong, Y., Liu, H., and Zhang, Z. (2024).
\newblock Advancements in humanoid robots: A comprehensive review and future prospects.
\newblock {\em IEEE/CAA Journal of Automatica Sinica}, 11(2):301--328.

\bibitem[Torabi et~al., 2018]{10.5555/3304652.3304697}
Torabi, F., Warnell, G., and Stone, P. (2018).
\newblock Behavioral cloning from observation.
\newblock In {\em Proceedings of the 27th International Joint Conference on Artificial Intelligence}, IJCAI'18, page 4950–4957. AAAI Press.

\bibitem[Tu et~al., 2025]{tu2025survey}
Tu, C., Zhang, X., Weng, R., Li, R., Zhang, C., Bai, Y., Yan, H., Wang, J., and Cai, X. (2025).
\newblock A survey on llm mid-training.
\newblock {\em arXiv preprint arXiv:2510.23081}.

\bibitem[Vasudevan et~al., 2025]{vasudevan2025strategic}
Vasudevan, S., Sagar, S., and Senanayake, R. (2025).
\newblock Strategic vantage selection for learning viewpoint-agnostic manipulation policies.
\newblock {\em arXiv e-prints}, pages arXiv--2506.

\bibitem[Vorvoreanu et~al., 2023]{vorvoreanu2023responsible}
Vorvoreanu, M., Heger, A., Passi, S., Dhanorkar, S., Kahn, Z., and Wang, R. (2023).
\newblock Responsible ai maturity model.
\newblock {\em Technical Report MSR-TR-2023-26}.

\bibitem[Wan et~al., 2023]{wan2023performance}
Wan, Y., Sun, J., Peers, C., Humphreys, J., Kanoulas, D., and Zhou, C. (2023).
\newblock Performance and usability evaluation scheme for mobile manipulator teleoperation.
\newblock {\em IEEE Transactions on Human-Machine Systems}, 53(5):844--854.

\bibitem[Wang et~al., 2024]{wang2024lami}
Wang, C., Hasler, S., Tanneberg, D., Ocker, F., Joublin, F., Ceravola, A., Deigmoeller, J., and Gienger, M. (2024).
\newblock Lami: Large language models for multi-modal human-robot interaction.
\newblock In {\em Extended Abstracts of the CHI Conference on Human Factors in Computing Systems}, pages 1--10.

\bibitem[Wang et~al., 2023]{wang2023differential}
Wang, Y., Wang, Q., Zhao, L., and Wang, C. (2023).
\newblock Differential privacy in deep learning: Privacy and beyond.
\newblock {\em Future Generation Computer Systems}, 148:408--424.

\bibitem[Watanabe et~al., 2013]{watanabe2013cooking}
Watanabe, Y., Nagahama, K., Yamazaki, K., Okada, K., and Inaba, M. (2013).
\newblock Cooking behavior with handling general cooking tools based on a system integration for a life-sized humanoid robot.
\newblock {\em Paladyn, Journal of Behavioral Robotics}, 4(2):63--72.

\bibitem[Wei et~al., 2022]{wei2022emergent}
Wei, J., Tay, Y., Bommasani, R., Raffel, C., Zoph, B., Borgeaud, S., Yogatama, D., Bosma, M., Zhou, D., Metzler, D., Chi, E.~H., Hashimoto, T., Vinyals, O., Liang, P., Dean, J., and Fedus, W. (2022).
\newblock Emergent abilities of large language models.
\newblock {\em Transactions on Machine Learning Research}.

\bibitem[Wei et~al., 2025]{wei2025memorization}
Wei, J., Zhang, Y., Zhang, L.~Y., Ding, M., Chen, C., Ong, K.-L., Zhang, J., and Xiang, Y. (2025).
\newblock Memorization in deep learning: A survey.
\newblock {\em ACM Computing Surveys}, 58(4):1--35.

\bibitem[Wilding and Boffey, 2025]{wilding2025police}
Wilding, M. and Boffey, D. (2025).
\newblock Uk police forces lobbied to use biased facial recognition technology.
\newblock Accessed: 2026-02-04.

\bibitem[Winfield and Jirotka, 2018]{winfield2018ethical}
Winfield, A.~F. and Jirotka, M. (2018).
\newblock Ethical governance is essential to building trust in robotics and artificial intelligence systems.
\newblock {\em Philosophical Transactions of the Royal Society A: Mathematical, Physical and Engineering Sciences}, 376(2133):20180085.

\bibitem[Wolf et~al., 2025]{wolf2025diffusion}
Wolf, R.~P., Shi, Y., Liu, S., and Rayyes, R. (2025).
\newblock Diffusion models for robotic manipulation: A survey.
\newblock {\em Frontiers in Robotics and AI}, 12:1606247.

\bibitem[Wu et~al., 2023]{wu2023ar}
Wu, T., Fan, Z., Liu, X., Zheng, H.-T., Gong, Y., Jiao, J., Li, J., Guo, J., Duan, N., Chen, W., et~al. (2023).
\newblock Ar-diffusion: Auto-regressive diffusion model for text generation.
\newblock {\em Advances in Neural Information Processing Systems}, 36:39957--39974.

\bibitem[Xiao et~al., 2025]{xiao2025robot}
Xiao, X., Liu, J., Wang, Z., Zhou, Y., Qi, Y., Jiang, S., He, B., and Cheng, Q. (2025).
\newblock Robot learning in the era of foundation models: A survey.
\newblock {\em Neurocomputing}, page 129963.

\bibitem[Xie et~al., 2023]{xie2023fitts}
Xie, Y., Zhou, R., and Qu, J. (2023).
\newblock Fitts’ law on the flight deck: evaluating touchscreens for aircraft tasks in actual flight scenarios.
\newblock {\em Ergonomics}, 66(4):506--523.

\bibitem[Xu and Gao, 2023]{xu2023enabling}
Xu, W. and Gao, Z. (2023).
\newblock Enabling human-centered ai: A methodological perspective.
\newblock {\em arXiv preprint arXiv:2311.06703}.

\bibitem[Xu et~al., 2024]{xu2024surveyroboticsfoundationmodels}
Xu, Z., Wu, K., Wen, J., Li, J., Liu, N., Che, Z., and Tang, J. (2024).
\newblock A survey on robotics with foundation models: toward embodied ai.

\bibitem[Yaacoub et~al., 2022]{yaacoub2022robotics}
Yaacoub, J.-P.~A., Noura, H.~N., Salman, O., and Chehab, A. (2022).
\newblock Robotics cyber security: Vulnerabilities, attacks, countermeasures, and recommendations.
\newblock {\em International Journal of Information Security}, 21(1):115--158.

\bibitem[Zare et~al., 2024]{zare2024survey}
Zare, M., Kebria, P.~M., Khosravi, A., and Nahavandi, S. (2024).
\newblock A survey of imitation learning: Algorithms, recent developments, and challenges.
\newblock {\em IEEE Transactions on Cybernetics}.

\bibitem[Zhang et~al., 2025a]{zhang2025review}
Zhang, H., Wu, J., Fan, J., An, Y., Jin, X., Cui, D., and Yang, Y. (2025a).
\newblock A review of fall coping strategies for humanoid robots.
\newblock {\em Journal of Bionic Engineering}, 22(2):480--512.

\bibitem[Zhang and Zhong, 2025]{zhang2025decoding}
Zhang, J. and Zhong, L. (2025).
\newblock Decoding emotion in the deep: A systematic study of how llms represent, retain, and express emotion.
\newblock {\em arXiv preprint arXiv:2510.04064}.

\bibitem[Zhang et~al., 2025b]{zhang2025vlabench}
Zhang, S., Xu, Z., Liu, P., Yu, X., Li, Y., Gao, Q., Fei, Z., Yin, Z., Wu, Z., Jiang, Y.-G., et~al. (2025b).
\newblock Vlabench: A large-scale benchmark for language-conditioned robotics manipulation with long-horizon reasoning tasks.
\newblock In {\em Proceedings of the IEEE/CVF International Conference on Computer Vision}, pages 11142--11152.

\bibitem[Zhao et~al., 2025]{zhao2025multimodal}
Zhao, W., Gangaraju, K., and Yuan, F. (2025).
\newblock Multimodal perception-driven decision-making for human-robot interaction: a survey.
\newblock {\em Frontiers in Robotics and AI}, 12:1604472.

\bibitem[Zhou et~al., 2026]{zhou2026spread}
Zhou, S., Duan, W., Yin, X., Scalia, M., Hao, R., Weng, N., Funke, G., Tolston, M., Freeman, G., Schelble, B., et~al. (2026).
\newblock The spread of trust and distrust in human-ai teams.
\newblock {\em Applied Ergonomics}, 130:104648.

\bibitem[Zhu, 2024]{zhu2024datapyramid}
Zhu, Y. (2024).
\newblock Data pyramid and data flywheel for robotic foundation models.
\newblock Technical talk slides, University of Texas at Austin / Robot Perception and Learning Lab.

\bibitem[Zimmerli et~al., 2012]{zimmerli2012validation}
Zimmerli, L., Krewer, C., Gassert, R., M{\"u}ller, F., Riener, R., and L{\"u}nenburger, L. (2012).
\newblock Validation of a mechanism to balance exercise difficulty in robot-assisted upper-extremity rehabilitation after stroke.
\newblock {\em Journal of neuroengineering and rehabilitation}, 9(1):6.

\bibitem[Zitkovich et~al., 2023]{zitkovich2023rt}
Zitkovich, B., Yu, T., Xu, S., Xu, P., Xiao, T., Xia, F., Wu, J., Wohlhart, P., Welker, S., Wahid, A., et~al. (2023).
\newblock Rt-2: Vision-language-action models transfer web knowledge to robotic control.
\newblock In {\em Conference on Robot Learning}, pages 2165--2183. PMLR.

\bibitem[Zou et~al., 2025]{zou2025synergy}
Zou, L., Zhang, Z., Mavilidi, M., Chen, Y., Herold, F., Ouwehand, K., and Paas, F. (2025).
\newblock The synergy of embodied cognition and cognitive load theory for optimized learning.
\newblock {\em Nature Human Behaviour}, pages 1--9.

\end{thebibliography}
\end{document}